%% file: survey.tex
\documentclass{article}
\usepackage[preprint]{colm2026_conference}

\usepackage{microtype}
\usepackage{hyperref}
\usepackage{url}
\usepackage{booktabs}
\usepackage{pifont}
\newcommand{\cmark}{\ding{51}}
\newcommand{\xmark}{\ding{55}}
\usepackage{graphicx}
\usepackage{forest}
\usepackage{tikz}
\usepackage{colortbl}
\usetikzlibrary{decorations.pathreplacing,fit,backgrounds,calc,arrows.meta,shadows}
\usepackage{xcolor}
\usepackage{lineno}
\usepackage{amsmath}
\usepackage{amssymb}
\usepackage{amsthm}
\usepackage{algorithm}
\usepackage{algorithmic}

\newtheorem{proposition}{Proposition}

\definecolor{darkblue}{rgb}{0, 0, 0.5}
\hypersetup{colorlinks=true, citecolor=darkblue, linkcolor=darkblue, urlcolor=darkblue, hypertexnames=false}

\definecolor{hidden-draw}{RGB}{20,68,106}
\definecolor{hidden-pink}{RGB}{255,245,247}
\tikzset{ver/.style={rounded corners, draw=hidden-draw, fill=white, text=hidden-draw, font=\bfseries}}
\tikzset{leaf/.style={draw=hidden-draw, fill=hidden-pink, text width=20em, text=black, font=\small}}

\title{Model Merging in the Era of Large Language Models\\Methods, Applications, and Future Directions}

\author{Mingyang Song \& Mao Zheng \\
Large Language Model Department \\Tencent, China\\
\texttt{nickmysong@tencent.com}}

\begin{document}

\ifcolmsubmission
\linenumbers
\fi

\maketitle

\begin{abstract}
Model merging combines the parameters of multiple neural networks into a single model without additional training. As fine-tuned large language models (LLMs) proliferate, merging offers a computationally efficient alternative to ensembles and full retraining, enabling practitioners to compose specialized capabilities at minimal cost. This survey examines model merging in the LLM era through the \textbf{FUSE} taxonomy, organized along \textbf{F}oundations, \textbf{U}nification Strategies, \textbf{S}cenarios, and \textbf{E}cosystem. We first establish the theoretical underpinnings of merging, including loss landscape geometry and mode connectivity, then systematically review the algorithmic space spanning weight averaging, task vector arithmetic, sparsification-enhanced methods, mixture-of-experts architectures, and evolutionary optimization. We further examine downstream applications across multi-task learning, safety alignment, domain specialization, and federated learning, and survey the supporting ecosystem of tools and evaluation benchmarks. Finally, we identify key open challenges and future directions, aiming to equip researchers and practitioners with a structured foundation for advancing model merging.
\end{abstract}

\section{Introduction}

Model merging combines the parameters of multiple trained neural networks into a single unified model that inherits their respective capabilities without the inference-time overhead of ensembles. Whereas ensembles aggregate predictions from separate models at runtime, merging operates directly in weight space, producing a single model whose computational cost matches that of any individual constituent. The idea traces back to the discovery that independently trained networks are connected by low-loss curves in weight space~\citep{DBLP:conf/nips/GaripovIPVW18} and to Stochastic Weight Averaging~\citep{DBLP:conf/uai/IzmailovPGVW18}, which showed that averaging checkpoints along an optimization trajectory yields flatter, better-generalizing solutions. Large-scale pretrained language models~\citep{DBLP:conf/naacl/DevlinCLT19} broadened the scope dramatically; models fine-tuned for entirely different tasks could now be merged into unified systems with multi-task capabilities~\citep{DBLP:conf/iclr/IlharcoRWSHF23}, turning what had been a checkpoint-averaging heuristic into a principled approach to neural network knowledge composition.

Pretrained foundation models such as LLaMA~\citep{DBLP:journals/corr/abs-2302-13971}, Mistral~\citep{DBLP:journals/corr/abs-2310-06825}, Qwen~\citep{DBLP:journals/corr/abs-2309-16609,qwen3}, DeepSeek~\citep{DBLP:journals/corr/abs-2401-02954}, Gemma~\citep{DBLP:journals/corr/abs-2403-08295}, and CLIP~\citep{DBLP:conf/icml/RadfordKHRGASAM21} have created conditions that make model merging both practically viable and scientifically compelling. The key insight is that models fine-tuned from a shared pretrained initialization tend to reside within the same loss basin, exhibiting linear mode connectivity~\citep{DBLP:conf/icml/FrankleDBMG20} that permits direct weight interpolation without traversing high-loss barriers~\citep{DBLP:conf/nips/GaripovIPVW18}. This theoretical grounding, combined with the proliferation of openly available fine-tuned large language models, has sparked broad interest in merging methodologies.

Rapid methodological progress over the past two years has transformed model merging from a niche technique into a mainstream component of LLM development. Task vector arithmetic~\citep{DBLP:conf/iclr/IlharcoRWSHF23}, sparsification-enhanced methods such as TIES-Merging~\citep{DBLP:conf/nips/YadavTCRB23} and DARE~\citep{DBLP:journals/corr/abs-2311-03099}, and activation-informed approaches~\citep{DBLP:journals/corr/abs-2502-02421} now enable the principled combination of specialized models~\citep{DBLP:journals/corr/abs-2405-17461}. Merged models have reached top rankings on competitive leaderboards including the Open LLM Leaderboard, with strategic model combination yielding capabilities exceeding those of individual fine-tuned variants~\citep{DBLP:conf/icml/WortsmanIGRLMNF22,DBLP:journals/corr/abs-2403-19522}. Dedicated toolkits such as MergeKit~\citep{DBLP:journals/corr/abs-2403-13257} have further democratized access, putting advanced merging strategies within reach of practitioners without deep algorithmic expertise.

Several recent surveys have begun to address aspects of this rapidly evolving field. \citet{DBLP:journals/corr/abs-2408-07666} provide a broad treatment of model merging techniques, while \citet{DBLP:journals/corr/abs-2408-07057} focus specifically on mixture-of-experts style composition. Concurrent works include \citet{ruan2025from}, who review model merging from task-specific models to unified systems, and \citet{lotfi2026tradeoffs}, who systematically analyze trade-offs among ensembling, merging, and routing for parameter-efficient experts. However, no existing survey provides a unified treatment that simultaneously connects theoretical foundations explaining \emph{why} merging succeeds, algorithmic strategies for \emph{how} to merge effectively, downstream scenarios illustrating \emph{where} merging provides value, and the practical ecosystem supporting deployment. In particular, the full spectrum from loss surface geometry and mode connectivity theory through evaluation methodologies and community tooling has not been covered within a single coherent taxonomic framework.

\begin{figure*}[t]
\centering
\resizebox{0.92\textwidth}{!}{
\begin{tikzpicture}[>=latex, semithick]
\definecolor{moyu}{HTML}{2B2D2F}       
\definecolor{qinghua}{HTML}{2E5FA1}    
\definecolor{songbai}{HTML}{2B6E4F}    
\definecolor{liuli}{HTML}{C68B2C}      
\definecolor{zhusha}{HTML}{C03F3D}     
\definecolor{yanzhi}{HTML}{8B2252}     
\definecolor{qinghui}{HTML}{6B818C}    
\tikzset{
 nd/.style={draw=#1!45!black, fill=#1!6, rounded corners=3pt,
 minimum width=1.6cm, minimum height=0.6cm, align=center,
 font=\footnotesize, line width=0.5pt},
 nd/.default={qinghui},
 mn/.style={draw=qinghua!45!black, fill=qinghua!#1, rounded corners=3pt,
 minimum width=2.4cm, minimum height=0.55cm, align=center,
 font=\footnotesize, line width=0.5pt},
}
\node[nd=moyu] (pt) at (0, 0) {$\theta_0$};
\node[nd=orange] (m1) at (2.6, 1.2) {$\theta_1$};
\node[nd=orange] (m2) at (2.6, 0) {$\theta_2$};
\node[nd=orange] (mk) at (2.6,-1.2) {$\theta_K$};
\node[color=gray!40, font=\small] at (2.6,-0.6) {$\vdots$};
\draw[->, dashed, thin, gray!40] (pt.east) -- ++(0.3,0) |- (m1.west);
\draw[->, dashed, thin, gray!40] (pt.east) -- (m2.west);
\draw[->, dashed, thin, gray!40] (pt.east) -- ++(0.3,0) |- (mk.west);
\coordinate (busL) at (4.4, 0);
\draw[gray!25, line width=4pt, rounded corners=2pt] (4.4, 1.2) -- (4.4,-1.2);
\draw[->, thin, gray!35] (m1.east) -- (4.4, 1.2);
\draw[->, thin, gray!35] (m2.east) -- (4.4, 0);
\draw[->, thin, gray!35] (mk.east) -- (4.4,-1.2);
\node[mn=8] (w) at (6.6, 1.2) {Weight Avg.~(\S3)};
\node[mn=6] (t) at (6.6, 0.4) {Task Vectors~(\S4)};
\node[mn=5] (s) at (6.6,-0.4) {MoE / Routing~(\S5)};
\node[mn=4] (r) at (6.6,-1.2) {Search-Based~(\S5)};
\draw[->, thin, gray!35] (4.4, 1.2) -- (w.west);
\draw[->, thin, gray!35] (4.4, 0.4) -- (t.west);
\draw[->, thin, gray!35] (4.4,-0.4) -- (s.west);
\draw[->, thin, gray!35] (4.4,-1.2) -- (r.west);
\draw[gray!25, line width=4pt, rounded corners=2pt] (8.6, 1.2) -- (8.6,-1.2);
\draw[->, thin, blue!30] (w.east) -- (8.6, 1.2);
\draw[->, thin, blue!30] (t.east) -- (8.6, 0.4);
\draw[->, thin, blue!30] (s.east) -- (8.6,-0.4);
\draw[->, thin, blue!30] (r.east) -- (8.6,-1.2);
\node[nd=red, minimum width=1.8cm] (mg) at (10.2, 0) {$\theta_{\text{merged}}$};
\draw[->, semithick, gray!50] (8.6, 0) -- (mg.west);
\node[nd=gray, minimum width=1.8cm] (ev) at (12.4, 0) {Evaluation};
\draw[->, semithick, gray!50] (mg) -- (ev);
\foreach \x/\lab in {0/{Pretrained}, 2.6/{Fine-tuned}, 6.6/{Merging}, 10.2/{Output}, 12.4/{Evaluate}} {
 \node[font=\tiny\bfseries, color=gray!45, above=0pt] at (\x, 1.9) {\textsc{\lab}};
}
\end{tikzpicture}
}
\caption{Overview of the model merging pipeline. Task-specific models $\{\theta_1,\dots,\theta_K\}$, obtained by fine-tuning a shared pretrained checkpoint~$\theta_0$, are combined via one of several strategies (Sections~3--5) into a single model~$\theta_{\text{merged}}$, which is then evaluated on all source tasks. Gray vertical bars represent the many-to-many mapping between model inputs and merging strategies.}
\label{fig:pipeline}
\end{figure*}
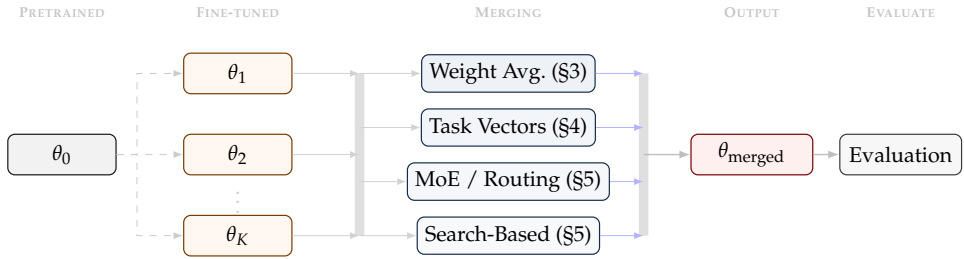

To fill this gap, this survey provides a structured treatment of model merging in the LLM era, bridging theoretical understanding with practical application. Table~\ref{tab:foundation_models} summarizes the major open-weight foundation models commonly used in merging research. As shown in Figure~\ref{fig:taxonomy}, we organize the field through the \textbf{FUSE} taxonomy, a four-dimensional scheme encompassing \textbf{F}oundations (why merging works), \textbf{U}nification Strategies (how merging is performed), \textbf{S}cenarios (where merging provides value), and \textbf{E}cosystem (what resources support merging). Figure~\ref{fig:pipeline} provides a schematic overview of the model merging pipeline.

The main contributions are as follows.
\begin{enumerate}
\item We propose the FUSE taxonomy as a coherent organizational structure that categorizes model merging research across theoretical foundations, algorithmic strategies, application scenarios, and ecosystem infrastructure.
\item We provide an in-depth technical analysis of merging methodologies, explaining the mathematical principles underlying each approach and offering comparative insights into their strengths and limitations.
\item We conduct a systematic examination of application domains, synthesizing empirical evidence on when and how merging delivers practical benefits across multi-task generalization, multilingual transfer, preference alignment, and federated learning.
\item We identify open challenges and promising future research directions~\citep{DBLP:journals/corr/abs-2505-12082}, including theoretical gaps in understanding mergeability, scalability limitations, and opportunities for automated merging systems.
\end{enumerate}
The remainder of this survey proceeds as follows. Section~\ref{sec:theory} establishes theoretical foundations. Sections~\ref{sec:weight_avg}--\ref{sec:search} present unification strategies with increasing sophistication. Section~\ref{sec:applications} examines application scenarios. Section~\ref{sec:ecosystem} discusses the ecosystem, systematic failure modes (Section~\ref{sec:failure-modes}), open challenges, and future directions.

\begin{table}[t]
 \centering
 \footnotesize
 \caption{Overview of major open-weight foundation models commonly used in model merging research. These models serve as the base for fine-tuning and subsequent merging experiments.}
 \label{tab:foundation_models}
 \resizebox{\textwidth}{!}{
 \begin{tabular}{llccll}
 \toprule
 \textbf{Model Family} & \textbf{Organization} & \textbf{Parameters} & \textbf{Year} & \textbf{Architecture} & \textbf{Key Features} \\
 \midrule
 LLaMA~\citep{DBLP:journals/corr/abs-2302-13971} & Meta & 7B--65B & 2023 & Decoder-only & RoPE, Pre-norm, SwiGLU \\
 LLaMA-2~\citep{DBLP:journals/corr/abs-2307-09288} & Meta & 7B--70B & 2023 & Decoder-only & GQA, Extended context \\
 LLaMA-3~\citep{DBLP:journals/corr/abs-2407-21783} & Meta & 8B--405B & 2024 & Decoder-only & 128K context, GQA \\
 \midrule
 Mistral~\citep{DBLP:journals/corr/abs-2310-06825} & Mistral AI & 7B & 2023 & Decoder-only & Sliding window attention \\
 Mixtral~\citep{DBLP:journals/corr/abs-2401-04088} & Mistral AI & 8$\times$7B & 2024 & Sparse MoE & 8 experts, top-2 routing \\
 \midrule
 Qwen~\citep{DBLP:journals/corr/abs-2309-16609} & Alibaba & 1.8B--72B & 2023 & Decoder-only & Multilingual, Code \\
 Qwen2~\citep{DBLP:journals/corr/abs-2407-10671} & Alibaba & 0.5B--72B & 2024 & Decoder-only & GQA, Extended context \\
 Qwen3~\citep{qwen3} & Alibaba & 0.6B--235B & 2025 & Dense + MoE & Thinking mode, 32K--128K context \\
 \midrule
 DeepSeek~\citep{DBLP:journals/corr/abs-2401-02954} & DeepSeek & 7B--67B & 2024 & Decoder-only & MHA, Scaling law design \\
 DeepSeek-V2~\citep{DBLP:journals/corr/abs-2405-04434} & DeepSeek & 236B (21B active) & 2024 & MoE & MLA, DeepSeekMoE \\
 \midrule
 Gemma~\citep{DBLP:journals/corr/abs-2403-08295} & Google & 2B--7B & 2024 & Decoder-only & RoPE, GeGLU, MQA (2B) \\
 \bottomrule
 \end{tabular}
 }
\end{table}

\section{Theoretical Foundations of Model Merging}
\label{sec:theory}

\input{taxonomy}

Why does model merging work? We examine three interconnected pillars. First, the geometric structure of loss landscapes that permits effective weight interpolation~\citep{DBLP:conf/iclr/RenC025}, the phenomenon of mode connectivity that permits low-loss paths between independently trained solutions~\citep{DBLP:conf/nips/DraxlerVGH18,DBLP:conf/icml/LubanaBDKT23}, and weight-space symmetries, particularly permutation invariance~\citep{DBLP:journals/corr/abs-2211-08403,DBLP:conf/iclr/EntezariSDN22}, that must be resolved for effective merging~\citep{DBLP:conf/iclr/AinsworthHS23}.

\subsection{Loss Landscape Geometry and Convexity Properties in Neural Networks}

Successful model merging critically depends on the geometric properties of the loss surface. Neural network loss surfaces exhibit complex topologies with numerous local minima, saddle points, and regions of varying curvature. Yet a crucial observation underpinning model merging is that modern overparameterized networks display favorable geometric properties, deviating markedly from the pathological non-convex landscapes one might naively expect. \citet{DBLP:journals/corr/abs-2406-16300} systematize these properties, showing that when parameters vastly outnumber the minimum required for the task, the loss surface exhibits large connected regions of near-optimal solutions rather than isolated minima.

Basin characteristics, encompassing both width and flatness, play a central role in determining whether weight interpolation between two solutions will preserve model functionality. A loss basin can be formally characterized by considering the set of parameters $\mathcal{B}_\epsilon(\theta^*) = \{\theta : \mathcal{L}(\theta) < \mathcal{L}(\theta^*) + \epsilon\}$ surrounding a local minimum $\theta^*$. The width of this basin, often measured through the eigenspectrum of the Hessian matrix $\nabla^2 \mathcal{L}(\theta^*)$, determines how much perturbations from the optimum increase loss. Wide, flat basins permit considerable movement in parameter space without appreciable performance degradation, directly supporting weight averaging.

\citet{DBLP:conf/nips/GaripovIPVW18} discovered that independently trained neural networks can be connected by low-loss nonlinear paths (parameterized as B\'{e}zier curves) in weight space, even when linear interpolation between them encounters high-loss barriers. This finding established that the loss landscape contains simple, curved corridors connecting distinct modes, motivating subsequent work on identifying conditions under which even simpler \emph{linear} interpolation suffices. \citet{DBLP:conf/icml/FrankleDBMG20} later formalized the notion of \emph{linear} mode connectivity, showing that models sharing a common training trajectory, particularly those fine-tuned from a shared initialization, can be linearly interpolated without traversing high-loss barriers, a finding with important implications for model merging strategies that rely on simple averaging operations.

Loss surface geometry also has direct implications for generalization after merging. Flat minima, characterized by low curvature in the Hessian eigenspectrum, have been associated with improved generalization properties~\citep{DBLP:conf/uai/IzmailovPGVW18}. When merging models that reside in flat regions of the loss landscape, the resulting averaged parameters are more likely to remain within a low-loss region, thereby preserving generalization capabilities. Conversely, sharp minima~\citep{DBLP:conf/iclr/KeskarMNST17} with high curvature present challenges for merging, as even small perturbations induced by averaging can push parameters into high-loss regions. \citet{DBLP:conf/icml/LubanaBDKT23} reinforce this from a mechanistic perspective, showing that low-loss linear connectivity between two models implies similarity in their underlying computational mechanisms; models relying on fundamentally different features cannot be connected by simple interpolation paths.

Mathematical analysis of overparameterized networks reveals that the loss surface increasingly resembles a high-dimensional valley structure as model capacity grows. Let $\mathcal{M}$ denote the manifold of global minima; in the overparameterized regime, $\mathcal{M}$ becomes a connected, often convex subspace within which arbitrary linear combinations of solutions remain optimal. This geometric property, combined with rigorous proofs that wide networks trained with SGD are linearly mode connected~\citep{DBLP:conf/aistats/FerbachGGD24}, provides theoretical justification for why weight averaging succeeds when models share sufficient structural commonality, since their solutions lie within the same connected low-loss region, rendering interpolation a geometrically valid operation.

\subsection{Linear Mode Connectivity Theory and Basin Characteristics}
Mode connectivity, the existence of low-loss paths connecting different trained solutions in weight space, provides the central theoretical lens for understanding when and why model merging succeeds. Building on the geometric insights of the preceding section, this framework formalizes the conditions under which interpolation between independently trained solutions remains viable. This concept has become central to explaining the empirical success of weight interpolation methods, as thoroughly surveyed in~\citep{DBLP:journals/corr/abs-2408-07666}.

Formally, two neural network solutions $\theta_1$ and $\theta_2$ are said to exhibit \textbf{linear mode connectivity}~\citep{DBLP:conf/icml/FrankleDBMG20} if the loss along the linear interpolation path $\theta(\alpha) = (1-\alpha)\theta_1 + \alpha\theta_2$ for $\alpha \in [0,1]$ satisfies $\mathcal{L}(\theta(\alpha)) \leq \max(\mathcal{L}(\theta_1), \mathcal{L}(\theta_2)) + \epsilon$ for some small tolerance $\epsilon \geq 0$. When $\epsilon = 0$, the models reside in the same loss basin with no interpolation barrier. More generally, \textbf{nonlinear mode connectivity}~\citep{DBLP:conf/nips/GaripovIPVW18,DBLP:conf/nips/DraxlerVGH18} relaxes the path constraint to allow curved trajectories, typically parameterized as Bézier curves or piecewise linear paths. This distinction has practical implications, as linear mode connectivity~\citep{DBLP:conf/icml/FrankleDBMG20} enables direct weight averaging, while nonlinear connectivity may require more involved merging procedures.

Analysis of loss barriers, the maximum loss increase encountered along interpolation paths, provides quantitative measures for characterizing mode connectivity strength. \citet{DBLP:journals/corr/abs-2406-16300} present a systematization of methods for computing and interpreting these barriers, noting that barrier height correlates inversely with merging success. Specifically, given solutions $\theta_1$ and $\theta_2$, the loss barrier is defined as follows:
\begin{equation}
\Delta\mathcal{L} = \max_{\alpha \in [0,1]} \mathcal{L}(\theta(\alpha)) - \max(\mathcal{L}(\theta_1), \mathcal{L}(\theta_2)).
\label{eq:loss_barrier}
\end{equation}

\noindent When $\Delta\mathcal{L} \leq \epsilon$ for some small threshold $\epsilon$, we say the models exhibit \textbf{$\epsilon$-linear mode connectivity}~\citep{DBLP:conf/icml/FrankleDBMG20}. The relationship between loss barrier magnitude and merging success can be formally characterized through the following theorem.

\begin{proposition}[Merging Error Bound]
Let $\theta_1, \theta_2$ be two models with shared pretrained initialization $\theta_0$, and let $\theta_{\text{avg}} = \frac{1}{2}(\theta_1 + \theta_2)$. Under $L$-smoothness of the loss function $\mathcal{L}$ and the assumption that both models are at approximate stationary points ($\|\nabla\mathcal{L}(\theta_i)\| \approx 0$, as is typical for trained models), a standard result from convex analysis yields the following bound on the performance degradation of the averaged model (see, e.g., \citealt{DBLP:journals/corr/abs-2406-16300} for a refined, Hessian-dependent generalization):
\begin{equation}
\mathcal{L}(\theta_{\text{avg}}) - \frac{1}{2}(\mathcal{L}(\theta_1) + \mathcal{L}(\theta_2)) \leq \frac{L}{8}\|\theta_1 - \theta_2\|^2.
\label{eq:merging_bound}
\end{equation}
\end{proposition}

\noindent This bound reveals why shared initialization is critical, since models fine-tuned from the same pretrained checkpoint exhibit smaller $\|\theta_1 - \theta_2\|$, yielding tighter merging guarantees.

Shared pretrained initialization is the single most important factor enabling strong mode connectivity in the LLM era. When multiple models are fine-tuned from a common pretrained checkpoint, they exhibit considerably stronger linear mode connectivity~\citep{DBLP:conf/icml/FrankleDBMG20} compared to independently trained models~\citep{DBLP:conf/nips/GaripovIPVW18}. Fine-tuning typically induces relatively small parameter perturbations that remain within the same loss basin as the pretrained model, whereas training from random initialization explores disconnected regions of the loss surface.

Empirical investigations have used various path parameterizations to probe connectivity structure. Quadratic Bézier curves, parameterized as $\theta(\alpha) = (1-\alpha)^2\theta_1 + 2\alpha(1-\alpha)\theta_m + \alpha^2\theta_2$ with learnable midpoint $\theta_m$, can discover low-loss nonlinear paths even when linear barriers exist. \citet{DBLP:conf/iclr/RenC025} revisited mode connectivity through the lens of B\'{e}zier surfaces, providing richer characterizations of the loss landscape structure between trained solutions beyond simple linear paths.

Prior work on activation renormalization~\citep{DBLP:journals/corr/abs-2211-08403} revealed that apparent loss barriers often arise from feature distribution mismatches rather than genuine geometric separation, proposing renormalization procedures to restore connectivity. This finding suggests that many observed failures of linear mode connectivity~\citep{DBLP:conf/icml/FrankleDBMG20} stem from correctable statistical artifacts rather than core incompatibility between solutions.

Understanding basin characteristics extends beyond barrier analysis to encompass basin width and shape. As shown by \citet{DBLP:conf/icml/LubanaBDKT23}, the success of interpolation between two models depends not only on loss barrier height but also on the degree to which the models share underlying computational mechanisms. These theoretical insights establish that shared pretraining serves as a powerful mechanism for ensuring models occupy connected regions amenable to linear combination. However, even when models theoretically reside in connected basins, the success of direct weight averaging depends critically on how the models are represented, specifically whether their hidden units are aligned, a consideration governed by the symmetries inherent in neural network weight spaces.

\subsection{Weight Space Symmetries and Permutation Invariance}
These fundamental symmetries, particularly permutation invariance, substantially impact the feasibility and methodology of model merging. Permutation invariance arises from the fact that reordering hidden units within a layer, along with corresponding adjustments to incoming and outgoing weight matrices, produces a functionally equivalent network. This mathematical property implies that the number of distinct parameterizations encoding identical input-output mappings grows combinatorially with network depth and width, creating major challenges for direct weight space operations.

Consider a feedforward network with weight matrices $W^{(l)}$ connecting layer $l-1$ to layer $l$. For any permutation matrix $P^{(l)}$ applied to the hidden units of layer $l$, the transformation $(W^{(l)}, W^{(l+1)}) \mapsto (P^{(l)}W^{(l)}, W^{(l+1)}(P^{(l)})^T)$ preserves the network's function exactly. For a network with $L$ hidden layers each containing $n$ units, this yields $(n!)^L$ equivalent parameterizations, a symmetry group whose size renders the loss surface highly non-convex despite potential functional simplicity. The detailed treatment in~\citep{DBLP:conf/iclr/AinsworthHS23} emphasizes that this symmetry greatly complicates the interpretation of distances and interpolations in weight space, as two networks computing identical functions may appear arbitrarily distant under Euclidean metrics.

For model merging, the implications are immediate and severe. When averaging weights from two independently trained networks, the hidden unit orderings are essentially arbitrary and unrelated, causing the averaged network to combine misaligned features in semantically meaningless ways. Formally, if $\theta_A$ and $\theta_B$ represent two solutions related by permutation such that $\theta_B = \pi(\theta_A)$ for some permutation $\pi$, then $\frac{1}{2}(\theta_A + \theta_B)$ generally lies far from both solutions in functional space despite occupying their geometric midpoint. This observation explains why naive weight averaging of independently trained networks typically produces catastrophic performance degradation, even when both source networks achieve comparable accuracy.

A theoretical framework connecting permutation alignment~\citep{DBLP:conf/iclr/AinsworthHS23} to optimal transport provides rigorous methods for resolving this challenge. Given two networks with weight matrices $W_A$ and $W_B$, finding the optimal permutation reduces to solving an assignment problem that minimizes some distance metric between aligned representations. Practical implementations use the Hungarian method or approximate transport solvers to establish correspondence between hidden units based on their functional roles rather than arbitrary indexing. In a complementary line of work, \citet{DBLP:conf/aistats/FerbachGGD24} prove that sufficiently wide two-layer networks trained with SGD are linearly mode connected with high probability, using convergence rates of empirical measures in Wasserstein distance; this provides rigorous theoretical support for the empirical success of permutation-based alignment methods.

The REPAIR framework~\citep{DBLP:journals/corr/abs-2211-08403} showed that even after permutation alignment, residual distribution mismatches in intermediate activations can create apparent barriers to interpolation. Permutation-aligned networks may still exhibit incompatible feature statistics that cause interpolated networks to produce out-of-distribution internal representations. This insight motivates post-alignment normalization procedures that reconcile activation distributions, further reducing barriers to successful merging.

Shared initialization provides a natural mechanism for symmetry breaking that circumvents permutation ambiguity entirely. When multiple models are fine-tuned from identical pretrained weights, the optimization trajectories implicitly preserve hidden unit correspondence established during pretraining. This explains the empirical observation that fine-tuned models exhibit considerably better mergeability than independently trained counterparts~\citep{DBLP:conf/icml/WortsmanIGRLMNF22}. The pretrained initialization effectively anchors each hidden unit to a specific functional role, ensuring that corresponding parameters across fine-tuned variants encode semantically related modifications.

Recent systematic investigations have quantified the relationship between alignment quality and merging success, revealing that architectural choices and training procedures influence the degree to which permutation alignment is necessary. Transformer architectures with their structured attention mechanisms exhibit different symmetry properties than fully-connected networks. Together with the loss surface geometry and mode connectivity results of the preceding subsections, these symmetry considerations form a complete theoretical picture, namely that merging succeeds when models share a loss basin (geometry), can be connected by low-loss paths (connectivity), and have their hidden units aligned (symmetry). The practical question becomes how to verify and enforce these conditions, which we formalize next as mergeability prerequisites.

\subsection{Prerequisites and Conditions for Successful Model Merging}
The theoretical pillars above can be consolidated into a practical framework of \textit{mergeability conditions}, i.e., the set of prerequisites that must be satisfied for model merging to yield high-quality results. Understanding these conditions is essential for practitioners seeking to predict whether a given set of models can be successfully combined and for researchers developing new merging algorithms with broader applicability.

Perhaps the most critical prerequisite for successful model merging is shared pretrained initialization, which establishes a common reference point in weight space from which all candidate models diverge during fine-tuning. Formally, consider a pretrained model with parameters $\theta_{\text{pre}}$ and a set of fine-tuned models $\{\theta_1, \theta_2, \ldots, \theta_K\}$ each obtained by optimizing task-specific objectives starting from $\theta_{\text{pre}}$. The shared initialization ensures that fine-tuned parameters remain within the same loss basin, so that linear interpolation $\theta_{\alpha} = \sum_{k=1}^{K} \alpha_k \theta_k$ with $\sum_k \alpha_k = 1$ traverses low-loss regions rather than crossing high-loss barriers~\citep{DBLP:conf/nips/GaripovIPVW18,DBLP:conf/icml/FrankleDBMG20}. This basin-sharing property sharply distinguishes the fine-tuning regime from independent training, where models converge to disconnected minima separated by substantial energy barriers.

Architectural compatibility is a second essential condition, requiring both structural identity and representational alignment. Models must share identical architectures (layer configurations, activation functions, and normalization schemes) for element-wise parameter operations to be well-defined. Transformer architectures~\citep{DBLP:conf/nips/VaswaniSPUJGKP17}, with their residual connections and layer normalization, maintain more consistent feature representations across fine-tuning variants, enabling parameter combination without the severe interference observed in earlier network designs. The residual stream structure effectively constrains the magnitude of per-layer modifications, keeping fine-tuned models closer to the pretrained initialization.

Training procedure similarity also strongly influences merging success. Models fine-tuned with similar hyperparameters (learning rate, batch size, optimizer) tend to follow comparable trajectories through weight space, yielding stronger linear mode connectivity~\citep{DBLP:conf/icml/FrankleDBMG20}. Conversely, marked procedural divergence can cause models to settle in distinct sub-basins within the broader pretrained basin, introducing interpolation barriers that degrade merge quality.

A complete theoretical framework for predicting capability preservation during merging remains elusive but can be partially characterized through the lens of task vector interference. When fine-tuning induces parameter modifications in complementary subspaces, different parameters or opposing directions for different tasks, merging preserves individual capabilities through linear superposition. However, when task-specific adaptations overlap appreciably and conflict in direction, the merged model exhibits degraded performance on all constituent tasks. Activation-informed approaches~\citep{DBLP:journals/corr/abs-2502-02421} further show that incorporating activation-level information into the merging process can help preserve capabilities by selectively prioritizing critical weights, highlighting that mergeability assessment should account for both parameter-space and representation-space factors. This observation motivates the development of mergeability prediction methods that analyze both parameter-space and representation-space compatibility, supporting informed decisions about which models can be successfully combined. Yet while these conditions provide actionable guidance, the theoretical foundations underlying them remain incompletely understood, particularly for large-scale language models with billions of parameters.

\subsection{Open Theoretical Questions and Research Frontiers}
Several important gaps persist between observed merging phenomena and rigorous theoretical explanations, presenting fertile ground for future research. The most pressing open question concerns the mechanisms by which pretrained large language models exhibit such strong linear mode connectivity~\citep{DBLP:conf/icml/FrankleDBMG20} following independent fine-tuning on disparate tasks. While shared initialization creates a common loss basin, our understanding of why this basin remains sufficiently broad and convex to accommodate diverse task-specific adaptations remains incomplete, particularly as model scale increases to billions of parameters. \citet{DBLP:journals/corr/abs-2406-16300} analyze the topological conditions under which linear mode connectivity arises, but extending such analysis to the scale and complexity of modern LLMs remains an open challenge.

How model scale relates to mergeability properties presents another theoretically unresolved frontier. Empirical observations suggest that larger models exhibit more favorable merging characteristics, with reduced loss barriers and improved capability retention compared to smaller counterparts. However, a formal mathematical account of this phenomenon remains elusive. One hypothesis posits that increased overparameterization expands the effective dimensionality of the loss basin, providing more "room" for task-specific adaptations to coexist without interference. An alternative perspective suggests that the emergent representations in larger models may be more compositional and modular, naturally supporting the superposition of task-specific modifications. Resolving these hypotheses through controlled scaling experiments remains an important open direction.

Emerging research directions seek to develop predictive frameworks for merge compatibility that could obviate the need for exhaustive experimentation. The challenge can be formalized as learning a function $\mathcal{C}: (\theta_1, \theta_2, \ldots, \theta_k) \rightarrow [0, 1]$ that estimates the expected performance of merging models $\theta_1, \ldots, \theta_k$ without actually performing the merge. Promising approaches analyze task vector statistics, including angular relationships and magnitude distributions, to predict interference severity. However, translating these geometric insights into reliable predictive models requires overcoming the computational challenge of efficiently approximating high-dimensional loss surface properties.

Theoretical investigations into the fundamental limits of merging constitute another critical research frontier. While sparsification-based methods such as TIES-Merging and DARE have shown that parameter interference can be mitigated, the question of when interference becomes irreconcilable, representing a fundamental rather than technical limitation, remains open. \citet{DBLP:journals/corr/abs-2502-02421} show that activation-level statistics provide complementary information to parameter-level analysis, suggesting that a complete mergeability theory must account for both weight space and representation space phenomena. In addition, the conditions under which nonlinear interpolation paths offer strictly superior outcomes compared to linear combinations, and how such paths might be efficiently discovered, represent underexplored theoretical territories with large practical implications.

Finally, extending merging theory to accommodate architectural heterogeneity presents formidable theoretical challenges. Current frameworks assume exact architectural correspondence, but practical scenarios increasingly involve models with variations in layer counts, hidden dimensions, or attention mechanisms. Developing rigorous theoretical foundations for cross-architecture merging, potentially using representation alignment or functional correspondence rather than direct parameter mapping, could substantially expand the applicability of model merging techniques while requiring novel mathematical frameworks that transcend current weight-space formulations.

\begin{table}[t]
 \centering
 \scriptsize
 \caption{Theoretical foundations underlying model merging success. Each row summarizes a key theoretical concept, its mathematical basis, its implications for merging, and relevant seminal references. Together, these concepts explain \emph{when} and \emph{why} weight-space combination of fine-tuned models preserves or enhances performance.}
 \label{tab:theoretical_foundations_of_mod}
 \resizebox{\textwidth}{!}{
 \begin{tabular}{p{3.2cm}p{3.5cm}p{3.5cm}p{3.5cm}c}
 \toprule
 \textbf{Theoretical Concept} & \textbf{Key Approach} & \textbf{Key Feature(s)} & \textbf{Limitations/Notes} & \textbf{Year} \\
 \midrule
 \textbf{Loss Surface Geometry} \newline {\scriptsize \citep{DBLP:journals/corr/abs-2406-16300}} & Analyzes geometric properties of high-dimensional parameter space & Explains why weight interpolation yields functional networks & Complex landscapes make complete characterization difficult & 2024 \\
 \addlinespace
 \textbf{Linear Mode Connectivity} \newline {\scriptsize \citep{DBLP:conf/icml/FrankleDBMG20}} & Establishes that models from shared initialization are connected by low-loss linear paths & Central to explaining success of weight interpolation methods & Requires shared initialization for strong connectivity & 2020 \\
 \addlinespace
 \textbf{Permutation Invariance} \newline {\scriptsize \citep{DBLP:conf/iclr/AinsworthHS23}} & Exploits symmetry from reordering hidden units with weight adjustments & Multiple equivalent parameterizations for same function & Alignment required before merging independently trained models & 2023 \\
 \addlinespace
 \textbf{Weight Space Symmetries} \newline {\scriptsize \citep{DBLP:conf/iclr/AinsworthHS23}} & Analyzes fundamental symmetry properties affecting merging feasibility & Enables functionally equivalent network representations & Symmetry breaking needed for meaningful comparisons & 2023 \\
 \addlinespace
 \textbf{Mergeability Conditions} \newline {\scriptsize \citep{DBLP:journals/corr/abs-2406-16300}} & Establishes unified framework of prerequisites for successful merging & Predictive criteria for merge quality assessment & Conditions may vary across model architectures and tasks & 2024 \\
 \addlinespace
 \textbf{Shared Initialization Theory} \newline {\scriptsize \citep{DBLP:conf/icml/FrankleDBMG20}} & Explains role of common pretrained weights in enabling merging & Strong predictor of linear mode connectivity & Theoretical mechanisms not fully understood for LLMs & 2020 \\
 \bottomrule
 \end{tabular}
 }
\end{table}

Taken together, the theoretical foundations examined in this section, encompassing loss surface geometry, linear mode connectivity~\citep{DBLP:conf/icml/FrankleDBMG20}, and weight space symmetries, collectively establish the mathematical conditions under which model merging can succeed (summarized in Table~\ref{tab:theoretical_foundations_of_mod}). These insights guide understanding of when simple averaging will preserve model capabilities and when more targeted interventions become necessary. Building on these theoretical principles, the next section examines how they translate into practical merging algorithms.

\section{Weight-Space Averaging and Geometric Interpolation Methods}
\label{sec:weight_avg}
This section examines weight-space averaging and geometric interpolation, the most fundamental class of merging techniques. We organize these approaches along a spectrum of geometric sophistication, from static averaging with fixed coefficients, through trajectory-based methods that exploit optimization dynamics, to geometric interpolation (e.g., SLERP) that respects the manifold structure of weight spaces.

\subsection{Linear Averaging and Static Model Soups}
Weight averaging represents the most fundamental instantiation of model merging. It computes the element-wise arithmetic mean of model parameters, treating weight space as a Euclidean vector space. This formula admits a natural generalization to \emph{weighted averaging} $\theta_{\text{avg}} = \sum_i \alpha_i\theta_i$ with $\sum_i\alpha_i=1$, where the coefficients $\alpha_i$ can be tuned on a held-out set or determined via importance metrics such as Fisher information~\citep{DBLP:conf/nips/MatenaR22}. Given a collection of $N$ models with parameters $\{\theta_1, \theta_2, \ldots, \theta_N\}$, uniform weight averaging produces a merged model with parameters:

\[\theta_{\text{merged}} = \frac{1}{N} \sum_{i=1}^{N} \theta_i\]

This formulation assumes all source models share identical architectures with correspondingly positioned parameters. This works because of linear mode connectivity~\citep{DBLP:conf/icml/FrankleDBMG20}, whereby models fine-tuned from a common pretrained initialization tend to reside within the same loss basin, so that linear interpolation between solutions avoids high-loss regions. However, this connectivity assumption breaks down when models originate from independent random initializations or when fine-tuning procedures cause parameters to diverge substantially from the shared pretrained manifold, resulting in interpolation paths that traverse high-loss barriers~\citep{DBLP:journals/corr/abs-2406-16300}. For models lacking shared initialization, optimal transport methods~\citep{DBLP:conf/icml/SinghJ20} can align hidden units before averaging.

The Model Soups framework, introduced by~\citet{DBLP:conf/icml/WortsmanIGRLMNF22}, advances weight-space averaging by systematically studying the aggregation of multiple fine-tuned checkpoints. The framework establishes two primary aggregation strategies, the uniform soup and the greedy soup.

The \textbf{uniform soup} computes the arithmetic mean of all $N$ available model checkpoints without selection.

\[\theta_{\text{uniform}} = \frac{1}{N} \sum_{i=1}^{N} \theta_i\]

This baseline approach assumes all fine-tuned models contribute constructively to the aggregate solution. While computationally straightforward, the uniform soup may incorporate checkpoints whose parameter contributions introduce interference that degrades overall performance.

The \textbf{greedy soup} overcomes this limitation through iterative conditional selection, incorporating models into the aggregate only when their inclusion improves held-out validation performance. Formally, starting with an initial model $\theta_{\text{soup}} = \theta_1$, each candidate $\theta_i$ is added only if

\[\mathcal{L}\left(\frac{k \cdot \theta_{\text{soup}} + \theta_i}{k+1}\right) < \mathcal{L}(\theta_{\text{soup}})\]

where $k$ denotes the current number of models in the soup and $\mathcal{L}$ represents the validation loss function.

Empirically, greedy soup consistently outperformed both uniform soup and individual fine-tuned models across multiple benchmarks, including ImageNet and its distribution-shifted variants. While initially validated on vision models, this approach has since been applied to large language models. The key advantage is enhanced robustness to distribution shift. These improvements~\citep{DBLP:conf/icml/WortsmanIGRLMNF22} incur no additional inference cost compared to deploying a single model, as the merged architecture maintains identical computational requirements during deployment.

This selective aggregation strategy addresses a fundamental limitation of uniform averaging, namely that not all models contribute equally, and certain checkpoints introduce parameter interference that degrades performance. \citet{DBLP:conf/icml/WortsmanIGRLMNF22} show that weight-space operations can enable constructive task composition when models are appropriately combined, while unstructured aggregation frequently results in destructive interference. Several extensions have broadened the applicability of this framework. ColD Fusion~\citep{arxiv_2212_01378} adapts the approach to distributed multi-task fine-tuning, showing that iterative collaborative descent through weight averaging supports continuous base model improvement without data sharing across participants. Model Ratatouille~\citep{DBLP:journals/corr/abs-2212-10445} recycles diverse auxiliary models as initializations for parallel fine-tunings on the target task before averaging, improving out-of-distribution generalization. Transformer Fusion with Optimal Transport~\citep{arxiv_2310_05719} uses optimal transport for neuron matching before averaging to handle alignment challenges specific to Transformer architectures.

Beyond these core extensions, multiple lines of work expand the frontier of weight-space merging. Bayesian approaches to checkpoint selection~\citep{arxiv_2403_19390} maximize intermediate checkpoint value during LLM pretraining through careful aggregation. Soup-of-Experts~\citep{arxiv_2502_01804} learns to linearly combine a bank of expert parameters, enabling test-time instantiation of specialist models without retraining, while Extrapolation Merging~\citep{arxiv_2503_04834} shows that performance can continue improving beyond simple interpolation through extrapolation in weight space. Researchers have also explored the relationship between model aggregation and federated learning extensively, adapting weighted averaging strategies to scenarios involving heterogeneous client contributions~\citep{DBLP:conf/mlsys/LiSZSTS20,wang2024fedma} and non-identically distributed data.

Unlike exhaustive combinatorial search over model subsets, which becomes computationally prohibitive as the candidate pool grows exponentially, greedy selection provides an efficient approximation that scales linearly with the number of available checkpoints. However, this approach assumes that validation performance serves as a reliable proxy for generalization, an assumption that may not hold under substantial distribution shift between validation and deployment domains. Furthermore, theoretical analyses indicate that linear averaging methods, while computationally efficient, do not account for the permutation symmetries inherent in neural network weight spaces, motivating the development of alignment-based interpolation techniques discussed in later subsections. Beyond these structural considerations, the uniform weighting scheme treats all parameters as equally important, ignoring their heterogeneous contributions to model functionality, a limitation that importance-weighted approaches directly address.

\subsection{Importance-Weighted Averaging and Fisher Information}
Importance-weighted averaging methods confront this limitation directly by weighting each parameter according to its estimated contribution to model performance. In practice, certain parameters encode critical task-specific information while others remain largely unchanged from pretraining or exhibit high variance across fine-tuning runs. By incorporating second-order information about parameter sensitivity, these methods prioritize high-information parameters while attenuating noise from less consequential weights.

Fisher-weighted merging draws its theoretical foundation from the Fisher Information Matrix (FIM), which characterizes the local curvature of the loss landscape regarding model parameters. For a model with parameters $\theta$ and loss function $\mathcal{L}$, the Fisher Information Matrix is defined as $F = \mathbb{E}[\nabla_\theta \log p(x|\theta) \nabla_\theta \log p(x|\theta)^\top]$, where the expectation is taken over the data distribution. Parameters associated with high Fisher information values correspond to directions in weight space where small perturbations strongly affect model predictions, indicating their importance for task performance. Fisher-weighted averaging exploits this insight by computing the merged parameters as $\theta_{\text{merged}} = (\sum_{i} F_i)^{-1} \sum_{i} F_i\theta_i$, where $F_i$ denotes the Fisher Information Matrix computed for model $i$. This formulation effectively weights each model's contribution inversely proportional to its uncertainty, yielding a maximum a posteriori estimate under Gaussian assumptions about parameter distributions.

Full Fisher Information Matrix estimation imposes computational demands that render it intractable for contemporary large language models with billions of parameters. Practical implementations therefore use diagonal approximations, where only the diagonal entries $F_{i,jj}$ are retained, reducing both memory requirements and computational complexity from $O(d^2)$ to $O(d)$ where $d$ represents the parameter count. This diagonal Fisher-weighted averaging computes element-wise weighted means: $\theta_{\text{merged},j} = (\sum_{i} F_{i,jj} \theta_{i,j}) / (\sum_{i} F_{i,jj})$, treating each parameter independently. While this approximation discards correlation structure between parameters, empirical investigations show that diagonal Fisher weighting consistently outperforms uniform averaging across diverse task combinations~\citep{DBLP:conf/nips/MatenaR22}.

Alternative formulations use covariance statistics rather than Fisher information to estimate parameter importance. The RegMean approach~\citep{DBLP:conf/iclr/Jin0P023} computes inner products between model activations to construct weight matrices that account for the representational structure learned by each model. Specifically, RegMean solves a linear regression problem that minimizes the expected squared difference between the merged model's outputs and those of the constituent models, weighted by activation covariances. This representation-aware weighting implicitly captures parameter dependencies that diagonal Fisher approximations neglect, potentially yielding superior merging outcomes when models have learned complementary feature representations. WiSE-FT~\citep{DBLP:journals/corr/abs-2109-01903} shows that robust fine-tuning of zero-shot models can be obtained through simple weight interpolation between the pretrained and fine-tuned parameters, providing a rigorous approach for preserving zero-shot capabilities while acquiring task-specific expertise. Bayesian perspectives on weighted aggregation naturally extend these importance-based schemes, treating merged parameters as posterior estimates given observations from multiple training procedures. Localize-and-Stitch~\citep{DBLP:journals/corr/abs-2408-13656} introduces localized sparse merging, identifying tiny regions (1\% of parameters) essential for each task and stitching only those back into the pretrained model, enabling flexible and continual skill composition with minimal storage overhead. \citet{DBLP:conf/iclr/StoicaBHH24} propose ZipIt! for merging models trained on disjoint tasks by exploiting feature correlations, and~\citet{arxiv_2410_03617} provide systematic benchmarking of merging methods under controlled conditions.

Catastrophic forgetting~\citep{DBLP:journals/corr/KirkpatrickPRVD16} presents another critical consideration for model merging. When fine-tuned models are merged, the resulting model must retain capabilities from all constituent tasks without catastrophically forgetting any individual specialization. Fuse to Forget~\citep{arxiv_2311_07682} shows that model merging can be strategically used for bias reduction and selective memorization, revealing that the forgetting properties of merging can be harnessed as a feature rather than a limitation. Merging by Matching Models in Task Parameter Subspaces~\citep{arxiv_2312_04339} proposes identifying and aligning task-specific parameter subspaces before merging, yielding improved capability preservation. Elastic Weight Consolidation (EWC) and related continual learning methods provide theoretical frameworks for understanding which parameters are most critical for specific capabilities, insights that directly inform importance-weighted merging strategies.

Despite their theoretical appeal, importance-weighted methods introduce several practical challenges. Fisher information estimates require access to representative data samples from each constituent model's training distribution, information that may be unavailable when merging pretrained models obtained from external sources. Moreover, the approximation quality of diagonal Fisher estimates degrades in highly overparameterized networks where off-diagonal correlations capture essential structural relationships~\citep{DBLP:conf/nips/MatenaR22}. Unlike uniform averaging, which requires only the model weights themselves, importance-weighted approaches demand additional computational overhead for Fisher or covariance estimation, creating tradeoffs between merging quality and practical deployability that practitioners must navigate based on their specific resource constraints and accuracy requirements. Crucially, both uniform and importance-weighted averaging share a common assumption in that they operate on models trained independently, treating the aggregation problem as combining discrete endpoints without exploiting any structure from the optimization process itself.

\paragraph{Limitations and failure modes.}
Despite its simplicity and empirical success, linear averaging exhibits well-characterized failure modes. When source models are fine-tuned with markedly different learning rates or regularization strengths, the resulting parameter vectors may reside in distinct loss basins, causing their average to land on a high-loss barrier~\citep{DBLP:conf/icml/FrankleDBMG20}. Performance also degrades when the number of merged models grows large and the tasks are semantically diverse, a phenomenon termed \emph{task interference}. \citet{DBLP:journals/corr/abs-2212-10445} partially address this by recycling intermediate fine-tuned checkpoints from diverse auxiliary tasks (\emph{Model Ratatouille}), showing that averaging models exposed to a broader training distribution yields more robust merged solutions. At larger scales, \citet{arxiv_2412_04144} show that optimizing merging weights across a pool of checkpoints at the 100B-parameter scale can recover near-Pareto-optimal performance by recycling suboptimal checkpoints that would otherwise be discarded. Nevertheless, characterizing the precise conditions under which linear averaging provably succeeds remains an open theoretical challenge (see Section~\ref{sec:theory}).

\subsection{Trajectory-Based Averaging and Optimization Dynamics}
Trajectory-based averaging represents a fundamentally distinct approach that directly exploits the temporal structure of optimization, precisely the information that endpoint-focused methods discard. Rather than combining final checkpoints of separate training procedures, these techniques aggregate intermediate states collected along the optimization trajectory, exploiting the insight that weight-space averages of sequential iterates can approximate solutions residing near the center of flat loss basins. This geometric centering yields models exhibiting enhanced robustness to distribution shifts and improved generalization characteristics.

\textbf{Polyak Averaging.} The theoretical foundations for trajectory-based averaging trace to Polyak and Juditsky's seminal work on iterate averaging for stochastic approximation. Their analysis established that averaging SGD iterates achieves optimal asymptotic convergence rates, with the averaged solution $\bar{\theta}_n = \frac{1}{n}\sum_{i=1}^{n}\theta_i$ converging at rate $O(1/n)$ for strongly convex objectives, compared to $O(1/\sqrt{n})$ for the final iterate alone. This classical result provides the theoretical underpinning for modern trajectory averaging methods, showing that temporal averaging serves as an implicit regularizer that suppresses gradient noise while preserving convergence to stationary points.

\textbf{Stochastic Weight Averaging.} Stochastic Weight Averaging (SWA) adapts iterate averaging to modern deep learning by averaging checkpoints collected during the latter stages of training with cyclic or high constant learning rates~\citep{DBLP:conf/uai/IzmailovPGVW18}. Formally, given checkpoints $\theta_1, \theta_2, \ldots, \theta_n$ collected after initial convergence, SWA computes $\theta_{\text{SWA}} = \frac{1}{n}\sum_{i=1}^{n}\theta_i$. The key insight is that SGD with appropriately tuned learning rates explores a connected region corresponding to a single basin of attraction, and averaging over this exploration identifies the basin's centroid. In the context of LLMs, SWA has been successfully applied to smooth out the optimization trajectory during the final stages of pretraining or instruction tuning, leading to models with improved generation quality and reduced perplexity on held-out data. The flatness of these averaged solutions, measured via local entropy metrics, correlates strongly with improved generalization.

\textbf{SWAG and Uncertainty Quantification.} Stochastic Weight Averaging Gaussian (SWAG) extends SWA to approximate Bayesian inference by additionally capturing the covariance structure of the optimization trajectory. Beyond computing the mean $\theta_{\text{SWA}}$, SWAG maintains a low-rank approximation to the covariance matrix $\Sigma \approx \frac{1}{n-1}\sum_{i=1}^{n}(\theta_i - \theta_{\text{SWA}})(\theta_i - \theta_{\text{SWA}})^\top$, enabling posterior sampling for uncertainty estimation. This extension proves particularly valuable for calibrated predictions and out-of-distribution detection, achieving competitive performance with ensemble methods at considerably reduced computational cost.

\textbf{Lookahead Optimizer.} The Lookahead optimizer provides a complementary perspective by integrating trajectory averaging directly into the optimization procedure rather than applying it post-hoc. Lookahead maintains slow weights $\phi$ updated every $k$ steps toward the fast weights $\theta$ via $\phi \leftarrow \phi + \alpha(\theta - \phi)$, effectively performing online trajectory averaging. This approach reduces variance in gradient estimates and shows improved stability across learning rate choices, often reaching faster convergence while maintaining or improving final performance.

\textbf{Exponential Moving Average.} Exponential Moving Average (EMA) represents a continuously updated variant that maintains a shadow copy of parameters throughout training. At each iteration $t$, the EMA parameters are updated according to $\theta_{\text{EMA}}^{(t)} = \alpha \theta_{\text{EMA}}^{(t-1)} + (1-\alpha)\theta^{(t)}$, where $\alpha$ controls the temporal smoothing strength. Unlike SWA's uniform averaging, EMA implements exponentially-decaying weights prioritizing recent iterates while retaining memory of the entire trajectory. The selection of decay coefficient $\alpha$ varies systematically with model scale and training duration. For large language models, values close to 1 are typical, with larger models and longer training runs generally benefiting from stronger smoothing.

\textbf{Applications in LLM Training.} Trajectory averaging has shown clear benefits in large-scale language model development. In neural machine translation, checkpoint averaging of the final training checkpoints is standard practice~\citep{DBLP:conf/nips/VaswaniSPUJGKP17}, and SWA has been successfully applied to improve generalization in image classification~\citep{DBLP:conf/uai/IzmailovPGVW18}. In the model merging context, averaging models from different fine-tuning runs with varying hyperparameters can improve robustness on downstream tasks~\citep{DBLP:conf/icml/WortsmanIGRLMNF22}.

A bridge between trajectory-based averaging and broader model merging frameworks emerges when considering multiple training runs initialized from common starting points. When models share a pretrained initialization and undergo fine-tuning on related tasks, their optimization trajectories remain confined to overlapping basins of attraction, so that checkpoints can be meaningfully averaged despite originating from nominally independent procedures. This perspective bridges trajectory-based methods with task vector approaches, as temporal averaging within a single fine-tuning run and spatial averaging across related runs both operate by centering solutions within shared loss basins, with combined approaches yielding additive benefits.

Practical implementations of trajectory-based averaging require attention to checkpoint selection strategy and computational overhead. Regarding collection frequency, \citet{DBLP:conf/uai/IzmailovPGVW18} established that averaging checkpoints collected during the later stages of training with cyclic or constant learning rates provides near-optimal results, while more frequent collection yields diminishing returns. For memory efficiency, running average formulations avoid storing multiple checkpoints. SWA requires only two model copies (current and averaged), while EMA maintains constant memory overhead regardless of training duration. The robustness benefits to distribution shift have been validated in multiple studies, with SWA-averaged models exhibiting smaller performance degradation on domain-shifted evaluation sets compared to single checkpoints, attributed to the implicit regularization toward flatter minima.

Despite their effectiveness, trajectory-based methods share a common characteristic with the importance-weighted approaches discussed earlier, in that they perform arithmetic averaging in weight space, implicitly treating the parameter manifold as Euclidean. This assumption, while reasonable for iterates within a single optimization trajectory that remain geometrically proximate, becomes increasingly problematic when averaging models that have diverged substantially during independent fine-tuning, a limitation that motivates geometric interpolation techniques designed to preserve the statistical properties of constituent parameter sets.

\paragraph{Limitations.} Trajectory-based methods inherently require access to intermediate checkpoints from a single training run, making them inapplicable when only final model weights are available, as is typical in open-weight model repositories. In addition, the memory overhead of storing $T$ checkpoints ($T \times |\theta|$ parameters) can be prohibitive for billion-parameter models, though rank-1 covariance approximations such as those in SWAG partially mitigate this.

\subsection{Geometric Interpolation and Manifold-Aware Methods}
Since Euclidean assumptions become problematic for divergent models, geometric interpolation techniques offer structured alternatives that account for the curved structure of parameter spaces. When combining two parameter vectors $\theta_A$ and $\theta_B$ via linear interpolation $\theta_{\text{merged}} = (1-\alpha)\theta_A + \alpha\theta_B$, the resulting vector's magnitude typically satisfies $\|\theta_{\text{merged}}\| < (1-\alpha)\|\theta_A\| + \alpha\|\theta_B\|$ unless the vectors are perfectly aligned, leading to systematic magnitude shrinkage that can degrade model representations. This phenomenon proves particularly acute when merging models with markedly different weight configurations.

Spherical Linear Interpolation (SLERP) tackles magnitude preservation by constraining the interpolation path to lie on the hypersphere connecting two points in weight space. For two parameter vectors $\theta_A$ and $\theta_B$, SLERP computes the interpolated point as:

\[\text{SLERP}(\theta_A, \theta_B; \alpha) = \frac{\sin((1-\alpha)\Omega)}{\sin(\Omega)}\theta_A + \frac{\sin(\alpha\Omega)}{\sin(\Omega)}\theta_B\]

where $\Omega = \arccos(\theta_A \cdot \theta_B / \|\theta_A\|\|\theta_B\|)$ represents the angular distance between the two vectors. This formulation ensures that the interpolated vector maintains constant magnitude throughout the interpolation trajectory, traversing a geodesic arc rather than cutting through the interior of the sphere. Empirical evidence has shown that SLERP-based merging preserves representational geometry more faithfully than linear averaging, particularly for attention weight matrices where directional information carries semantic significance.

In practice, applying SLERP to neural network merging requires careful consideration of granularity. Layer-wise application, where SLERP operates independently on each layer's parameter tensor, has emerged as the dominant approach, as global application across flattened parameter vectors fails to capture the hierarchical structure of deep networks. Layer-wise SLERP yields superior capability retention compared to uniform linear averaging when merging models fine-tuned on semantically distant tasks, attributed to the preservation of layer-specific activation statistics.

Beyond spherical interpolation, manifold-aware methods encompass techniques that explicitly model the non-Euclidean structure of parameter spaces. Latent space interpolation approaches project weight configurations into lower-dimensional manifolds where interpolation can be performed with reduced interference before mapping back to the original parameter space. This approach acknowledges that the effective dimensionality of viable weight configurations is far lower than the ambient parameter count, allowing interpolation paths that remain within the manifold of high-performing solutions.

Variance preservation presents additional challenges beyond magnitude shrinkage. When averaging independent parameter vectors, the variance of the resulting distribution decreases proportionally to the number of models merged, potentially collapsing learned features toward uninformative mean values. Rescaling strategies that adjust merged parameters to match the original variance statistics partially mitigate this limitation, though they introduce hyperparameters governing the appropriate variance target. Geometric interpolation methods incorporating both magnitude and variance preservation consistently outperform naive averaging when source models exhibit low angular similarity, though the benefits diminish for models sharing high linear mode connectivity~\citep{DBLP:conf/icml/FrankleDBMG20} where Euclidean assumptions remain approximately valid. Yet all methods discussed so far treat model parameters as quantities to be combined, without explicitly modeling \emph{what changed} during fine-tuning. The next section introduces a complementary perspective, namely reconceptualizing fine-tuning as vector displacement in weight space, which enables targeted operations such as capability addition, negation, and interference-aware composition. Table~\ref{tab:weight_space_averaging_and_geo} compares the weight-space averaging and geometric interpolation methods discussed above.

\begin{table}[t]
 \centering
 \footnotesize
 \caption{Weight-space averaging and geometric interpolation methods for model merging. Methods range from simple uniform averaging to manifold-aware interpolation, progressively incorporating parameter importance (Fisher), temporal structure (SWA), and geometric properties (SLERP, manifold) to improve merge quality. ``Key Approach'' summarizes the merging mechanism; ``Limitations'' notes the primary constraint for each method.}
 \label{tab:weight_space_averaging_and_geo}
 \resizebox{\textwidth}{!}{
 \begin{tabular}{p{4.5cm}p{3.5cm}p{4.2cm}p{3.8cm}l}
 \toprule
 \textbf{Method} & \textbf{Key Approach} & \textbf{Key Feature(s)} & \textbf{Limitations/Notes} & \textbf{Year} \\
 \midrule
 \textbf{Uniform Averaging} \newline {\scriptsize (Model Soups)} & Element-wise arithmetic mean: $\theta_{\text{merged}} = \frac{1}{N}\sum_{i=1}^{N}\theta_i$ & Simple; no hyperparameters; treats weight space as Euclidean & Assumes equal parameter importance; suboptimal when models diverge substantially & 2022 \\
 \addlinespace
 \textbf{Fisher-Weighted Averaging} & Importance weighting via Fisher Information matrix & Accounts for heterogeneous parameter importance; preserves critical task-specific weights & Computationally expensive; requires data for Fisher estimation & 2022 \\
 \addlinespace
 \textbf{Stochastic Weight Averaging} \newline {\scriptsize (SWA) \citep{DBLP:conf/uai/IzmailovPGVW18}} & Average checkpoints along optimization trajectory & Exploits temporal structure; finds flatter minima; improves generalization & Limited to single training run; requires checkpoint storage & 2018 \\
 \addlinespace
 \textbf{SLERP} & Spherical linear interpolation on hypersphere & Preserves weight vector magnitude; avoids norm shrinkage in linear interpolation & Limited to pairwise merging; requires careful normalization & 2023 \\
 \addlinespace
  \textbf{Latent Space Merging} & Interpolation in activation/latent space rather than weights & Bypasses weight space geometry issues; often outperforms weight averaging & May require data access; increased inference complexity & 2024 \\
 \addlinespace
 \textbf{Manifold-Aware Interpolation} & Geodesic paths on parameter manifolds with permutation alignment & Respects non-Euclidean geometry; handles mode connectivity & Computationally intensive; requires solving alignment problem & 2023 \\
 \bottomrule
 \end{tabular}
 }
\end{table}

As reviewed above, weight-space averaging, whether through uniform linear combinations, Fisher-weighted importance schemes, trajectory-based accumulation along optimization paths, or geometry-aware interpolation on parameter manifolds, provides a powerful basis for combining multiple models without additional training, with performance gains stemming from the exploitation of loss surface connectivity and mode connectivity properties. Recent advances continue to refine these foundations along several complementary axes, revealing a consistent trend, namely that the field is moving from uniform, method-agnostic averaging toward context-sensitive strategies that adapt merging behavior to the specific characteristics of the models and layers being combined.

A first line of work pursues \textit{data-free adaptive merging} that replaces uniform coefficients with automatically derived, layer-wise or parameter-wise weights. FroM~\citep{arxiv_2506_02478} uses Frobenius norm statistics, MAGIC~\citep{arxiv_2512_19320} exploits magnitude calibration, and Orthogonal Model Merging~\citep{arxiv_2602_05943} preserves orthogonality during combination. The shared insight across these methods is that different layers contribute unequally to task-specific behavior, and norm-based or structural heuristics can approximate this heterogeneity without requiring any training data.

A second direction extends averaging from Euclidean weight space to \textit{richer geometric structures}. ACE-Merging~\citep{arxiv_2603_02945} uses adaptive covariance estimation, while Functionality-Oriented Merging~\citep{arxiv_2603_04972} operates on the Fisher--Rao manifold. These approaches trade computational simplicity for tighter alignment with the functional geometry of neural networks, yielding improvements particularly when source models have undergone substantial fine-tuning that violates local linearity assumptions.

Third, \textit{importance-weighted refinements} such as Dynamic Fisher-weighted Merging~\citep{arxiv_2504_18992}, which optimizes Fisher weights via Bayesian optimization, and RegMean++~\citep{arxiv_2508_03121} revisit the classical idea of curvature-based weighting with modern optimization tools, reducing the sensitivity of Fisher-based methods to hyperparameter choices.

Fourth, a growing body of work integrates merging into the \textit{training pipeline itself} rather than applying it only post hoc. WSM~\citep{arxiv_2507_17634} introduces decay-free learning rate schedules via checkpoint merging, Anytime Pretraining~\citep{arxiv_2602_03702} develops horizon-free schedules with weight averaging, and Mashup Learning~\citep{arxiv_2603_10156} accelerates fine-tuning by remixing checkpoints. This trend blurs the boundary between optimization and merging, suggesting that future training protocols may be co-designed with merging objectives.

Finally, \textit{scalable and training-free} approaches target practical deployment at LLM scale. Training-free LLM Merging~\citep{arxiv_2506_12379} proposes hierarchical pruning and scaling across model and layer levels for multi-task merging on LLaMA-3 and Qwen, while Harmonizing Diverse Models~\citep{arxiv_2510_14915} introduces consistency-aware layer-wise weights computed from intermediate activations to improve generation coherence in RAG systems.
Further works tackle weight averaging from the perspective of pretraining data mixture optimization~\citep{arxiv_2601_17858,arxiv_2602_00747,arxiv_2601_21115,arxiv_2505_16066}, parameter-efficient checkpoint merging~\citep{arxiv_2504_18580}, and bagging-based merging for robust text embeddings~\citep{arxiv_2602_05787}. While these approaches focus on holistically merging entire model checkpoints through various averaging strategies, the next section examines task vector arithmetic and sparsification-enhanced methods, which shift the perspective from combining full parameter sets to treating fine-tuning-induced weight differences as modular, composable units supporting addition, negation, and scaling for targeted multi-task composition.

\section{Task Vector Arithmetic and Sparsification-Enhanced Methods}
\label{sec:task-vector}
Rather than treating model parameters as interchangeable quantities to be averaged, an alternative perspective focuses on isolating and manipulating the task-specific knowledge encoded during fine-tuning. Task vector arithmetic, introduced by~\citet{DBLP:conf/iclr/IlharcoRWSHF23}, reconceptualizes model merging by treating the weight differences between fine-tuned models and their shared pretrained ancestor as composable vectors that support algebraic operations such as addition, negation, and scaling. This section also covers sparsification-enhanced variants such as TIES-Merging~\citep{DBLP:conf/nips/YadavTCRB23} and DARE~\citep{DBLP:journals/corr/abs-2311-03099} that tackle parameter interference and sign conflicts.

\subsection{Task Vector Formulation and Conceptual Framework}
The task vector framework formalizes how fine-tuning encodes task-specific knowledge within neural network parameters. \citet{DBLP:conf/iclr/IlharcoRWSHF23} introduced this foundational framework by proposing that the knowledge acquired during fine-tuning can be isolated, manipulated, and transferred through simple algebraic operations on model weights, making model specialization amenable to systematic composition and modification.

Formally, given a pretrained model with parameters $\theta_{\text{pre}}$ (denoted $\theta_0$ in Section~\ref{sec:theory}) and a model fine-tuned on task $t$ with resulting parameters $\theta_t$, the task vector $\tau_t$ is defined as the element-wise difference:
\begin{equation}
\tau_t = \theta_t - \theta_{\text{pre}}.
\label{eq:task_vector}
\end{equation}

This definition captures the parametric displacement induced by fine-tuning, effectively encoding the directional shift in weight space that transforms a general-purpose foundation model into a task-specialized variant. Based on this formulation, \citet{DBLP:conf/iclr/IlharcoRWSHF23} defined three fundamental arithmetic operations that constitute a versatile toolkit for training-free model editing, namely

\textbf{Task Addition (Multi-task Fusion).} To combine capabilities from multiple tasks, their respective task vectors can be added and applied to the pretrained model:
\begin{equation}
\theta_{\text{multi}} = \theta_{\text{pre}} + \sum_{i=1}^{n} \lambda_i \tau_i,
\label{eq:task_addition}
\end{equation}
where $\lambda_i$ denotes the scaling coefficient for task $i$. This operation supports the creation of multi-task models without joint training, assuming the task vectors reside in approximately orthogonal subspaces.

\textbf{Task Negation (Forgetting/Unlearning).} To remove an undesirable capability or behavior (e.g., toxicity) encoded in a task vector $\tau$, the vector can be subtracted from the pretrained model:
\begin{equation}
\theta_{\text{forget}} = \theta_{\text{pre}} - \lambda \tau,
\label{eq:task_negation}
\end{equation}
where $\lambda > 0$ controls the strength of the unlearning effect. This operation has found practical application in detoxification and bias removal for LLMs.

\textbf{Task Scaling (Capability Modulation).} The strength of a specific capability can be continuously modulated by scaling its task vector:
\begin{equation}
\theta_{\text{scaled}} = \theta_{\text{pre}} + \alpha \tau,
\label{eq:task_scaling}
\end{equation}
where $\alpha$ controls the intensity of the specialized behavior ($\alpha > 1$ amplifies the capability, while $0 < \alpha < 1$ attenuates it).

\textbf{Task Analogy (Relationship Transfer).} Extending the vector algebra metaphor, task analogy enables transferring the relationship between two tasks to a third domain. Given task vectors $\tau_A$, $\tau_B$, and a target task vector $\tau_C$, the analogy operation applies the directional difference between two tasks to a third as follows.
\begin{equation}
\theta_{\text{analogy}} = \theta_{\text{pre}} + \tau_C + (\tau_B - \tau_A).
\end{equation}
This operation mirrors classical word analogy reasoning (e.g., ``king $-$ man $+$ woman $=$ queen'') in embedding spaces, but operates directly on model parameters. As illustrated in Figure~\ref{fig:task_vector_ops}(d), task analogy permits compositional transfer of inter-task relationships without additional training.

Conceptually, this formulation's elegance lies in its interpretation. Fine-tuning is reconceptualized not as an opaque optimization process but as the application of a structured transformation that can be decomposed, analyzed, and recombined with other such transformations.

The task vector framework derives its power from treating these parameter deltas as first-class mathematical objects inhabiting a vector space~\citep{DBLP:conf/iclr/IlharcoRWSHF23}. Under this interpretation, the weight space of neural networks admits linear algebraic operations that preserve semantic meaning. When a model is fine-tuned on a specific capability, whether instruction following, mathematical reasoning, or domain expertise, the resulting task vector encodes a compact representation of that capability that remains meaningful when extracted from its original context and applied to compatible base models.

This formulation maintains deep connections to the theoretical foundations of mode connectivity discussed in earlier sections. The success of task vector arithmetic implicitly relies on the phenomenon of linear mode connectivity~\citep{DBLP:conf/icml/FrankleDBMG20} between fine-tuned models sharing a common pretrained initialization. When fine-tuned models reside within the same loss basin as their shared pretrained ancestor, the linear interpolation paths between them traverse regions of relatively low loss, permitting meaningful combination through arithmetic operations on their corresponding task vectors.

Task vector compatibility imposes important constraints on practical applications. Task vector operations are inherently limited to models with identical architectures since they rely on element-wise parameter correspondence. Both source models must derive from the same pretrained checkpoint, ensuring that their parameter spaces are aligned and that the extracted task vectors encode semantically coherent transformations. This architectural constraint reflects the key observation that task vectors represent relative displacements within a shared reference frame established by the common initialization.

Both the magnitude and distribution of task vector components carry meaningful information about the nature of fine-tuning adaptations. \citet{DBLP:conf/iclr/IlharcoRWSHF23} investigated how different fine-tuning procedures produce task vectors with distinct statistical properties, affecting downstream merging behavior. Empirical analyses reveal that task vectors exhibit considerable sparsity, with relatively few parameters undergoing large modifications during fine-tuning while the majority remain close to their pretrained values~\citep{DBLP:journals/corr/abs-2311-03099}. This sparsity observation has motivated various enhancement strategies, though the immediate practical power of task vectors emerges from the fundamental algebraic operations they support, addition, negation, and scaling, which enable training-free model editing without access to training data. Similar to how knowledge distillation~\citep{DBLP:journals/corr/HintonVD15} transfers knowledge between models through output matching, task vectors enable knowledge transfer through direct parameter manipulation. Figure~\ref{fig:task_vector_ops} illustrates the three primary task vector operations.

\input{task_vector}

\subsection{Basic Arithmetic Operations for Model Editing}
The three primary operations, addition, negation, and scaling, transform task vectors into versatile tools for training-free model editing.

\textbf{Task Vector Addition} represents the most widely used operation, enabling the combination of multiple specialized capabilities into a unified model. Given task vectors $\{\tau_1, \tau_2, \ldots, \tau_n\}$ from $n$ fine-tuned models, the merged parameters follow Eq.~\ref{eq:task_addition}. The seminal work~\citet{DBLP:conf/iclr/IlharcoRWSHF23} showed that this simple additive combination successfully transfers capabilities across diverse task categories, including instruction following, mathematical reasoning, and domain-specific knowledge adaptation in LLMs. The effectiveness of addition relies critically on the assumption that task-specific knowledge encoded in different task vectors occupies approximately orthogonal subspaces of the parameter manifold, allowing superposition without destructive interference. Empirical evidence indicates that task addition achieves competitive performance across numerous NLP benchmarks when source tasks exhibit complementary rather than conflicting requirements.

\textbf{Task Vector Negation} offers an effective mechanism for targeted behavior removal without requiring access to negative training examples or explicit unlearning procedures. By computing $\theta_{\text{edited}} = \theta_{\text{pre}} - \lambda \tau_{\text{unwanted}}$, practitioners can attenuate or eliminate specific model behaviors such as toxic language generation, demographic biases, or restricted domain knowledge~\citep{DBLP:conf/iclr/IlharcoRWSHF23}. The mathematical intuition follows from interpreting task vectors as directional displacements in capability space; reversing the direction moves the model away from the associated capability manifold. However, care must be taken, as the interaction between negated and retained capabilities may produce unintended side effects. Over-aggressive negation (large $\lambda$) can degrade general capabilities beyond the targeted behavior, and the impact may vary across demographic groups~\citep{DBLP:conf/iclr/IlharcoRWSHF23}.

Several extensions have refined the negation operation for specific applications. Elastic Weight Removal~\citep{arxiv_2303_17574} uses Fisher information to weigh parameter importance when subtracting a ``negative expert'' fine-tuned on hallucinated outputs, enabling more precise removal of unfaithful generation while preserving abstractive capabilities. Separate the Wheat from the Chaff~\citep{arxiv_2308_08090} improves negation through an extraction-before-subtraction approach that isolates deficiency-specific parameters before removal. Controlled text generation via language model arithmetic~\citep{dekoninck2023controlled} extends the arithmetic metaphor to the inference level, composing multiple LLMs through algebraic operations on output distributions rather than model parameters for fine-grained control over generation attributes. Forgetting before Learning~\citep{arxiv_2311_08011} demonstrates that parametric arithmetic can first subtract parameters encoding obsolete information before fine-tuning on new knowledge.

\textbf{Scaling Coefficients} provide fine-grained control over the magnitude of task contributions, enabling practitioners to balance competing objectives and prevent any single task from dominating the merged representation. The choice of scaling parameter $\lambda$ presents a fundamental tradeoff, as larger values increase task-specific performance but risk catastrophic interference with the pretrained model's general capabilities, while smaller values preserve base model competence at the cost of diminished task specialization. Unlike~\citet{DBLP:conf/iclr/IlharcoRWSHF23}, which primarily explored uniform scaling across tasks, subsequent investigations have revealed the benefits of task-specific coefficient tuning. \citet{DBLP:journals/corr/abs-2310-04742} showed that partially linearizing adapter modules and applying task arithmetic over the linearized adapters enables more principled coefficient selection by improving the linearity assumptions underlying weight-space composition.

An analogy between task arithmetic and classical linear algebra operations provides both conceptual clarity and practical limitations. While vector addition in Euclidean space enjoys exact superposition properties, task vector addition operates in a highly nonlinear parameter space where interactions between components can produce emergent behaviors not predicted by simple linear models. These nonlinear interactions manifest as parameter-level conflicts that can substantially degrade merged model performance, interference phenomena that demand systematic examination to support principled mitigation strategies.

\subsection{Parameter Interference and Sign Conflict Analysis}
Parameter-level conflicts identified in the preceding discussion manifest through three distinct interference mechanisms that emerge when multiple task vectors are combined through element-wise operations. Detailed examination of these mechanisms is essential for developing principled mitigation strategies that preserve the compositional benefits of task arithmetic while minimizing performance degradation.

\textbf{Sign Conflicts} represent perhaps the most pernicious form of parameter interference, occurring when task vectors from different fine-tuning objectives contain opposing directional updates for identical weight positions. Formally, for task vectors $\tau_1, \tau_2, \ldots, \tau_n$ derived from $n$ distinct fine-tuning procedures, a sign conflict at parameter position $j$ occurs when $\exists i, k: \text{sign}(\tau_i^{(j)}) \neq \text{sign}(\tau_k^{(j)})$ and both $|\tau_i^{(j)}|, |\tau_k^{(j)}| > \epsilon$ for some threshold $\epsilon$. When such conflicts arise, naive averaging produces parameter values that represent a compromised intermediate state, potentially satisfying neither task's requirements. \citet{DBLP:conf/nips/YadavTCRB23} provided detailed empirical analysis showing that sign conflicts correlate strongly with performance degradation in merged models, motivating their three-step trim-elect-merge procedure for conflict resolution.

\textbf{Magnitude Disparities} constitute a second critical interference mechanism, arising from the heterogeneous scaling of parameter updates across different fine-tuning procedures. Even when sign agreement exists, large differences in update magnitudes, quantified as $\max_i |\tau_i^{(j)}| / \min_k |\tau_k^{(j)}|$ for non-trivial updates, can cause certain tasks to dominate the merged representation disproportionately. This phenomenon is particularly pronounced when combining models fine-tuned with different learning rates, dataset sizes, or training durations~\citep{DBLP:conf/nips/YadavTCRB23}. The resulting merged model exhibits biased behavior favoring tasks whose fine-tuning produced larger magnitude updates, effectively drowning out the contributions of more conservatively trained variants. Empirical investigations reveal that real-world task vectors exhibit heavy-tailed magnitude distributions with substantial inter-task variance.

\textbf{Redundant Parameter Modifications} introduce a subtler but equally important source of interference, wherein low-magnitude updates that do not encode meaningful task-specific knowledge accumulate noise during the merging process. These redundant parameters arise from optimization dynamics that produce non-zero gradients even for weights irrelevant to the target task, creating spurious modifications that dilute the signal-to-noise ratio of the merged representation. \citet{DBLP:journals/corr/abs-2311-03099} formalized this observation by showing that delta parameters from fine-tuning are extremely sparse, with the vast majority contributing minimally to task performance while actively degrading merge quality through interference with truly informative modifications.

Interactions between these interference mechanisms produce complex, often non-additive effects that resist simple analytical characterization. Statistical frameworks for analyzing the joint distribution of task vector parameters reveal that interference severity depends strongly on the correlation structure between task vectors rather than their marginal statistics alone. Visualization techniques including parameter-wise conflict heatmaps and layer-wise interference scores have become diagnostic tools for identifying problematic regions before merging. This careful characterization of interference phenomena, spanning sign conflicts, magnitude disparities, and redundant modifications, directly motivates the sparsification-enhanced approaches that selectively retain informative parameters while eliminating conflict-inducing modifications.

\subsection{Sparsification-Enhanced Merging Methods}
Sparsification-enhanced methods mitigate these interference phenomena by exploiting redundancy in task vectors, as strategic parameter elimination reduces conflicts while preserving task-critical information. These approaches transform dense task vectors into sparse representations with improved composability.

The seminal Trim-Elect-Sign (TIES) mechanism introduced a three-stage pipeline that systematically resolves each category of interference. Given task vectors $\{\tau_i\}_{i=1}^n$ derived from $n$ fine-tuned models, the trimming stage retains only the top-$k$\% of parameters by magnitude, producing sparse vectors $\tilde{\tau}_i$ where all other entries are set to zero. The election stage then resolves sign conflicts through majority voting, computing a consensus sign mask $\gamma_j = \text{sign}(\sum_i \tilde{\tau}_{i,j})$ that determines the dominant direction for each parameter across all task vectors. Finally, the sign alignment stage retains only those parameter updates consistent with the elected sign, yielding conflict-free task vectors suitable for aggregation.

\begin{algorithm}[t]
\caption{TIES-Merging: Trim, Elect Sign, and Merge}
\label{alg:ties}
\begin{algorithmic}[1]
\REQUIRE Task vectors $\{\tau_1, \tau_2, \ldots, \tau_n\}$, trim ratio $k$, scaling coefficient $\lambda$, pretrained weights $\theta_{\text{pre}}$
\ENSURE Merged model parameters $\theta_{\text{merged}}$
\STATE \textbf{// Step 1: Trim - Keep only top-$k$\% parameters by magnitude}
\FOR{$i = 1$ to $n$}
 \STATE $\tilde{\tau}_i \leftarrow \text{TopK}(\tau_i, k)$ \COMMENT{Zero out small-magnitude parameters}
\ENDFOR
\STATE \textbf{// Step 2: Elect Sign - Resolve sign conflicts via majority voting}
\FOR{each parameter position $j$}
 \STATE $\gamma_j \leftarrow \text{sign}\left(\sum_{i=1}^{n} \tilde{\tau}_{i,j}\right)$ \COMMENT{Determine dominant sign}
\ENDFOR
\STATE \textbf{// Step 3: Merge - Average parameters with elected signs}
\FOR{each parameter position $j$}
 \STATE $S_j \leftarrow \{i : \text{sign}(\tilde{\tau}_{i,j}) = \gamma_j\}$ \COMMENT{Select agreeing task vectors}
 \STATE $\bar{\tau}_j \leftarrow \frac{1}{|S_j|}\sum_{i \in S_j} \tilde{\tau}_{i,j}$ \COMMENT{Average aligned values}
\ENDFOR
\STATE $\theta_{\text{merged}} \leftarrow \theta_{\text{pre}} + \lambda \bar{\tau}$
\RETURN $\theta_{\text{merged}}$
\end{algorithmic}
\end{algorithm}

\citet{DBLP:conf/nips/YadavTCRB23} showed that this mechanism achieves substantial performance improvements over naive task arithmetic, particularly when merging more than three task vectors where sign conflicts proliferate.

Complementing the deterministic TIES approach, the DARE (Drop And REscale) framework~\citep{DBLP:journals/corr/abs-2311-03099} introduced probabilistic sparsification as an alternative conflict resolution strategy. Rather than using magnitude-based pruning, DARE applies random dropout to task vector parameters with probability $p$, followed by rescaling of the retained parameters by factor $1/(1-p)$ to preserve the expected contribution. Mathematically, the sparsified task vector becomes $\hat{\tau}_i = \frac{1}{1-p} \cdot \tau_i \odot m_i$ where $m_i$ is a binary mask with $m_{i,j} \sim \text{Bernoulli}(1-p)$. This approach draws theoretical justification from the observation that fine-tuning induces primarily low-rank modifications to pretrained weights, similar to findings in network pruning research~\citep{DBLP:conf/iclr/FrankleC19}, implying considerable redundancy that can be exploited through random sampling without catastrophic information loss. Recent extensions such as DELLA-Merging~\citep{DBLP:journals/corr/abs-2406-11617} further refine this approach through magnitude-based sampling strategies. The stochastic nature of DARE provides an implicit ensemble effect when multiple merging trials are conducted, though practical deployments typically use a single sparsification pass.

Building upon these foundational methods, importance-aware sparsification has been proposed that uses second-order information to guide parameter retention decisions. Rather than treating all parameters equivalently during the trimming process, such methods compute importance scores based on the Fisher information or gradient magnitude accumulated during fine-tuning, preferentially retaining parameters whose modification contributed most directly to task adaptation. This approach overcomes a key limitation of magnitude-based pruning, wherein parameters with small absolute values may nonetheless encode critical task-specific information.

Subsequent work extended the sparsification approach by introducing conflict-awareness into the retention decision process. Unlike TIES, which considers sign conflicts only after independent trimming of each task vector, conflict-aware methods jointly optimize sparsification masks across all source models to minimize expected interference while maximizing information retention. CABS (Conflict-Aware Balanced Sparsification)~\citep{arxiv_2503_01874} introduces explicit conflict detection and balanced sparsification that normalizes contribution magnitudes across tasks, preventing dominant task vectors from overwhelming weaker contributors during the merge. Concrete Subspace Learning~\citep{arxiv_2312_06173} takes an alternative approach by learning continuous relaxations of discrete sparsification masks through concrete distributions, allowing gradient-based optimization of interference elimination. Sens-Merging~\citep{arxiv_2502_12420} proposes sensitivity-guided parameter balancing that uses gradient sensitivity analysis to determine which parameters most require conflict resolution.

Further refinements have incorporated distributional statistics of task vector parameters into sparsification decisions. DPPA~\citep{zhu2024dppa} introduces a dual-stage method combining dynamic magnitude-based pruning with partition-based amplification (rescaling) to better preserve important parameters while resolving conflicts during merging. ImPart~\citep{yang2025impart} proposes importance-aware delta-sparsification that jointly considers compression and merging objectives, striking improved trade-offs between model size and merged performance. Recognizing that optimal trimming thresholds vary across layers and parameter groups, these methods adaptively calibrate sparsification intensity based on local magnitude distributions, producing more precise conflict resolution than global threshold approaches. While these sparsification techniques have proven effective at mitigating interference, they nonetheless operate within the constraints of direct weight-space manipulation, motivating alternative formulations that reconsider the mathematical foundations of task vector composition itself.

\subsection{Advanced Task Vector Manipulation and Recent Extensions}
Motivated by these limitations of direct weight-space manipulation, recent methodological advances have extended the task arithmetic framework toward more principled mathematical reformulations that tackle core limitations of standard parameter arithmetic. A particularly influential direction involves reconceptualizing task vector operations within the tangent space of neural network parameters, motivated by the observation that standard weight-space arithmetic implicitly assumes local linearity that may not hold for large parameter displacements.

Research in this area introduced a partial linearization formulation for parameter-efficient adapters~\citep{DBLP:journals/corr/abs-2310-04742}, which linearizes only the adapter modules and applies task arithmetic over the linearized representations. Given a pretrained model with parameters $\theta_{\text{pre}}$ and a fine-tuned model $\theta_t$, the standard task vector $\tau_t = \theta_t - \theta_{\text{pre}}$ captures parameter displacements without accounting for the nonlinear relationship between weights and model behavior. The tangent space approach instead considers the linearized model $f_{\text{lin}}(x; \theta) = f(x; \theta_{\text{pre}}) + \nabla_\theta f(x; \theta_{\text{pre}})^\top (\theta - \theta_{\text{pre}})$, which provides a first-order approximation of how parameter changes affect model outputs. Within this framework, task vector arithmetic operates on the linearized representations, yielding improved compositional properties when combining multiple task-specific adaptations. Empirically, tangent space task arithmetic yields superior performance on model editing benchmarks, particularly for negation operations where the standard formulation often produces unstable results. Recent theoretical analysis by \citet{li2025taskvector} provide formal guarantees for when task vectors are provably effective for model editing in nonlinear Transformers, establishing generalization bounds that help practitioners predict merging success.

The theoretical appeal of tangent space methods stems from their explicit accommodation of neural network nonlinearity. The linearization constrains operations to a regime where the superposition principle approximately holds, handling scenarios where direct weight interpolation violates the implicit linearity assumptions of standard task arithmetic. This reformulation proves especially valuable for combining task vectors with large magnitudes or when source models have undergone extensive fine-tuning that moves parameters appreciably from the pretrained initialization.

Complementary to these mathematical reformulations, adaptive rank-based approaches optimize the representational structure of task vectors. Methods in this direction adaptively determine optimal rank configurations for low-rank approximations of task vectors across different layers and modules. STAR (Spectral Truncation and Rescale)~\citep{arxiv_2502_10339} uses spectral analysis to identify and preserve the most informative components of task vectors while truncating noise. LoRE-Merging~\citep{arxiv_2502_10749} explores low-rank estimation specifically designed for large language model merging, delivering effective compression of task vectors with minimal capability loss. No Task Left Behind~\citep{arxiv_2502_04959} proposes isotropic model merging that operates in common and task-specific subspaces to ensure that all constituent tasks are adequately represented. Superpose Task-specific Features~\citep{arxiv_2502_10698} introduces a feature superposition framework that supports more effective combination of task-specific information. Unlike uniform rank reduction strategies, these approaches recognize that different network components encode task-specific information with varying degrees of redundancy, allocating representational capacity accordingly. The method formulates rank selection as an optimization problem that balances compression efficiency against information preservation, using singular value decomposition to identify the effective dimensionality of task-specific modifications.

Integrating adaptive rank methods with existing sparsification techniques yields synergistic benefits. While sparsification operates by zeroing individual parameters, rank-based approaches restructure the entire task vector representation, capturing different aspects of redundancy in fine-tuned models. Layer-Aware Task Arithmetic~\citep{chen2025layeraware} disentangles task-specific knowledge from instruction-following capabilities at the layer level, allowing more targeted composition. Adaptive Task Vectors~\citep{kang2025adaptive} dynamically generate input-conditioned task vectors using a small language model, enabling query-specific adaptation without fixed demonstration sets. Decom-Renorm-Merge~\citep{chaichana2025decomrenormmerge} proposes performing merging in a decomposed and renormalized space, showing that operating in the right mathematical space markedly improves multi-tasking performance. Optimal Brain Iterative Merging~\citep{wang2025optimal} draws inspiration from second-order pruning methods to iteratively mitigate interference during the merging process. Table~\ref{tab:task_vector_arithmetic_and_spa} summarizes these methods. This complementarity enables hybrid strategies that first reduce task vector rank to eliminate correlated modifications, then apply sparsification to resolve residual conflicts among the compressed representations.

From a practical standpoint, community toolkits such as MergeKit~\citep{DBLP:journals/corr/abs-2403-13257} have progressively incorporated these advanced manipulation methods, offering modular pipelines that compose tangent space projections, adaptive rank reduction, and conflict-aware sparsification into reproducible merging workflows. Despite their increasing complexity, however, all task vector manipulation techniques discussed thus far share a common assumption, namely that merging should produce a unified parameter set. An alternative approach instead preserves distinct expert components while allowing flexible, input-dependent use of specialized capabilities.

\begin{table}[t]
 \centering
 \footnotesize
 \caption{Task vector arithmetic and sparsification-enhanced methods for model merging. Each method addresses a specific aspect of parameter interference: sign conflicts (TIES-Merging), redundant updates (DARE), conflict-aware balancing (CABS), or tangent space reformulation. The ``Interference Handling'' column indicates whether the method explicitly resolves parameter conflicts between task vectors.}
 \label{tab:task_vector_arithmetic_and_spa}
 \resizebox{\textwidth}{!}{
 \begin{tabular}{p{4.5cm}p{3cm}p{4cm}p{3.5cm}c}
 \toprule
 \textbf{Method} & \textbf{Key Approach} & \textbf{Key Feature(s)} & \textbf{Limitations/Notes} & \textbf{Year} \\
 \midrule
  \textbf{Task Arithmetic} \newline {\scriptsize \citep{DBLP:conf/iclr/IlharcoRWSHF23}} & Linear task vector combination & Defines task vectors as weight differences; enables addition, negation, scaling operations & Suffers from parameter interference and sign conflicts & 2023 \\
 \addlinespace
  \textbf{TIES-Merging} \newline {\scriptsize \citep{DBLP:conf/nips/YadavTCRB23}} & Trim, Elect Sign, Merge & Resolves sign conflicts via magnitude-based trimming and democratic sign election & Requires hyperparameter tuning for sparsity threshold & 2023 \\
 \addlinespace
  \textbf{DARE} & Drop and Rescale & Randomly drops delta parameters and rescales remaining ones to preserve expectations & Stochastic nature may introduce variance & 2023 \\
 \addlinespace
  \textbf{CABS} & Conflict-Aware Balanced Sparsification & Explicitly identifies and eliminates conflicting parameters while maintaining balance & Limited to identical architectures & 2025 \\
 \addlinespace
  \textbf{Tangent Space Methods} & Tangent Space Operations & Reformulates task arithmetic in tangent space for improved linearity assumptions & Increased mathematical complexity & 2024 \\
 \bottomrule
 \end{tabular}
 }
\end{table}

Task vector arithmetic provides a powerful conceptual basis for model editing, supporting intuitive operations such as knowledge addition and negation, while sparsification-enhanced methods like TIES-Merging and DARE tackle critical challenges of parameter interference and sign conflicts by selectively pruning redundant or conflicting parameters before aggregation. Beyond these core methods, recent extensions reveal several converging trends that illuminate the field's trajectory.

A first trend is the refinement of \textit{sparsity and pruning criteria}. While TIES-Merging and DARE establish that sparsification improves merging, follow-up work shows that the choice of \emph{which} parameters to prune matters as much as the pruning rate. LEWIS~\citep{arxiv_2503_03874} proposes layer-wise sparsity for training-free guided merging, recognizing that different layers require different pruning intensities, while AdaRank~\citep{arxiv_2503_22178} replaces element-wise sparsity with adaptive rank pruning that operates on the singular value structure of task vectors. This progression from uniform to adaptive sparsification mirrors the broader shift observed in weight-space averaging methods.

A second, closely related trend moves conflict resolution from \textit{parameter space to representation space}. Representation Surgery~\citep{DBLP:journals/corr/abs-2402-02705} mitigates multi-task interference directly at the representation level, Subspace-Boosted Merging~\citep{arxiv_2506_16506} exploits subspace decomposition, and Purifying Task Vectors~\citep{arxiv_2510_14697} operates in knowledge-aware subspaces. The key insight is that parameter-level sign conflicts may be symptoms of deeper representational incompatibilities; by resolving interference in the representation domain, these methods can address root causes rather than surface-level symptoms.

Third, several methods introduce explicit \textit{stability constraints} to prevent catastrophic forgetting during merging. Task Arithmetic in Trust Region~\citep{arxiv_2501_15065} constrains merged parameters to remain within trust regions around the pretrained initialization, LINES~\citep{arxiv_2410_17146} prevents forgetting via post-training layer scaling, and Adaptive Weight Disentanglement~\citep{arxiv_2411_18729} separates task-specific from shared components. These approaches trade merging flexibility for stronger preservation guarantees, a tradeoff that becomes increasingly important as the number of merged tasks grows.

Fourth, \textit{differentiable and optimization-based formulations} recast merging as a continuous optimization problem rather than a discrete recipe. DAM~\citep{arxiv_2410_08371} provides a differentiable path from simple averaging to full model composition, while adaptive projective gradient descent~\citep{arxiv_2501_01230} enables gradient-based coefficient learning. By making the merging process end-to-end differentiable, these methods can jointly optimize merging coefficients, but at the cost of requiring validation data, a departure from the data-free approach that characterizes most task arithmetic methods.

Fifth, \textit{cross-model transfer} methods relax the strict shared-initialization requirement. Update Your Transformer~\citep{arxiv_2505_22697} supports re-basin of task vectors across pretrained model versions, Gradient-Sign Masking~\citep{arxiv_2510_09658} enables task vector transport, and Decomposing Task Vectors~\citep{arxiv_2512_22511} supports refined model editing across model families. These methods represent an important step toward broader applicability, though they typically incur additional alignment costs compared to same-initialization merging.

Finally, \textit{novel applications} show the versatility of the task vector abstraction beyond multi-task combination. Personality Vector~\citep{arxiv_2509_19727} extends task arithmetic to modulate personality traits of LLMs through continuous-strength control and multi-trait composition. Expert Merging~\citep{arxiv_2509_25712} proposes unsupervised expert alignment via hidden-state and logit matching with importance-guided layer chunking. Efficient Compositional Multi-tasking~\citep{arxiv_2507_16083} introduces learnable calibration for merging task adapters when a single input must be processed for multiple tasks simultaneously on edge devices. Additional contributions include SE-Merging~\citep{arxiv_2506_18135}, MetaGPT~\citep{arxiv_2406_11385}, MoD~\citep{arxiv_2411_00406}, distribution-aware sparse fusion~\citep{arxiv_2602_11717}, and ParamDelta~\citep{arxiv_2504_21023}.
While this section focused on parameter-space manipulations and sparsity-driven conflict resolution, the next section examines structured and information-guided merging approaches that move beyond element-wise operations to exploit architectural organization through Mixture-of-Experts~\citep{DBLP:conf/iclr/ShazeerMMDLHD17} compositions, use activation statistics for guided merging decisions, and apply evolutionary search to automatically discover optimal merging configurations across the full design space.

\section{Structured and Information-Guided Merging Approaches}
\label{sec:search}
While task vector arithmetic and sparsification techniques offer powerful tools for mitigating parameter interference, they primarily operate on static weight representations without considering the underlying model architecture or runtime behavior. We now turn to a complementary class of approaches that exploit structural priors and dynamic information to guide the merging process more effectively. This section examines three advanced families of methods. These include architecture-level composition methods that preserve modularity through Mixture-of-Experts (MoE) style routing mechanisms, activation-informed techniques that exploit representation statistics for principled parameter alignment, and evolutionary or search-based meta-strategies that automatically discover optimal merging configurations across the design space.

\begin{table*}[t]
\caption{Unified comparison of representative model merging methods. \textbf{Data-Free}: no training/calibration data required. \textbf{Interference}: explicit handling of parameter conflicts. \textbf{Shared Init}: requires common pretrained initialization.}
\label{tab:unified-comparison}
\centering
\resizebox{\textwidth}{!}{
\begin{tabular}{p{4.5cm}lccccc}
\toprule
\textbf{Method} & \textbf{Category} & \textbf{Data-Free} & \textbf{Interference Handling} & \textbf{Shared Init} & \textbf{Scalability} & \textbf{Year} \\
\midrule
\multicolumn{7}{l}{\textit{Weight-Space Averaging (Section~3)}} \\
\midrule
\textbf{Simple Averaging} & Static & \cmark & \xmark & \cmark & High & -- \\
\textbf{Model Soups} \newline {\scriptsize \citep{DBLP:conf/icml/WortsmanIGRLMNF22}} & Greedy selection & \cmark & \xmark & \cmark & High & 2022 \\
\textbf{Fisher Merging} \newline {\scriptsize \citep{DBLP:conf/nips/MatenaR22}} & Importance-weighted & \xmark & Partial & \cmark & Medium & 2022 \\
\textbf{RegMean} \newline {\scriptsize \citep{DBLP:conf/iclr/Jin0P023}} & Linear regression & \xmark & Partial & \cmark & Medium & 2023 \\
\textbf{SLERP} & Geometric & \cmark & \xmark & \cmark & High & 2023 \\
\midrule
\multicolumn{7}{l}{\textit{Task Vector \& Sparsification (Section~4)}} \\
\midrule
\textbf{Task Arithmetic} \newline {\scriptsize \citep{DBLP:conf/iclr/IlharcoRWSHF23}} & Additive & \cmark & \xmark & \cmark & High & 2023 \\
\textbf{TIES-Merging} \newline {\scriptsize \citep{DBLP:conf/nips/YadavTCRB23}} & Sign resolution & \cmark & \cmark & \cmark & High & 2023 \\
\textbf{DARE} \newline {\scriptsize \citep{DBLP:journals/corr/abs-2311-03099}} & Sparsification & \cmark & \cmark & \cmark & High & 2023 \\
\textbf{Model Breadcrumbs} \newline {\scriptsize \citep{DBLP:journals/corr/abs-2312-06795}} & Sparse masking & \cmark & \cmark & \cmark & High & 2024 \\
\textbf{DELLA} \newline {\scriptsize \citep{DBLP:journals/corr/abs-2406-11617}} & Magnitude sampling & \cmark & \cmark & \cmark & High & 2024 \\
\textbf{Model Stock} \newline {\scriptsize \citep{DBLP:journals/corr/abs-2403-19522}} & Geometric selection & \cmark & Implicit & \cmark & High & 2024 \\
\textbf{AdaMerging} \newline {\scriptsize \citep{DBLP:conf/iclr/YangWLWGLLQYCL24}} & Adaptive coefficients & \xmark & \cmark & \cmark & Medium & 2024 \\
\midrule
\multicolumn{7}{l}{\textit{Structured \& Routing (Section~5)}} \\
\midrule
\textbf{Git Re-Basin} \newline {\scriptsize \citep{DBLP:conf/iclr/AinsworthHS23}} & Permutation alignment & \xmark & N/A & \xmark & Low & 2023 \\
\textbf{ZipIt!} \newline {\scriptsize \citep{DBLP:conf/iclr/StoicaBHH24}} & Feature matching & \xmark & N/A & \xmark & Low & 2024 \\
\textbf{Branch-Train-Merge} \newline {\scriptsize \citep{DBLP:conf/icml/LiDTG0H22}} & MoE routing & \xmark & N/A & \cmark & High & 2022 \\
\textbf{PHATGOOSE} \newline {\scriptsize \citep{DBLP:journals/corr/abs-2402-05859}} & Learned routing & \xmark & N/A & \cmark & Medium & 2024 \\
\textbf{Evol.\ Merging} \newline {\scriptsize \citep{DBLP:journals/corr/abs-2403-13187}} & Search-based & \xmark & Implicit & \cmark & Medium & 2024 \\
\midrule
\multicolumn{7}{l}{\textit{LoRA~\citep{DBLP:conf/iclr/HuSWALWWC22}-Specific}} \\
\midrule
\textbf{LoRAHub} \newline {\scriptsize \citep{DBLP:journals/corr/abs-2307-13269}} & Gradient-free composition & \xmark & Partial & \cmark & High & 2023 \\
\textbf{MoLE} \newline {\scriptsize \citep{DBLP:journals/corr/abs-2404-13628}} & Expert routing & \xmark & N/A & \cmark & High & 2024 \\
\bottomrule
\end{tabular}
}
\end{table*}

\subsection{Mixture-of-Experts Style Merging and Expert Routing Strategies}
Mixture-of-Experts (MoE) style merging preserves specialized capabilities by organizing distinct expert pathways, directly countering destructive interference when combining divergent models~\citep{DBLP:conf/icml/LiDTG0H22,DBLP:journals/corr/abs-2306-03745}. By maintaining separate parameter spaces for each expert and using learned routing mechanisms, MoE-style merging strikes a principled balance between capability preservation and unified deployment.

The Branch-Train-Merge (BTM) framework provides the conceptual foundation for scalable expert training within this approach~\citep{DBLP:conf/icml/LiDTG0H22}. In BTM, a shared pretrained LLM is first replicated into multiple branches, each subsequently fine-tuned independently on distinct data subsets or task distributions. The original BTM framework combines expert predictions through simple ensemble averaging at inference time. Subsequent work extended this paradigm by introducing learned routing mechanisms that direct inputs to appropriate experts, yielding a general MoE-style formulation. Formally, given a set of $K$ expert models $\{E_1, E_2, \ldots, E_K\}$ derived from a common pretrained initialization, the merged output for an input $x$ is computed as:
\begin{equation}
y = \sum_{i=1}^{K} g_i(x) \cdot E_i(x),
\label{eq:moe_routing}
\end{equation}

where $g_i(x)$ represents the routing weight assigned to expert $i$ by a gating network $G$, typically computed via a softmax transformation over learned routing logits:
\begin{equation}
g_i(x) = \frac{\exp(w_i^\top h(x))}{\sum_{j=1}^{K} \exp(w_j^\top h(x))},
\label{eq:softmax_gate}
\end{equation}
with $h(x)$ denoting an intermediate hidden representation and $\{w_1, \ldots, w_K\}$ the learnable routing parameters. To promote computational efficiency, top-$k$ sparse routing activates only a subset of experts per input:
\begin{equation}
g_i^{\text{top-}k}(x) = \begin{cases} g_i(x) & \text{if } i \in \text{TopK}(\{g_j(x)\}_{j=1}^K) \\ 0 & \text{otherwise} \end{cases},
\label{eq:topk_routing}
\end{equation}
enabling inference costs that scale as $O(k/K)$ relative to the full ensemble. CALM (Composition to Augment Language Models)~\citep{DBLP:journals/corr/abs-2401-02412} shows that LLMs can be augmented through composition with specialized modules, allowing capability expansion without full model retraining. The WEMoE framework~\citep{DBLP:journals/corr/abs-2402-00433} shows how this weight-ensembling approach can effectively integrate multiple task-specific models into a unified MoE architecture, achieving superior multi-task performance compared to naive averaging strategies by constructing experts from merged weight differences rather than full model copies.

Several prominent methods have been proposed for constructing MoE systems from independently trained models. PHATGOOSE (Post-Hoc Adaptive Tokenwise Gating Over an Ocean of Specialized Experts)~\citep{DBLP:journals/corr/abs-2402-05859} addresses the challenge of routing over pre-existing fine-tuned models by learning lightweight gating parameters while keeping expert weights frozen. For each expert $E_i$, PHATGOOSE computes routing scores based on the dot product between a learned gate vector $v_i$ and the input representation: $g_i(x) = v_i^\top h(x)$, where gate vectors are optimized on each expert's original training data to identify inputs the expert handles well. MoLE (Mixture of LoRA Experts)~\citep{DBLP:journals/corr/abs-2404-13628} takes a parameter-efficient approach by treating individual LoRA adapters as experts, composing the output as $y = W_0 x + \sum_{i=1}^{K} g_i(x) \cdot B_i A_i x$, where $W_0$ represents the frozen pretrained weights and $B_i A_i$ denotes the low-rank decomposition of expert $i$. This LoRA-as-experts design greatly reduces memory overhead since each expert contributes only the low-rank adapter parameters rather than full model weights. LoRAHub~\citep{DBLP:journals/corr/abs-2307-13269} enables efficient cross-task generalization via dynamic LoRA composition, while LoRAMoE~\citep{DBLP:journals/corr/abs-2312-09979} extends this concept by using LoRA-based experts as plugins that preserve world knowledge during instruction tuning.

Beyond the basic softmax gating (Eq.~\ref{eq:softmax_gate}) and top-$k$ routing (Eq.~\ref{eq:topk_routing}) introduced above, several design choices critically determine MoE effectiveness. Load balancing mechanisms prevent expert collapse, where routing degenerates to selecting only a few experts. The auxiliary load balancing loss $\mathcal{L}_{\text{balance}} = \alpha \cdot K \cdot \sum_{i=1}^{K} f_i \cdot p_i$ penalizes uneven expert usage, where $f_i$ represents the fraction of tokens routed to expert $i$ and $p_i$ denotes the average routing probability for expert $i$ across the batch~\citep{DBLP:journals/corr/abs-2401-04088,DBLP:journals/jmlr/FedusZS22}. Switch Transformers~\citep{DBLP:journals/jmlr/FedusZS22} scaled to trillion parameter models with simple and efficient sparsity mechanisms. Branch-Train-MiX~\citep{DBLP:journals/corr/abs-2403-07816} provides a framework for mixing expert LLMs into a unified MoE architecture through branched training procedures. Alternative routing strategies beyond learned softmax gating include expert-choice routing, which inverts the selection process by having experts select their top-$k$ preferred tokens rather than tokens selecting experts, naturally achieving load balance. The granularity of routing decisions, whether at token-level for fine-grained adaptation or sequence-level for reduced overhead, depends considerably on expert specialization patterns and input diversity~\citep{DBLP:journals/corr/abs-2408-07057}.

Post-hoc integration of specialized fine-tuned models as experts presents unique challenges distinct from training MoE systems from scratch, as experts were never optimized for selective activation. PHATGOOSE tackles this through a two-stage process, first computing per-expert gate vectors by optimizing routing scores on each expert's training data independently, then combining these gates for joint inference without requiring access to all training data simultaneously. Merge, Then Compress~\citep{arxiv_2310_01334} demystifies efficient sparse MoE by first merging redundant experts and then compressing the resulting architecture, providing insights into the relationship between expert similarity and routing behavior. LoRA-Flow~\citep{arxiv_2402_11455} introduces dynamic LoRA fusion that adaptively combines LoRA adapters during generation based on input characteristics, allowing more fine-grained expert use than static routing. LoraRetriever~\citep{arxiv_2402_09997} proposes input-aware LoRA retrieval and composition for mixed tasks, dynamically selecting and combining relevant adapters from a library. XFT~\citep{ding2024xft} unlocks the power of code instruction tuning by merging upcycled mixture-of-experts, showing that MoE structures can emerge from strategic combination of specialized models. Model Kinship~\citep{DBLP:journals/corr/abs-2410-12613} provides a framework for exploring and quantifying the similarity relationships between LLMs that predict merging performance, supporting principled selection of candidate models for merging through a kinship-based greedy strategy.

MoE-style merging incurs distinct resource tradeoffs compared to parameter-averaging approaches. While parameter averaging produces a single model with memory requirements identical to any constituent, MoE architectures must store all $K$ expert parameter sets, resulting in memory scaling of $O(K \cdot |E|)$ where $|E|$ denotes expert size. LoRA-based experts greatly mitigate this overhead~\citep{DBLP:conf/iclr/HuSWALWWC22}, and SMILE~\citep{tang2024smile} further reduces costs by constructing sparse mixture of low-rank experts from pretrained models in a zero-shot manner. Computationally, sparse top-$k$ routing reduces inference FLOPs to approximately $k/K$ of the dense ensemble cost but introduces routing overhead and memory bandwidth costs from loading selected expert weights. In contrast, merged models via TIES-Merging~\citep{DBLP:conf/nips/YadavTCRB23} or DARE~\citep{DBLP:journals/corr/abs-2311-03099} incur no additional inference cost beyond a single model but cannot dynamically adapt computation to input characteristics.

Several recent methods further improve LoRA-based MoE composition. LoRA Soups~\citep{prabhakar2024lora} show that model merging outperforms data mixing for skill composition, with their concatenation-of-LoRAs (CAT) method achieving optimal weighting of individually trained modules. MergeME~\citep{zhou2025mergeme} extends merging techniques to both homogeneous and heterogeneous MoE architectures for growing MoE-based LLM deployments. DLP-LoRA~\citep{zhang2024dlplora} introduces dynamic lightweight plugins for efficient task-specific fusion, while Decouple and Orthogonalize~\citep{zheng2025decouple} proposes data-free LoRA merging through orthogonal subspace decomposition, and \citet{zhang2025unraveling} analyze LoRA interference patterns and propose orthogonal subspace methods for robust merging. The choice between MoE and parameter-averaging approaches thus depends on deployment constraints. Specifically, MoE architectures favor scenarios requiring preserved specialist capabilities and sufficient memory, while parameter averaging suits resource-constrained settings where moderate capability blending suffices. Table~\ref{tab:moe_lora_comparison} compares representative MoE-style and LoRA-based composition methods.

\begin{table}[t]
 \centering
 \footnotesize
 \caption{Comparison of MoE-style and LoRA-based composition methods for model merging. Memory overhead is relative to base model size.}
 \label{tab:moe_lora_comparison}
 \resizebox{\textwidth}{!}{
 \begin{tabular}{p{4.5cm}lcccp{3.5cm}}
 \toprule
 \textbf{Method} & \textbf{Expert Type} & \textbf{Memory} & \textbf{Router Training} & \textbf{Year} & \textbf{Key Innovation} \\
 \midrule
 \multicolumn{6}{l}{\textit{Full-Model Expert Methods}} \\
 \midrule
  \textbf{Switch Transformers} \newline {\scriptsize \citep{DBLP:journals/jmlr/FedusZS22}} & Dense FFN & $K\times$ FFN & From scratch & 2022 & Simplified MoE with single expert routing \\
  \textbf{Branch-Train-MiX} \newline {\scriptsize \citep{DBLP:journals/corr/abs-2403-07816}} & Full model & $K\times$ & Post-hoc & 2024 & Branched training for expert specialization \\
  \textbf{PHATGOOSE} \newline {\scriptsize \citep{DBLP:journals/corr/abs-2402-05859}} & Full model & $K\times$ & Per-expert & 2024 & Post-hoc gating without joint training \\
 \midrule
 \multicolumn{6}{l}{\textit{LoRA-Based Expert Methods}} \\
 \midrule
  \textbf{LoRAHub} \newline {\scriptsize \citep{DBLP:journals/corr/abs-2307-13269}} & LoRA adapter & $K\times$ LoRA & Few-shot & 2023 & Dynamic composition via gradient-free search \\
  \textbf{MoLE} \newline {\scriptsize \citep{DBLP:journals/corr/abs-2404-13628}} & LoRA adapter & $K\times$ LoRA & Learned gating & 2024 & Mixture-of-LoRA-Experts routing \\
  \textbf{LoRAMoE} \newline {\scriptsize \citep{DBLP:journals/corr/abs-2312-09979}} & LoRA adapter & $K\times$ LoRA & Balanced routing & 2023 & Preserves world knowledge via localized balancing \\
 \midrule
 \multicolumn{6}{l}{\textit{Hybrid and Advanced Methods}} \\
 \midrule
  \textbf{WEMoE} \newline {\scriptsize \citep{DBLP:journals/corr/abs-2402-00433}} & Weight diff & $K\times$ delta & Task-aware & 2024 & Weight-ensembling from task vectors \\
  \textbf{Twin-Merging} \newline {\scriptsize \citep{DBLP:journals/corr/abs-2406-15479}} & Modular & Dynamic & Adaptive & 2024 & Dynamic integration of modular expertise \\
  \textbf{CALM} \newline {\scriptsize \citep{DBLP:journals/corr/abs-2401-02412}} & Cross-attn & Anchor + Aug & Composed & 2024 & LLM augmentation through composition \\
 \bottomrule
 \end{tabular}
 }
\end{table}

Beyond these structural approaches to combining models, an orthogonal direction exploits empirical observations of how models process inputs to guide the merging process itself.

\paragraph{Limitations of MoE-style merging.} Despite the architectural elegance, MoE-style merging introduces several practical challenges. The memory footprint scales linearly with the number of experts, as each expert retains a full parameter copy (or full LoRA adapter), making deployment of $K$-expert systems $K$-times more expensive than a single merged model. Router training requires representative held-out data for each task, and poorly calibrated routers may exhibit \emph{expert collapse}, where the router directs all inputs to a single dominant expert, particularly when task distributions are imbalanced. \citet{DBLP:conf/icml/LiDTG0H22} showed in the Branch-Train-Merge framework that expert domain specialization is essential for effective merging, as random data splits fail to produce well-performing merged models. Federated variants further exacerbate these issues, since heterogeneous client data distributions can cause routers to overfit local patterns~\citep{DBLP:conf/mlsys/LiSZSTS20}.

\subsection{Activation-Informed and Representation-Guided Merging Methods}
Activation-informed and representation-guided merging methods embody this empirical approach, representing a departure from static weight-averaging by incorporating direct observations of model behavior into the merging process. While methods such as Fisher-weighted averaging use second-order gradient information computed at parameter positions, activation-based approaches instead examine how models actually process inputs through their forward computational pathways, thereby capturing functional importance that may not be evident from gradient-based measures alone.

A key insight motivating these approaches is that parameter importance for a given task manifests directly in activation patterns during inference. When a model processes task-relevant inputs, certain neurons and attention heads exhibit consistently high activation magnitudes or frequencies, indicating their centrality to the computational process. Conversely, components with sparse or negligible activations contribute minimally to output formation and thus warrant reduced influence during merging. Formally, given a calibration dataset $\mathcal{D}$ representative of target tasks, activation-informed methods compute importance scores $\alpha_i$ for parameter groups by aggregating statistics over forward passes: $\alpha_i = f(\{a_i(x) : x \in \mathcal{D}\})$, where $a_i(x)$ denotes the activation vector associated with parameter group $i$ when processing input $x$, and $f$ is an aggregation function such as mean absolute activation or activation variance.

Representation-level alignment extends beyond scalar importance scores to account for the geometric structure of intermediate representations across models. Given two models $\theta_A$ and $\theta_B$ fine-tuned from a common pretrained initialization, their intermediate representations may occupy differently oriented subspaces despite encoding semantically similar information. Centered Kernel Alignment (CKA) and related similarity metrics can quantify representational correspondence across layers, providing guidance for matching functionally equivalent components before parameter combination. This alignment proves particularly valuable when models have undergone divergent fine-tuning trajectories that disrupt the implicit correspondence established by shared initialization.

Semantic-guided merging approaches exploit the rich structure of intermediate representations to inform merging decisions at a finer granularity than global importance scores permit. Multimodal model development has increasingly adopted MoE-style architectures, with MoE-LLaVA~\citep{lin2024moellava} showing that sparse MoE training strategies can effectively scale large vision-language models while maintaining computational efficiency. More broadly, merging techniques applied to multimodal models must account for cross-modal alignment, as parameter combination may disrupt the correspondence between vision and language representations established during joint training. These methods typically construct auxiliary objectives that encourage merged representations to maintain task-relevant semantic properties observed in constituent models, effectively treating representation preservation as a soft constraint during the merging optimization.

\citet{DBLP:journals/corr/abs-2408-07666} provide a taxonomy of activation-informed techniques, categorizing them by information source, ranging from lightweight activation magnitudes to detailed attention pattern analysis. Activated Parameter Locating~\citep{kong2024activated} introduces causal intervention to identify task-relevant parameters, providing a principled mechanism for determining which parameters should be prioritized during merging. Activation-Guided Consensus Merging~\citep{yao2025activationguided} extends this direction by computing consensus across multiple models' activation patterns to guide parameter combination decisions. Curvature-informed merging~\citep{mahdavinia2025harnessing} harnesses optimization dynamics to incorporate second-order curvature information into the merging process, providing a bridge between activation-based and gradient-based importance estimation. These methods achieve their advantages through task-relevant calibration data, raising questions about data requirements and calibration set composition.

Computational overhead of activation-informed merging varies substantially across methods. Simple activation magnitude averaging requires only forward passes through calibration data, adding modest preprocessing cost. However, more involved representation alignment techniques may require solving optimization problems over representation matching objectives, considerably increasing computational requirements. The optimal tradeoff between computational investment and merging quality improvement remains context-dependent~\citep{DBLP:journals/corr/abs-2502-02421}.

Empirical evidence suggests activation-informed methods show the greatest advantages when merging models with heterogeneous fine-tuning histories or when constituent models exhibit substantial parameter interference under naive averaging. In scenarios where linear mode connectivity~\citep{DBLP:conf/icml/FrankleDBMG20} holds strongly, typically when models share recent fine-tuning from identical pretrained checkpoints, the marginal benefit of activation guidance diminishes relative to simpler static methods. This aligns with the understanding that shared initialization establishes common loss basins within which parameter averaging succeeds without explicit importance weighting. Consequently, practitioners should consider activation-informed approaches primarily when merging distantly related models or when calibration data clearly reveals differential parameter usage across tasks. Yet even with principled importance weighting, the broader challenge remains. Given a collection of candidate models, how should one select which models to merge, determine layer-wise mixing coefficients, and configure method-specific hyperparameters? This combinatorial complexity motivates automated optimization strategies that can systematically explore the space of possible merge configurations.

\subsection{Evolutionary and Search-Based Optimization for Merging Configurations}

Since principled importance weighting alone cannot resolve the broader challenge of configuration selection, evolutionary and search-based approaches reformulate merging as an optimization problem over a structured configuration space. These methods open the door to principled exploration of merge recipes that would be infeasible to discover through exhaustive enumeration or expert intuition alone. \citet{DBLP:conf/kdd/AkibaSTN19} provide hyperparameter optimization tools commonly used in this setting. The search space for model merging can be formally characterized as a product of discrete and continuous decision variables. Let $\mathcal{M} = \{M_1, \ldots, M_K\}$ denote a collection of candidate source models. A merge recipe $\theta$ typically consists of a selected model subset $S \subseteq \mathcal{M}$, per-layer mixing coefficients $\alpha^{(l)} \in \Delta^{|S|-1}$ (where $l$ ranges from 1 to 32 or more, depending on model depth), and topological decisions regarding layer permutation or repetition. The optimization objective is defined as $\theta^* = \arg\max_\theta \mathcal{F}(\text{Merge}(\theta))$, where $\mathcal{F}$ represents a fitness function derived from specific validation metrics. In practice, $\mathcal{F}$ is often a composite score from the Open LLM Leaderboard or task-specific benchmarks such as GSM8K for reasoning and HumanEval for coding. This formulation reveals the immense scale of the challenge. A merge of two 70-billion parameter models involves optimizing hundreds of continuous coefficients alongside discrete architectural choices, creating a non-convex landscape rife with local optima.

Evolutionary algorithms (EAs) are the dominant framework for navigating this space, most notably shown by the landmark work of~\citet{DBLP:journals/corr/abs-2403-13187}. This approach generalizes model merging into two distinct optimization domains, namely Parameter Space (PS), which optimizes mixing weights (e.g., via SLERP or Task Arithmetic), and Data Flow Space (DFS), which optimizes the inference path by permuting or repeating layers from different models. By applying CMA-ES (Covariance Matrix Adaptation Evolution Strategy) to optimize continuous mixing weights in PS and evolutionary search over discrete layer permutations in DFS, the authors successfully merged a Japanese LLM with a Math-specialized LLM to create a model that achieved top performance on Japanese math reasoning, a capability present in neither parent model individually. CMA-ES adaptively learns the covariance structure of high-fitness solutions, concentrating the search in promising regions of the coefficient space and allowing the algorithm to discover non-intuitive architectures, such as the ``Franken-merge'' structures where layers from different models are interleaved in novel sequences. EvoMerge~\citep{jiang2024evomerge} extends neuroevolution principles specifically for LLM merging, while CycleQD~\citep{kuroki2024agent} shows that quality-diversity algorithms can discover diverse merging configurations for agent skill acquisition. Mergenetic~\citep{minut2025mergenetic} provides a simple and accessible evolutionary model merging library that democratizes these techniques for practitioners.

While~\citet{DBLP:journals/corr/abs-2403-13187} demonstrated the effectiveness of CMA-ES for the continuous subspace of mixing coefficients, the computational cost of fitness evaluation remains a bottleneck; typical evolutionary runs reported in the literature may require hundreds of candidate evaluations, consuming significant GPU hours. To mitigate this, practitioners often use lower-fidelity proxies during the search phase, such as evaluating on small subsets of the validation data (e.g., minibatches of perplexity scores) before verifying top candidates on full benchmarks.

As an alternative to evolutionary methods, Bayesian Optimization (BO) offers a more sample-efficient strategy for finding optimal merge parameters. BO constructs a probabilistic surrogate model (typically a Gaussian Process) to approximate the fitness function $\mathcal{F}$. By using an acquisition function to balance exploration (sampling uncertain regions) and exploitation (sampling high-performing regions), BO can converge to optimal mixing coefficients with far fewer function evaluations than genetic algorithms. \citet{jang2024bayesian} show that Bayesian optimization produces competitive merging quality with markedly reduced evaluation budgets compared to evolutionary approaches, particularly when the search space is restricted to continuous mixing coefficients. SIP-BMM~\citep{chen2025sipbmm} extends this direction by constructing capability-efficiency Pareto sets through Bayesian model merging with structural importance priors. However, standard BO struggles with the high-dimensional, mixed-discrete nature of Data Flow Space optimization, making it more suitable for refining mixing coefficients ($\alpha$) once the architectural topology is fixed.

Multi-objective formulations are essential when merged models must satisfy competing criteria, such as balancing safety alignment with reasoning capability. Pareto-based evolutionary algorithms, such as NSGA-II, maintain a population approximating the Pareto frontier, allowing practitioners to select trade-off solutions post-hoc. Such approaches are critical for "generalist" merging, where a single scalar fitness function fails to capture the degradation in instruction following that often accompanies gains in domain-specific knowledge~\citep{DBLP:journals/corr/abs-2311-13534}.

Table~\ref{tab:search_paradigms} summarizes the trade-offs between the primary search paradigms used in automated model merging.

\begin{table}[t]
\centering
\caption{Search paradigms for automated model merging configuration discovery. Evolutionary algorithms excel at exploring discrete architectural choices (e.g., layer permutation) but require many fitness evaluations; Bayesian optimization is more sample-efficient for continuous coefficient tuning; random search serves as a scalable baseline. ``Variable Handling'' indicates support for mixed discrete-continuous search spaces.}
\label{tab:search_paradigms}
\footnotesize
\resizebox{\columnwidth}{!}{
\begin{tabular}{lllll}
\toprule
\textbf{Search Paradigm} & \textbf{Sample Efficiency} & \textbf{Parallelizability} & \textbf{Variable Handling} & \textbf{Best Use Case} \\
\midrule
Evolutionary Algorithms & Low & High & Excellent & Novel layer architectures \\
Bayesian Optimization & High & Low & Limited & Fine-tuning coefficients \\
Random Search & Very Low & Very High & Good & Baseline benchmarking \\
\bottomrule
\end{tabular}
}
\end{table}

Search-based model merging has shifted the field from manual heuristic design to automated discovery. While evolutionary methods currently lead in discovering complex architectural combinations, the high computational cost of evaluation remains a barrier. Future research will likely converge on hybrid approaches that combine the structural exploration of evolution with the sample efficiency of Bayesian methods or surrogate-assisted optimization.

\paragraph{Benchmarking evolutionary merging.} Quantitative comparisons on the FusionBench suite~\citep{tang2024fusionbench} reveal that CMA-ES-based search over layer-wise coefficients typically recovers a large fraction of the best oracle merge performance at much lower evaluation cost compared to exhaustive grid search, though the absolute gap varies with model scale. A multi-fidelity framework~\citep{arxiv_2502_04030} for automated model merging further reduces search costs by hierarchically evaluating candidates at increasing fidelity levels. MERGE3~\citep{arxiv_2502_10436} makes evolutionary merging feasible on consumer-grade GPUs, showing that search-based approaches need not be restricted to resource-rich settings. Nature-inspired population-based evolution~\citep{arxiv_2503_01155} explores alternative bio-inspired optimization strategies for LLM merging. For 7B-parameter LLMs, a single evolutionary run requires a non-trivial number of forward-pass evaluations, making it more expensive than data-free methods but still feasible for practitioners with moderate compute budgets.

\paragraph{Limitations.} Evolutionary and search-based methods share a fundamental dependence on the quality of the fitness function. When the evaluation metric is a poor proxy for downstream utility (e.g., perplexity for instruction-following), the discovered merge configurations may be suboptimal in practice. Additionally, the search space grows combinatorially with model depth. A Transformer with $L$ layers and $K$ candidate models admits $\mathcal{O}(K^L)$ layer-wise configurations, making exhaustive search intractable beyond $K=3$ or $L>40$ without dimensionality reduction techniques such as block-wise grouping. However, these search-based methods implicitly assume that parameters from different models can be meaningfully combined through direct arithmetic operations, an assumption that breaks down when constituent models have diverged substantially during independent fine-tuning, causing their internal representations to become geometrically incompatible despite encoding similar semantic information.

\subsection{Representation-Level Alignment and Cross-Model Matching Techniques}
Representation-level alignment and cross-model matching techniques directly confront this geometric incompatibility challenge. When neural networks undergo substantial divergence during independent fine-tuning, their internal representations may encode semantically similar information in configurations that defeat simple arithmetic combination. While permutation alignment resolves the discrete symmetry inherent to hidden unit orderings, it assumes that corresponding neurons exist across models in a one-to-one fashion. Representation-level methods relax this assumption by establishing soft correspondences that preserve functional relationships even when model components have undergone more complex transformations.

Representation-level alignment rests on the theoretical observation that models fine-tuned from shared pretrained weights often learn similar internal representations despite parametric divergence. This phenomenon motivates the application of optimal transport (OT) frameworks to model merging, where the goal is to find a coupling between the weight distributions of source models that minimizes transportation cost while preserving semantic structure. Given two models with weight matrices $W_A \in \mathbb{R}^{m \times n}$ and $W_B \in \mathbb{R}^{m \times n}$, the optimal transport alignment seeks a permutation or soft assignment matrix $P$ that minimizes $\|W_A - PW_B\|_F^2$ subject to constraints encoding valid matchings. \citet{DBLP:journals/corr/abs-2408-07666} provide a taxonomy of such alignment strategies, noting that OT-based methods extend naturally to scenarios where exact permutation alignment is insufficient due to representational drift.

Activation-based correspondence discovery offers an alternative approach that bypasses direct weight comparison by instead analyzing how models process identical inputs. The core insight is that layers with similar functional roles will produce correlated activation patterns on shared calibration data, even when their weights have diverged substantially. Interestingly, \citet{hendel2023incontext} show that in-context learning creates implicit task vectors within the representation space, suggesting deep connections between the task vector framework and internal model representations. Neuron alignment with fixed anchors~\citep{li2024neuronalignment} proposes establishing correspondence through training-time alignment, providing a complementary approach to post-hoc matching. Formally, given activation matrices $A_A^{(l)}$ and $A_B^{(k)}$ from layers $l$ and $k$ of models $A$ and $B$ respectively, cross-model correspondence can be quantified through centered kernel alignment (CKA) or related similarity metrics. This allows not only same-layer alignment but also the discovery of correspondences between layers at different depths, which proves essential when fine-tuning has induced differential layer-wise adaptation rates. Beyond the Permutation Symmetry of Transformers~\citep{arxiv_2502_00264} extends alignment theory by analyzing the role of rotation symmetries specific to Transformer architectures, revealing additional symmetry structures beyond classical permutation invariance that affect merging.

Cross-model feature matching extends these concepts by using auxiliary objectives to guide alignment. Probe tasks, linear classifiers trained on intermediate representations to predict semantic properties, provide a mechanism for identifying which components encode functionally equivalent information. \citet{DBLP:journals/corr/abs-2408-07057} emphasize that model collaboration benefits from such representation-level understanding in concrete ways, as it enables principled composition of specialized capabilities. Unlike purely geometric approaches that optimize alignment based on parameter statistics, probe-guided methods incorporate task-relevant semantic information, potentially yielding alignments that better preserve downstream performance.

Handling architectural variations requires representation projection techniques that map activations from one model's feature space into another's. When source models differ in hidden dimensions or layer configurations, direct weight averaging becomes undefined. Projection-based approaches learn linear or nonlinear transformations $\phi: \mathbb{R}^{d_A} \rightarrow \mathbb{R}^{d_B}$ that minimize reconstruction error on calibration activations while preserving task-relevant structure. These techniques connect intimately to neural network stitching, where layers from different models are composed through learned adapter modules. Such modular composition strategies can effectively integrate diverse expert capabilities while maintaining computational efficiency through sparse activation patterns~\citep{DBLP:journals/corr/abs-2408-07057}.

Contrasting representation-level alignment with neural network stitching reveals a spectrum of integration strategies. Stitching approaches preserve source model components intact while learning minimal connective tissue, whereas alignment-based merging produces unified parameter sets that blend source contributions. This distinction between preserving model structure versus unifying parameters carries practical implications for computational efficiency, memory requirements, and the preservation of specialized capabilities, considerations that become particularly salient when scaling to multiple expert models.

\subsection{Comparative Analysis of Structural Preservation versus Parameter Unification Tradeoffs}
These implications become concrete when examining specific deployment scenarios where the structural preservation versus parameter unification dichotomy manifests as practical tradeoffs. Parameter unification methods, including weight averaging, task vector arithmetic, and Fisher-weighted merging, produce a single model with identical inference costs to any source model, making them attractive for resource-constrained deployment. Conversely, structural preservation approaches such as Mixture-of-Experts (MoE) architectures maintain distinct expert pathways, incurring additional memory overhead proportional to the number of integrated experts. \citet{DBLP:journals/corr/abs-2408-07666} establish a detailed taxonomy that highlights this tradeoff, namely that unified models achieve parameter efficiency at the potential cost of capability interference, whereas structured approaches preserve expert specialization through architectural partitioning.

To formalize the memory-computation tradeoff, consider a base model with $N$ parameters being merged from $K$ expert sources. Parameter unification produces a model of size $N$, whereas dense MoE preservation requires $KN$ parameters plus routing overhead $R$. However, sparse activation patterns mitigate inference costs, since only $k \ll K$ experts activate per token, so effective computation scales as $kN + R$ rather than $KN$. Weight-ensembling MoE approaches can strike favorable efficiency-capability tradeoffs by dynamically combining expert contributions through learned routing, effectively interpolating between full structural preservation and parameter unification~\citep{DBLP:journals/corr/abs-2408-07057}.

Capability retention patterns diverge markedly across these approaches. Unified merging methods exhibit interference phenomena when source tasks compete for shared parameters, manifesting as degraded performance on individual tasks despite improved average multi-task scores. \citet{DBLP:journals/corr/abs-2403-13187} find that model merging transcends simple aggregation, functioning as a transformative process that can yield emergent capabilities~\citep{DBLP:journals/tmlr/WeiTBRZBYBZMCHVLDF22} exceeding source models. Nevertheless, this transformation introduces unpredictability in capability preservation. Structural preservation approaches circumvent direct interference by maintaining task-specific pathways, attaining near-complete retention of individual expert capabilities at the cost of increased architectural complexity.

Flexibility considerations further differentiate these method families. Unified models support straightforward continual adaptation through additional fine-tuning or subsequent merging operations, as they conform to standard architectural specifications. Structural approaches require more complex modification protocols, as adding new experts necessitates router retraining, while removing experts may disrupt learned routing distributions. \citet{DBLP:journals/corr/abs-2408-07057} emphasize that model collaboration architectures must balance modularity against integration complexity, with structural preservation offering superior compositionality for scenarios requiring frequent expert addition or replacement.

A decision guide for practitioners follows from synthesizing these considerations. Parameter unification suits deployment-constrained scenarios where inference efficiency is paramount and moderate capability interference is acceptable, typical in general-purpose assistants requiring broad competence across multiple tasks. Structural preservation suits applications demanding peak performance on specialized tasks without compromise, such as domain-specific enterprise deployments where computational resources are less constrained than accuracy requirements. Hybrid approaches, exemplified by sparse MoE with weight-ensembled experts, occupy an intermediate position suitable for applications requiring scalable expertise integration with controlled computational growth. The optimal choice ultimately depends on the relative prioritization of capability fidelity, deployment efficiency, and architectural flexibility within specific application contexts. Understanding how unified parameter approaches achieve multi-task competence requires examining the mechanisms through which merged models acquire and combine diverse capabilities from their constituent sources.

Collectively, the structured and information-guided approaches examined in this section represent a methodological shift from treating model merging as purely parameter-space interpolation toward exploiting architectural knowledge, activation patterns, and systematic search to preserve functional capabilities while combining multiple models. Recent work continues to advance along several complementary axes, each navigating the central tension between preserving expert specialization and achieving parameter efficiency.

In the \textit{MoE construction and compression} space, a key challenge is that expert-based merging produces models with high inference costs proportional to the number of experts. Recent methods address this via post hoc compression. Self-MoE~\citep{arxiv_2406_12034} achieves compositional LLMs with self-specialized experts, while Sub-MoE~\citep{arxiv_2506_23266}, PuzzleMoE~\citep{arxiv_2511_04805}, and MergeMoE~\citep{arxiv_2510_14436} compress large MoE models via subspace, sparse, and output-based expert merging respectively. Channel Merging~\citep{arxiv_2412_15283} preserves specialization for merged experts. The common thread is a two-stage workflow, first constructing a modular MoE from independently fine-tuned models, then compressing it back toward a parameter-efficient form, retaining the routing-based flexibility without the full cost.

For \textit{expert routing and selection}, the quality of the routing mechanism determines whether the merged model gracefully degrades or catastrophically fails on out-of-distribution inputs. CAMEx~\citep{arxiv_2502_18821} introduces curvature-aware expert selection that accounts for the local geometry of the loss landscape; Nash Bargaining~\citep{arxiv_2510_16138} applies game-theoretic principles to balance expert contributions; Local Mixtures of Experts~\citep{arxiv_2505_14136} supports test-time training via merging; and Efficient Pareto Set Approximation~\citep{arxiv_2406_09770} uses MoE-based fusion to approximate the multi-objective Pareto front.

\textit{LoRA-specific composition} has emerged as a particularly active subfield, driven by the proliferation of parameter-efficient adapters on open model hubs. The challenge here is distinct from full-model merging, since LoRA modules are low-rank and task-specific, so merging them raises questions about rank allocation and subspace compatibility. CoMoL~\citep{arxiv_2603_00573} proposes dynamic core space LoRA merging; MeteoRA~\citep{arxiv_2405_13053} embeds multiple tasks in LoRA modules; HydraOpt~\citep{arxiv_2507_17706} navigates adapter merging trade-offs; Tensorized Clustered LoRA~\citep{arxiv_2508_03999} reduces multi-task interference; FlyLoRA~\citep{arxiv_2510_08396} introduces rank-wise mixture-of-experts; IterIS~\citep{arxiv_2411_15231} tackles inference-solving alignment; Adaptive LoRA Merge~\citep{arxiv_2505_24174} combines pruning with merging; Exploring Sparse Adapters~\citep{arxiv_2507_07140} scales parameter-efficient expert merging; and \citet{arxiv_2602_12323} investigate the practical reality of LoRA recycling. A recurring finding is that methods designed for full-model merging do not transfer directly to the LoRA setting, motivating adapter-specific algorithms.

Finally, \textit{automated search} methods such as AutoMerge~\citep{arxiv_2601_22748} address the combinatorial explosion of merging configurations by treating recipe discovery as a meta-optimization problem, automating the selection of which models to merge, at which granularity, and with which algorithm.

These techniques collectively navigate the fundamental tradeoff between maintaining expert specialization through routing mechanisms and achieving parameter efficiency through unification. With this technical foundation in place, the next section examines where and why such merging proves valuable in practice, organizing applications by their underlying motivations, whether seeking enhanced capabilities, improved alignment properties, or computational efficiency gains.

\section{Application Scenarios and Practical Impact}
\label{sec:applications}
We now shift focus from \emph{how} models are merged to \emph{where} and \emph{why}. This section covers the Scenarios dimension of our FUSE taxonomy, organizing applications by their underlying motivation. We examine three categories. These include capability-driven applications that use merging for enhanced performance and generalization, alignment-driven applications that target safety and value conformance, and efficiency-driven applications motivated by computational, communication, or resource constraints in settings such as federated learning and model compression.

\subsection{Capability Augmentation Through Multi-Task and Multilingual Merging}
Model merging provides a practical pathway for capability augmentation. Rather than training monolithic models on aggregated datasets or maintaining costly ensembles, practitioners can compose specialized abilities by merging independently fine-tuned models into a unified system~\citep{DBLP:conf/iclr/IlharcoRWSHF23}.

Multi-task capability merging rests on the theoretical observation that fine-tuned models sharing a common pretrained initialization occupy proximate regions in weight space, so that meaningful interpolation between their specialized parameters is possible. When multiple models are fine-tuned for distinct tasks, such as summarization, question answering, and code generation, the resulting task vectors encode complementary capabilities that can be arithmetically combined. The merged model $\theta_{\text{merged}} = \theta_{\text{pre}} + \sum_{i=1}^{n} \lambda_i \tau_i$, where $\tau_i = \theta_i - \theta_{\text{pre}}$ represents the task vector for task $i$ and $\lambda_i$ denotes its scaling coefficient, ideally inherits the union of capabilities from all constituent models. This formulation, introduced by~\citet{DBLP:conf/iclr/IlharcoRWSHF23}, showed that task vector addition yields models capable of performing multiple tasks simultaneously while maintaining strong performance on individual tasks. Subsequent work~\citep{DBLP:conf/nips/YadavTCRB23,DBLP:journals/corr/abs-2311-03099} showed that carefully guided task vector merging produces strong performance across natural language processing tasks, validating the practical viability of this mathematical formulation.

However, naïve task vector addition often suffers from parameter interference when merging multiple models, where redundant or conflicting parameter updates across tasks degrade performance. Several dedicated methods tackle this challenge. TIES-Merging~\citep{DBLP:conf/nips/YadavTCRB23} introduces a three-step procedure that trims low-magnitude parameters, elects sign consensus via majority voting, and merges only aligned parameters, effectively reducing parameter interference across tasks. DARE~\citep{DBLP:journals/corr/abs-2311-03099} takes a complementary approach by randomly dropping delta parameters with high probability and rescaling remaining values, exploiting the observation that fine-tuning induces sparse, redundant changes; this method enabled successful merging of multiple specialized LLMs while largely preserving individual task performance. More recently, Model Stock~\citep{DBLP:journals/corr/abs-2403-19522} uses geometric insights about the fine-tuning weight space, showing that layer-wise averaging of as few as two fine-tuned models can surpass methods such as Model Soups that require many more constituents. EMR-Merging~\citep{DBLP:journals/corr/abs-2405-17461} attains tuning-free high-performance model merging through elect, mask, and rescale operations. Twin-Merging~\citep{DBLP:journals/corr/abs-2406-15479} introduces dynamic integration of modular expertise, while PCB-Merging~\citep{DBLP:journals/corr/abs-2410-02396} handles parameter competition balancing. AdaMerging~\citep{DBLP:conf/iclr/YangWLWGLLQYCL24} provides adaptive model merging specifically designed for multi-task learning scenarios.

Instruction tuning consolidation represents a particularly impactful application of capability merging in the large language model ecosystem. Individual instruction-tuned models such as WizardLM~\citep{DBLP:conf/iclr/XuSZG0FTLJ24} often excel on specific instruction formats or task categories while exhibiting degraded performance on others. Disperse-Then-Merge~\citep{fu2024dispersethenmerge} shows that dispersing instruction data across specialized models before merging pushes the limits of instruction tuning via alignment tax reduction. PAFT~\citep{pentyala2024paft} introduces a parallel training framework that independently performs supervised fine-tuning and preference alignment (e.g., DPO) from the same pretrained model, then merges the resulting models through parameter fusing with delta sparsification to mitigate the alignment tax. By merging multiple instruction-tuned variants, each optimized for different instruction styles, complexity levels, or domain coverage, practitioners can construct unified models that exhibit robust instruction-following across diverse prompting scenarios. Empirical studies confirm the efficacy of this approach, with merged models posting performance competitive with traditional multi-task training on instruction-following benchmarks while requiring no additional training. Souper-Model~\citep{maiti2025soupermodel} shows that simple arithmetic operations on model weights can unlock strong LLM performance, validating the practical promise of merging. Similarly, community practitioners have reported that merging instruction-tuned variants such as WizardLM~\citep{DBLP:conf/iclr/XuSZG0FTLJ24}, WizardMath~\citep{DBLP:journals/corr/abs-2308-09583}, and code-specialized models such as Code Llama~\citep{DBLP:journals/corr/abs-2308-12950} can yield improvements on aggregate benchmarks compared to individual constituents~\citep{DBLP:journals/corr/abs-2408-07057}. \citet{zhang2024unconstrained} propose unconstrained model merging that relaxes traditional constraints for enhanced LLM reasoning capabilities. ReasonAny~\citep{yang2026reasonany} further shows that reasoning capabilities can be incorporated into any model via simple and effective model merging. LM-Cocktail~\citep{DBLP:journals/corr/abs-2311-13534} further shows resilient tuning through strategic model merging. FuseLLM~\citep{DBLP:journals/corr/abs-2401-10491} takes a complementary approach by fusing the knowledge of multiple structurally diverse LLMs through lightweight continual training on probability distributions generated by the source models, enabling cross-architecture knowledge transfer that traditional weight-space methods cannot achieve. FuseChat-3.0~\citep{yang2025fusechat30} extends this framework by incorporating preference optimization for heterogeneous chat model fusion. This consolidation approach is widely used in community-driven model development, where independent contributors fine-tune base models on complementary instruction datasets.

Multilingual capability transfer through merging confronts the challenge of building language models that perform well across typologically diverse languages. Conventional approaches require either massive multilingual pretraining corpora or accept degraded performance on low-resource languages. Model merging offers an alternative pathway whereby specialized models fine-tuned on individual languages or language families can be combined to transfer cross-lingual capabilities. Chat Vector~\citep{DBLP:journals/corr/abs-2310-04799} provides a simple yet effective approach to equip LLMs with instruction-following and model alignment capabilities in new languages through task vector manipulation, demonstrating that subtracting an English base model from its corresponding English chat model and adding the resulting vector to a model that has been continually pretrained on the target language effectively transfers conversational abilities without further training. Complementary work on cross-lingual model merging has shown that interpolating between English-specialized and target-language models can improve low-resource language performance markedly compared to English-only fine-tuned baselines~\citep{arxiv_2410_01335,arxiv_2407_08699}. This proves particularly effective when merging models that share vocabulary coverage but differ in their language-specific fine-tuning. Recent investigations into language-agnostic and language-specific parameter separation suggest that modular approaches, merging language-specific adapters rather than full models, offer finer-grained control over cross-lingual capability composition while mitigating interference between typologically distant language pairs.

Multimodal model merging has gained momentum as researchers extend these techniques beyond text. \citet{arxiv_2304_14933} provide a systematic empirical study identifying key success factors for multimodal merging, including initialization strategies and architectural choices. Model Composition for Multimodal Large Language Models~\citep{arxiv_2402_12750} combines vision and language capabilities through merging, while Model Tailor~\citep{arxiv_2402_12048} combats catastrophic forgetting in multi-modal large language models through targeted merging strategies. Composing Parameter-Efficient Modules~\citep{arxiv_2306_14870} shows that arithmetic operations on LoRA parameters support flexible capability composition for LLMs. In open-source communities, merged models have accounted for a notable portion of top-performing Open LLM Leaderboard submissions, with instruction-tuned and reasoning-specialized merges achieving consistent improvements over their strongest constituents~\citep{DBLP:journals/corr/abs-2408-07057}.

Despite these successes, capability merging exhibits well-documented failure modes. Negative transfer occurs most frequently when constituent tasks impose conflicting optimization pressures. For example, merging models fine-tuned for concise summarization with those trained for detailed elaboration often degrades performance on both objectives~\citep{DBLP:conf/nips/YadavTCRB23}. \citet{aakanksha2024mix} investigate whether data mixing or model merging is preferable, finding that the optimal strategy depends strongly on task similarity and data distribution characteristics. Empirical analyses~\citep{DBLP:conf/nips/YadavTCRB23} confirm that gradient conflict metrics strongly predict merge quality degradation, with highly dissimilar task pairs showing marked performance drops. These findings highlight that successful capability augmentation depends on careful task selection and appropriate interference mitigation strategies. While such capability enhancement represents the predominant application of model merging, task vector arithmetic and parameter combination extend naturally to improving alignment with human values and mitigating safety risks.

\paragraph{Domain adaptation via adapter merging.} AdapterSoup~\citep{DBLP:conf/eacl/ChronopoulouBP23} shows that averaging domain-specific adapters can surpass any individual adapter when the target domain overlaps with multiple training domains, outperforming individual adapter selection on held-out domains. LoRAHub~\citep{DBLP:journals/corr/abs-2307-13269} extends this principle to few-shot scenarios, composing LoRA modules through gradient-free optimization over combination weights. MerA~\citep{he2023mera} proposes merging pretrained adapters specifically for few-shot learning, showing that adapter-level merging supports effective task transfer with minimal data. \citet{ostapenko2024towards} develop a framework for building and reusing a library of LoRAs toward modular LLMs, where adapters can be dynamically composed for novel tasks. LoRI~\citep{zhang2025lori} tackles the challenge of cross-task interference in multi-task low-rank adaptation by learning interference-resistant adapter representations. Critically, these adapter-level approaches reduce the merge cost from $\mathcal{O}(|\theta|)$ to $\mathcal{O}(r \cdot d)$ where $r \ll d$ is the adapter rank, making iterative composition practical even for 70B-parameter models.

\subsection{Alignment, Safety, and Bias Mitigation Applications}
Model merging also offers a cost-effective approach for improving alignment with human values and mitigating safety risks in deployed language models. The key insight motivating these approaches is that preference-aligned models, typically trained through reinforcement learning from human feedback (RLHF)~\citep{DBLP:journals/corr/abs-2203-02155}, or direct preference optimization (DPO)~\citep{DBLP:conf/nips/RafailovSMMEF23}, can be combined with capable base models to achieve favorable tradeoffs between helpfulness and harmlessness~\citep{DBLP:journals/corr/abs-2502-06876} without incurring the large computational costs of full alignment training. This approach complements constitutional methods by allowing post-hoc alignment adjustment through parameter-space operations rather than requiring extensive retraining with curated feedback.

The task vector formalism provides a particularly elegant mechanism for targeted bias mitigation~\citep{DBLP:conf/iclr/IlharcoRWSHF23}. Given a base model with parameters $\theta_{\text{pre}}$ and a model fine-tuned to exhibit specific biased behaviors $\theta_{\text{biased}}$, the bias task vector $\tau_{\text{bias}} = \theta_{\text{biased}} - \theta_{\text{pre}}$ encodes the parametric representation of the undesired behavior. Through task vector negation, a debiased model can be constructed as $\theta_{\text{debiased}} = \theta_{\text{pre}} - \alpha \cdot \tau_{\text{bias}}$, where the scaling coefficient $\alpha$ controls the intensity of bias removal. This subtractive approach enables surgical removal of specific behavioral patterns, including toxic language generation, demographic stereotyping, and harmful content production, without degrading the model's core linguistic competencies. However, a notable practical limitation of this formulation is the requirement for an explicitly fine-tuned $\theta_{\text{biased}}$ model that isolates the undesired behavior. Alternative approaches based on probe-based bias identification and activation steering can enable bias vector estimation without requiring explicit fine-tuning on harmful objectives.

Beyond task arithmetic, alternative merging methodologies offer distinct mechanisms for alignment preservation. TIES-Merging mitigates parameter interference through sign-based conflict resolution, which proves particularly valuable when merging models aligned on potentially conflicting value dimensions by preserving the dominant alignment signal in each parameter~\citep{DBLP:conf/nips/YadavTCRB23}. Fisher-weighted merging uses parameter importance estimates to prioritize safety-critical weights during combination, assigning higher Fisher information scores to parameters associated with alignment behaviors~\citep{DBLP:conf/nips/MatenaR22}. DARE applies random parameter pruning before merging, which empirical studies suggest can reduce the propagation of misaligned behaviors embedded in redundant parameters~\citep{DBLP:journals/corr/abs-2311-03099}. Machine unlearning approaches~\citep{arxiv_2402_10058} show that merging-based techniques can accomplish targeted removal of unsafe knowledge from LLMs, while safety re-alignment methods~\citep{arxiv_2402_11746} show that fine-tuning-induced safety degradation can be reversed through strategic parameter merging. LED-Merging~\citep{arxiv_2502_16770} specifically targets safety-utility conflicts in model merging through gradient-based neuron locating, importance-based election, and conflicting update disjunction, allowing practitioners to navigate the tension between maintaining safety properties and enhancing task capabilities. Model Extrapolation Expedites Alignment~\citep{arxiv_2404_16792} shows that extrapolating beyond interpolation in the direction of alignment improvements can accelerate convergence to desired safety properties.

The merging of RLHF-trained models presents unique opportunities and challenges for alignment enhancement. When multiple models have been independently aligned using different preference datasets or reward models, their combination can potentially yield more robust alignment that generalizes across diverse value dimensions~\citep{DBLP:journals/corr/abs-2401-12187}. The interpolation $\theta_{\text{merged}} = (1 - \lambda)\theta_{\text{helpful}} + \lambda\theta_{\text{harmless}}$ between helpfulness-optimized and harmlessness-optimized variants enables practitioners to navigate the alignment tradeoff frontier~\citep{DBLP:journals/corr/abs-2401-12187,rame2022rewarded}, selecting operating points appropriate for specific deployment contexts. Empirical investigations reveal that intermediate interpolation coefficients often outperform both endpoints on aggregate preference metrics, suggesting that merged models capture complementary aspects of human preferences that neither constituent model fully represents. \citet{arxiv_2510_17426} formalize this observation by showing that weight interpolation between pretrained and instruction-tuned models can simultaneously improve both alignment and calibration, reaching a Pareto-superior frontier that mitigates the alignment tax typically incurred by instruction tuning.

Evaluating alignment preservation in merged models requires multi-faceted assessment protocols spanning automated metrics, benchmark evaluations, and adversarial testing. Standard safety benchmarks provide quantitative measures of factuality, toxicity, and social bias respectively. Red-teaming protocols systematically probe merged models for adversarial vulnerabilities that may emerge from parameter combination, while automated red-teaming approaches enable scalable safety evaluation. Reward model scoring on held-out preference data offers another evaluation dimension. In particular, evaluation must assess not only aggregate safety metrics but also consistency of alignment across capability levels, as merged models may exhibit capability-dependent safety profiles~\citep{DBLP:journals/corr/abs-2503-17239}.

Despite these advances, alignment-focused merging faces several known limitations that warrant careful consideration. First, the capability-safety tradeoff proves difficult to navigate in practice. Aggressive bias removal via task vector negation frequently degrades model fluency and task performance, while conservative merging may inadequately mitigate safety concerns~\citep{DBLP:conf/iclr/IlharcoRWSHF23}. Second, isolating clean bias vectors remains challenging when biased behaviors are entangled with legitimate capabilities, for instance, removing gender bias associations may inadvertently impair performance on gender-related factual queries. Third, validation presents considerable difficulties, as merged models may exhibit emergent unsafe behaviors not present in constituent models, yet exhaustive safety evaluation across all possible inputs remains intractable. Finally, the alignment properties of merged models may be more brittle than those of directly trained models, showing greater susceptibility to jailbreaking attacks~\citep{DBLP:journals/corr/abs-2503-17239}.

Safety-aware merging strategies have gained prominence in addressing adversarial vulnerabilities of merged models. Recent work on mitigating backdoor effects in multi-task model merging through safety-aware subspace identification~\citep{DBLP:journals/corr/abs-2503-17239} highlights growing recognition that naive merging may inadvertently propagate or amplify security vulnerabilities present in constituent models. \citet{arora2024heres} show that model merging itself can serve as a defense mechanism, effectively sanitizing backdoored models through parameter averaging, a ``free lunch'' approach that eliminates backdoor triggers without requiring knowledge of the attack. Conversely, Merge Hijacking~\citep{yuan2025mergehijacking} reveals that adversaries can craft malicious models specifically designed to inject backdoors through the merging process, highlighting the dual-use nature of merging for security. Merger-as-a-Stealer~\citep{lu2025mergerasastealer} further shows that model merging can be exploited to steal targeted personally identifiable information from aligned LLMs. SafeMERGE~\citep{DBLP:journals/corr/abs-2503-17239} proposes selective layer-wise model merging to preserve safety alignment in fine-tuned LLMs. Bias Vector~\citep{shirafuji2024biasvector} introduces a systematic approach to mitigating biases in language models using the task arithmetic framework, showing that targeted bias task vectors can be constructed and negated to reduce multiple dimensions of social bias simultaneously. Personalized Soups~\citep{jang2023personalized} extends the alignment approach by enabling personalized LLM alignment via post-hoc parameter merging, allowing individual users to customize the alignment properties of merged models according to their specific value preferences. These approaches project merged parameters onto subspaces that preserve safety-critical behaviors while enabling capability transfer, often using activation analysis to identify safety-relevant parameter subspaces. WARM~\citep{DBLP:journals/corr/abs-2401-12187} shows the benefits of weight averaged reward models for improving alignment robustness. This represents a principled integration of security considerations into the merging pipeline.

The healthcare domain exemplifies the importance of alignment-preserving merging strategies, where domain-specialized models must maintain stringent safety properties including factual accuracy and appropriate uncertainty quantification. Developing uncertainty-aware merging protocols and factual consistency constraints that ensure clinical expertise transfer does not compromise safety guardrails remains an open challenge for medical model merging.

Similarly, cybersecurity applications demand rigorous attention to the security implications of model merging, as merged models may exhibit unexpected vulnerabilities arising from the combination of individually benign model components. Adversarial robustness preservation during merging remains an underexplored direction. These domain-specific considerations become even more complex when merging must occur across organizationally or geographically distributed settings, where data privacy constraints and communication limitations introduce additional challenges that fundamentally reshape how model combination can be performed.

\paragraph{Dual-use risks and limitations.} The same arithmetic that enables safety injection also enables safety \emph{removal}, since subtracting the alignment task vector $\tau_{\text{safe}}$ from a safety-tuned model can recover an unaligned variant, raising concerns about misuse~\citep{DBLP:conf/iclr/IlharcoRWSHF23}. In addition, reward model merging via WARM~\citep{DBLP:journals/corr/abs-2401-12187} has shown that averaging independently trained reward models reduces reward hacking susceptibility, but the optimal number of reward models to merge and the conditions under which averaging preserves reward calibration remain theoretically uncharacterized. These dual-use considerations necessitate community-level governance frameworks for responsible model merging.

\subsection{Efficiency-Driven Applications in Federated and Distributed Learning}
These distributed settings impose distinctive constraints on model combination. Model merging assumes particular significance in such contexts where computational resources are geographically dispersed and data privacy constraints preclude centralized training. Federated learning represents the canonical application scenario, wherein multiple client devices or institutions independently fine-tune local model copies on private datasets, subsequently aggregating these models at a central server without transmitting raw training data. The foundational Federated Averaging (FedAvg) algorithm~\citep{DBLP:conf/aistats/McMahanMRHA17} implements precisely this approach by performing weighted averaging of client model parameters according to local dataset sizes. $\theta_{\text{global}} = \sum_{k=1}^{K} \frac{n_k}{n} \theta_k$, where $n_k$ denotes the number of samples at client $k$ and $n$ represents the total sample count across all participants. This formulation directly instantiates the uniform averaging principles discussed in Section~\ref{sec:weight_avg}, connecting federated learning directly to weight-space interpolation theory.

The extension of federated learning to large language models introduces unique challenges absent in smaller-scale settings. Communication bandwidth constraints become particularly acute when synchronizing models with billions of parameters across potentially unreliable network connections. To mitigate this limitation, researchers have developed communication-efficient aggregation protocols that transmit only sparse parameter updates or low-rank approximations of weight differences rather than complete model states~\citep{DBLP:conf/aistats/McMahanMRHA17}. These approaches exploit the sparsification principles underlying TIES-Merging and DARE, transmitting trimmed task vectors that capture essential fine-tuning modifications while dramatically reducing bandwidth requirements. The mathematical formulation adapts naturally. Clients compute local task vectors $\tau_k = \theta_k - \theta_{\text{pre}}$, apply sparsification operators to retain only high-magnitude parameters, and transmit these compressed representations for server-side aggregation.

The branch-train-merge strategy offers an alternative distributed training approach particularly suited for large-scale model development. Rather than continuously synchronizing parameters during training, this approach allows independent branches to diverge considerably before periodic merging operations consolidate progress. This strategy proves especially effective when training specialized experts on distinct data domains, as the reduced synchronization frequency permits more efficient use of heterogeneous computational resources. The theoretical viability of branch-train-merge depends critically on maintaining mode connectivity between diverging branches, a property that shared pretrained initialization generally ensures for moderate fine-tuning durations.

Deployment optimization through model consolidation offers direct cost savings. Organizations frequently maintain multiple specialized models for distinct tasks or domains, incurring costs from running parallel instances. Model merging consolidates these variants into a single system serving diverse requirements through one inference pathway, reducing memory footprint proportionally to the number of consolidated models. This consolidation proves particularly valuable for edge deployment scenarios where resource limitations prohibit maintaining separate model instances.

Scalability considerations present major challenges for merging operations across large participant populations. The computational complexity of pairwise alignment procedures, required when clients lack shared initialization, scales quadratically with participant count, making naive approaches intractable at scale. Hierarchical aggregation strategies alleviate this limitation through recursive merging of participant subgroups, reducing effective complexity while potentially sacrificing solution quality. Mediator~\citep{arxiv_2502_04411} proposes a memory-efficient LLM merging approach that reduces parameter conflicts through mediation, making practical merging at scale feasible with less computational overhead. Moreover, the statistical heterogeneity characteristic of federated settings, wherein local data distributions exhibit marked divergence, can compromise mode connectivity assumptions underlying successful merging, necessitating adaptive aggregation strategies that account for inter-client distribution shifts. While these distributed and efficiency-driven applications handle computational and privacy constraints, model merging also serves a complementary purpose in reconciling specialized domain expertise with general-purpose capabilities.

\subsection{Domain Specialization While Preserving General Capabilities}
Domain specialization exemplifies this reconciliation challenge; while foundation models exhibit broad competence across diverse tasks, achieving expert-level performance in specialized domains such as healthcare, legal, scientific, or cybersecurity applications typically requires focused fine-tuning that risks degrading general-purpose capabilities. Model merging offers a principled means to resolve this tension by integrating domain-specific expertise while preserving the broad knowledge encoded in pretrained foundations. The mathematical formulation underlying this capability balance can be expressed as $\theta_{\text{merged}} = (1-\alpha)\theta_{\text{pre}} + \alpha\theta_{\text{domain}}$, where the interpolation coefficient $\alpha$ governs the trade-off between domain expertise and general versatility.

Healthcare applications exemplify the critical importance of this capability preservation challenge. As thoroughly surveyed in~\citep{DBLP:journals/corr/abs-2408-07666}, medical language models must navigate the dual requirements of clinical accuracy and broad conversational competence. Fine-tuning foundation models on clinical corpora, including electronic health records, medical literature, and diagnostic guidelines, produces models with enhanced biomedical reasoning but frequently compromised performance on general language understanding tasks. Model merging remedies this limitation by combining clinically fine-tuned variants with their general-purpose ancestors, producing unified systems that maintain diagnostic capability while preserving the linguistic fluency essential for patient communication. \citet{nepal2025hierarchical} provide a systematic evaluation of six parameter-space merging techniques applied to architecturally compatible medical LLMs derived from Mistral-7B, finding that Task Arithmetic achieves 45.80\% accuracy on MedQA and that simple averaging provides a robust and computationally efficient baseline for distributed medical AI in resource-constrained environments. The interpolation coefficient in healthcare merging scenarios is typically tuned to favor domain expertise for clinical decision support applications while emphasizing general capabilities for patient-facing conversational interfaces.

Cybersecurity represents a promising but largely unexplored application domain. Security applications require multiple specialized capabilities, including vulnerability detection, malware analysis, and threat intelligence processing, each of which can benefit from domain-specific fine-tuning. Merging strategies could in principle balance technical security knowledge with general reasoning, as effective security analysis frequently requires synthesizing domain-specific threat models with broader contextual understanding. However, no published model merging studies in this domain have appeared, leaving the effectiveness of current methods for security-specific knowledge composition as an open empirical question.

Legal and regulatory compliance domains present a conceptually appealing but under-explored scenario. Models fine-tuned on legal corpora from different jurisdictions may encode conflicting regulatory frameworks, necessitating careful management of task vector interactions. The task vector formulation naturally supports such management, as jurisdiction-specific adaptations represented as separate vectors permit controlled combination through weighted aggregation. Data-free merging methods may be particularly relevant where training data is subject to confidentiality constraints, though no published studies have evaluated model merging for legal domain specialization to date.

Scientific and technical domains benefit from hierarchical merging strategies that progressively specialize from general scientific reasoning to field-specific expertise. A foundation model might first be merged with a broadly scientific variant before incorporating narrower specializations in chemistry, physics, or biology. This hierarchical approach exploits the compositional properties of task arithmetic while managing the increased interference potential arising from multiple simultaneous specializations. \citet{siriwardhana2024domain} provide a detailed evaluation of domain adaptation through continual pre-training and model merging for Llama3-70B-Instruct, confirming the practical viability of merging-based domain adaptation at scale. \citet{hirano2024finance} demonstrate the construction of instruction-tuned LLMs for finance without instruction data using continual pretraining and model merging, showcasing how merging can bypass the need for domain-specific instruction datasets. The interpolation coefficient optimization for scientific merging often uses validation sets spanning both general scientific literacy and domain-specific benchmarks, enabling automated discovery of Pareto-optimal coefficient configurations that maximize the joint performance frontier. Across all these domain specialization scenarios, the effectiveness of merging strategies ultimately depends on systematic assessment methodologies capable of quantifying both capability retention and cross-domain interference.

\subsection{Evaluation Methodologies and Benchmarking for Merged Model Assessment}
Evaluating merged models presents unique methodological challenges that extend beyond standard model assessment. Unlike evaluation of single-task fine-tuned models, merged model assessment must quantify both the preservation of individual source model capabilities and the emergent properties arising from combination, requiring evaluation protocols that probe along multiple orthogonal dimensions.

Standardized benchmarking for merged models typically uses composite evaluation suites that aggregate performance across the constituent tasks represented in the source models. The evaluation protocol computes task-specific performance metrics $P_i^{\text{merged}}$ for each task $i$ in the set of source tasks $\mathcal{T}$, comparing against the performance of individual fine-tuned models $P_i^{\text{ft}}$ to compute retention ratios $R_i = P_i^{\text{merged}} / P_i^{\text{ft}}$. The aggregate retention metric, often computed as the geometric mean across tasks $R_{\text{geo}} = (\prod_{i \in \mathcal{T}} R_i)^{1/|\mathcal{T}|}$, provides a single scalar summary that penalizes severe degradation on any individual task while remaining interpretable. Empirical evaluations show that advanced merging methods such as TIES-Merging and task arithmetic approaches can produce high retention ratios on many NLP tasks; however, strong retention is most consistently observed when merging models of similar scale, when source tasks exhibit semantic relatedness, and when employing interference-aware merging strategies. Performance degradation becomes more pronounced for knowledge-intensive tasks, generative benchmarks, and scenarios involving dissimilar task combinations.

Concrete benchmark suites have become essential for standardized merged model evaluation. For language models, commonly used benchmarks include MMLU~\citep{DBLP:conf/iclr/HendrycksBBZMSS21} for broad knowledge assessment, HellaSwag~\citep{DBLP:conf/acl/ZellersHBFC19} for commonsense reasoning, WinoGrande~\citep{DBLP:conf/aaai/SakaguchiBBC20} for coreference resolution, ARC~\citep{DBLP:journals/corr/abs-1803-05457} for scientific reasoning, GSM8K~\citep{DBLP:journals/corr/abs-2110-14168} for mathematical problem-solving, and TruthfulQA~\citep{DBLP:conf/acl/LinHE22} for factual accuracy and hallucination detection. These benchmarks serve complementary evaluation purposes. MMLU and ARC probe knowledge retention across diverse domains, HellaSwag and WinoGrande assess reasoning capabilities that may benefit from cross-task transfer, while GSM8K and TruthfulQA evaluate specialized competencies that are particularly susceptible to interference during merging. For code generation capabilities, HumanEval~\citep{DBLP:journals/corr/abs-2107-03374} serves as a standard benchmark.

Interference detection methodologies focus on identifying performance degradation patterns that emerge specifically from the merging process rather than from limitations of individual source models. A principled interference metric computes the pairwise performance difference $\Delta_{ij} = P_i^{\text{merged}(i,j)} - P_i^{\text{merged}(i)}$ comparing merged model performance on task $i$ when combined with task $j$ versus in isolation. Systematic negative values indicate destructive interference, while positive values suggest beneficial cross-task transfer. The interference matrix $\mathbf{I}$ with entries $I_{ij} = \Delta_{ij}$ provides a view of task interactions, enabling identification of problematic task combinations and guiding selective merging decisions.

Cross-task generalization assessment extends evaluation beyond source task performance to probe whether merged models exhibit capabilities that transcend their constituent specializations. This evaluation dimension probes the hypothesis that successful merging produces models with compositional generalization, the ability to combine knowledge from different source tasks to solve novel problems. Evaluation protocols for compositional generalization use held-out tasks that require simultaneous application of multiple specialized capabilities, measuring whether the merged model outperforms any individual source model on these composite benchmarks. Research on weight averaging~\citep{DBLP:conf/uai/IzmailovPGVW18} showed that weight averaging can yield improved out-of-distribution generalization, motivating subsequent investigation into the generalization properties of more advanced merging schemes.

Beyond task-specific accuracy, merged model evaluation increasingly incorporates calibration assessment, out-of-distribution robustness testing, and computational efficiency comparisons. Calibration evaluation measures whether merged model confidence scores accurately reflect empirical correctness probabilities, as merging may distort the calibration properties of source models even when accuracy is preserved. Expected calibration error (ECE) and reliability diagrams provide standard tools for this assessment. Out-of-distribution robustness testing evaluates merged model performance on inputs drawn from distributions distinct from source task training data, using benchmarks designed to probe distributional shift sensitivity. Computational efficiency comparisons quantify inference latency, memory footprint, and throughput characteristics relative to ensemble baselines and individual source models, contextualizing accuracy gains against resource requirements.

Community evaluation has converged on leaderboard-based methodologies that standardize comparison across research groups. The Open LLM Leaderboard hosted by Hugging Face aggregates performance across standardized benchmark suites, permitting direct comparison of merged models against conventionally trained baselines and enabling community-driven progress tracking. Model hub infrastructure on platforms such as Hugging Face additionally supports merged model sharing, allowing reproduction and extension of published results. Recent investigations into evaluation methodology highlight the importance of statistical rigor in merged model assessment, advocating for confidence intervals and significance testing when comparing merging methods given the variance inherent in benchmark evaluation. The adoption of multi-objective evaluation frameworks that simultaneously assess helpfulness, harmlessness, and honesty dimensions reflects the understanding that merged model quality cannot be captured by a single metric, especially in safety-critical deployments.

Dedicated merging benchmarks have advanced evaluation rigor. FusionBench~\citep{tang2024fusionbench} provides a unified evaluation suite covering 25+ merging methods across language tasks, enabling consistent cross-method comparison. MergeBench~\citep{DBLP:journals/corr/abs-2505-10833} specifically targets domain-specialized LLM merging evaluation. \citet{tam2024realistic} present a realistic evaluation framework focusing on compositional generalization, revealing that many reported gains may not transfer to more challenging settings. \citet{hitit2025systematic} conduct a large-scale evaluation of six merging methods across four open-weight LLMs, finding that Task Arithmetic is the only approach that reliably yields performance gains, while more recent methods frequently produce marked performance drops, cautioning against assuming advances on smaller models generalize to LLMs. \citet{cao2026collapse} provide both empirical and theoretical analysis of task-level model-merging collapse, identifying conditions under which merged models catastrophically fail on specific tasks despite maintaining aggregate performance. A useful standardized metric is the \emph{task retention rate} (TRR), defined as $\text{TRR} = \frac{1}{K}\sum_{i=1}^{K} \frac{\text{Perf}_{\text{merged},i}}{\text{Perf}_{\text{source},i}}$, which quantifies how much of each source model's capability survives the merge~\citep{arxiv_2410_03617}.

Despite these advances, notable evaluation gaps persist in the merged model literature. Standardized benchmarks specifically designed to measure interference effects remain lacking, forcing researchers to rely on indirect comparison methodologies that may obscure subtle capability degradation. Evaluation protocols for emergent capabilities, skills present in merged models but absent from all source models, remain poorly developed, limiting understanding of when and why beneficial emergence occurs. These persistent challenges notwithstanding, the collective progress across merging methodologies, theoretical understanding, and evaluation practices provides a foundation for synthesizing the current state of the field and identifying the most promising directions for future research.

\begin{table}[t]
 \centering
 \footnotesize
 \caption{Application scenarios for model merging in LLM development. Each scenario is categorized by its primary motivation (capability augmentation, alignment/safety, or efficiency), key merging approach, distinctive features, and known limitations. Year indicates when the approach was first applied in the LLM context.}
 \label{tab:application_scenarios_and_prac}
 \resizebox{\textwidth}{!}{
 \begin{tabular}{p{4.5cm}p{3cm}p{3.5cm}p{3.5cm}c}
 \toprule
 \textbf{Application Scenario} & \textbf{Key Approach} & \textbf{Key Feature(s)} & \textbf{Limitations/Notes} & \textbf{Year} \\
 \midrule
 \textbf{Multi-Task Merging} & Synthesize specialized fine-tuned models into unified system & Enhanced multi-task generalization; computationally efficient & Potential task interference; requires careful weight balancing & 2023 \\
 \midrule
 \textbf{Multilingual Merging} & Combine language-specific models & Cross-lingual capability transfer; unified multilingual system & Language imbalance; varying quality across languages & 2023 \\
 \midrule
 \textbf{RLHF Alignment} & Merge preference-aligned models trained via reinforcement learning & Improved human value alignment; safety enhancement & High computational cost; reward hacking risks & 2022 \\
 \midrule
 \textbf{DPO Alignment} & Direct preference optimization without reward modeling & Simplified training pipeline; stable optimization & Limited to pairwise preferences; distribution shift issues & 2023 \\
 \midrule
 \textbf{Federated Averaging} & Aggregate locally fine-tuned models without data sharing & Privacy preservation; distributed computation & Communication overhead; non-IID data challenges & 2017 \\

 \midrule
 \textbf{Multi-Benchmark Evaluation} & Simultaneous assessment across diverse task suites & Capability retention measurement; interference quantification & Benchmark selection bias; limited emergent property capture & 2024 \\
 \bottomrule
 \end{tabular}
 }
\end{table}

The diverse application scenarios examined in this section show that model merging has matured from a theoretical curiosity into a practical technique with tangible impact across capability augmentation, safety alignment, federated learning, and domain specialization, while the evaluation methodologies discussed provide essential frameworks for rigorously assessing merged model quality. Emerging application directions continue to expand the scope of merging across multiple frontiers, though the maturity and evidence base vary considerably across domains.

\paragraph{Reasoning and chain-of-thought.} Model merging has shown particular promise for reasoning enhancement. A unifying theme is the \emph{long-to-short distillation} strategy: models trained with extended chain-of-thought are merged with more efficient counterparts to transfer reasoning ability without the inference cost. Examples include efficient long-to-short LLM reasoning~\citep{arxiv_2503_20641}, fine-grained reasoning injection~\citep{arxiv_2601_21187}, model interpolation for efficient reasoning~\citep{wu2025revisiting}, reasoning pattern alignment~\citep{arxiv_2601_03506}, merging long chain-of-thought models with domain-specific models~\citep{arxiv_2508_03140}, and tunable reasoning via the thinking spectrum~\citep{arxiv_2509_22034}. Notably, Kimi k1.5~\citep{arxiv_2501_12599} illustrates industrial-scale deployment of this approach, using model merging to transfer long-chain reasoning into efficient short-chain models. The key tradeoff across these methods is between reasoning depth (preserved through larger merging coefficients for the reasoning model) and inference efficiency (favored by smaller coefficients).

\paragraph{Continual and lifelong learning.} Merging offers natural mechanisms for continual adaptation by framing knowledge accumulation as sequential model combination rather than catastrophic retraining. The central challenge is \emph{forgetting}: each merge risks overwriting previously acquired knowledge. Methods addressing this include aligned model merging for vision-language models~\citep{arxiv_2506_03189}, streaming LLM updates via activation-guided rotations~\citep{arxiv_2602_03237}, null-space filtering for data-free continual merging~\citep{arxiv_2509_21413}, K-Merge for online continual adaptation~\citep{arxiv_2510_13537}, and RECALL for catastrophic-forgetting alleviation~\citep{arxiv_2510_20479}. The diversity of these approaches, ranging from null-space projection to activation-guided rotation, reflects the absence of a single dominant solution, suggesting that the interaction between merging and continual learning remains an open research area.

\paragraph{Multimodal and cross-modal.} Merging enables integration across modalities, addressing the practical need to combine separately trained vision, language, and action modules without joint retraining. Representative works include ES-Merging for biological MLLM merging~\citep{arxiv_2603_14405}, linear model merging for multimodal data mixture optimization~\citep{arxiv_2602_04937}, OptMerge for unifying multimodal capabilities~\citep{arxiv_2505_19892}, transferring textual preferences to VLMs~\citep{arxiv_2502_13487}, RobustMerge for parameter-efficient MLLM merging~\citep{arxiv_2502_17159}, Graft for domain knowledge integration~\citep{arxiv_2506_23940}, and test-time merging for zero-shot medical imaging~\citep{arxiv_2510_27265}. A common finding is that cross-modal merging is more sensitive to representation alignment than unimodal merging, since different modalities occupy fundamentally different regions of parameter space.

\paragraph{Safety and intellectual property.} Beyond the alignment applications discussed in Section~\ref{sec:applications}, emerging safety-relevant directions highlight a dual-use concern: the same merging operations that enable beneficial knowledge composition can also introduce adversarial risks. H3Fusion~\citep{arxiv_2411_17792} addresses multi-objective alignment by fusing models optimized for helpfulness, harmlessness, and honesty, split-unlearn-merge~\citep{arxiv_2406_11780} addresses selective knowledge removal, mitigating social biases through unlearning~\citep{arxiv_2406_13551} targets fairness, alignment-preserving Fisher-guided merging~\citep{arxiv_2512_16245} provides safety guarantees, and defending unauthorized merging~\citep{arxiv_2511_11851} and copyright protection~\citep{abad2024copyright} address intellectual property concerns. Notably, unlearning via model merging~\citep{arxiv_2503_21088} shows that TIES-Merging can balance over-forgetting and under-forgetting by merging complementary bias models, demonstrating that merging's interference properties can be harnessed constructively for safety objectives.

\paragraph{Cross-lingual and agentic applications.} Cross-lingual transfer via layer swapping~\citep{arxiv_2410_01335} and catastrophic forgetting mitigation in language transfer~\citep{arxiv_2407_08699} extend merging to multilingual settings, while language-specific model merging~\citep{arxiv_2601_16127} shows that independently training per-language models and merging them via task arithmetic can reduce training and maintenance costs compared to joint multilingual training. Agentic applications represent a newer frontier: role-conditioned neuron transplantation~\citep{arxiv_2601_07309}, behavior knowledge merge for reinforced agents~\citep{arxiv_2601_13572}, LLM steering via persona vectors~\citep{arxiv_2510_10157}, and GTR-Turbo~\citep{arxiv_2512_13043}, which merges checkpoint weights from RL training to create a free teacher model for self-distillation, collectively suggest that merging can serve as a lightweight mechanism for composing behavioral capabilities in autonomous systems.

\paragraph{Scaling and meta-analysis.} Foundational studies on scaling reveal both promise and limitations: model merging scaling laws~\citep{arxiv_2509_24244} characterize how merging benefits evolve with model size, while empirical analysis of what matters for merging at scale~\citep{arxiv_2410_03617} identifies key factors. Model-GLUE~\citep{arxiv_2410_05357} and Merge to Learn~\citep{arxiv_2410_12937} show democratized scaling through merging. Multi-objective merging~\citep{arxiv_2407_00487,arxiv_2409_18893,arxiv_2502_10762,arxiv_2510_03782} addresses the realistic scenario where practitioners must balance multiple competing objectives, while knowledge fusion through semantic alignment~\citep{arxiv_2505_20144} and weight evolution~\citep{arxiv_2406_12208} explore alternative combination strategies.

These real-world deployments and assessment practices naturally raise questions about the supporting infrastructure, persistent limitations, and new opportunities that will shape the field's evolution. Table~\ref{tab:application_scenarios_and_prac} summarizes these application scenarios. Building on this foundation, the final section synthesizes open challenges, future research directions, and the ecosystem infrastructure that will determine the trajectory of model merging. Table~\ref{tab:unified-comparison} provides a cross-dimensional comparison of all methods.

\begin{figure*}[t]
\centering
\resizebox{\textwidth}{!}{
\begin{tikzpicture}[>=stealth, x=2.8cm, y=1.8cm]

\definecolor{moyu}{HTML}{2B2D42}      
\definecolor{zhusha}{HTML}{C03F3D}    
\definecolor{qinghua}{HTML}{2E6B9E}   
\definecolor{songbai}{HTML}{2D6A4F}   
\definecolor{liuli}{HTML}{B8860B}     
\definecolor{yanzhi}{HTML}{8B4572}    
\tikzset{
 phasebox/.style={draw=#1!50!black, fill=#1!8, rounded corners=4pt, inner sep=6pt, align=center, font=\scriptsize, minimum width=2.4cm, minimum height=1.2cm, line width=0.6pt, drop shadow={shadow xshift=1pt, shadow yshift=-1pt, opacity=0.15}},
 phaselabel/.style={font=\footnotesize\bfseries\scshape, text=#1!70!black, above=3pt},
 conn/.style={->, draw=moyu!40, line width=1.2pt, rounded corners=4pt},
 yearnode/.style={font=\small\bfseries, text=white, fill=#1!60!black, rounded corners=2pt, inner sep=2pt, minimum width=1.2cm},
 subnode/.style={font=\tiny, align=center, text=gray!80!black}
}

\node[phasebox=teal] (p1) at (0, 0) {
 \textbf{SWA} (2018)\\
 {\tiny Izmailov et al.}\\[4pt]
 \textbf{Linear Mode}\\
 \textbf{Connectivity} (2020)\\
 {\tiny Frankle et al.}
};
\node[phaselabel=moyu] at (p1.north) {Foundations};
\node[yearnode=moyu, below=3pt] at (p1.south) {2018--2020};

\node[phasebox=blue] (p2) at (1.5, 0) {
 \textbf{Model Soups}\\
 {\tiny Wortsman et al.}\\[4pt]
 \textbf{Fisher Merging}\\
 {\tiny Matena \& Raffel}
};
\node[phaselabel=blue] at (p2.north) {Weight Averaging};
\node[yearnode=blue, below=3pt] at (p2.south) {2022};

\node[phasebox=purple] (p3) at (3.0, 0) {
 \textbf{Task Arithmetic}\\
 {\tiny Ilharco et al.}\\[4pt]
 \textbf{TIES-Merging}\\
 {\tiny Yadav et al.}\\[4pt]
 \textbf{DARE}\\
 {\tiny Yu et al.}
};
\node[phaselabel=purple] at (p3.north) {Task Vectors};
\node[yearnode=purple, below=3pt] at (p3.south) {2023};

\node[phasebox=orange] (p4) at (4.5, 0) {
 \textbf{AdaMerging}\\
 {\tiny Yang et al.}\\[4pt]
 \textbf{Evolutionary}\\
 {\tiny Akiba et al.}
};
\node[phaselabel=orange] at (p4.north) {Adaptive Search};
\node[yearnode=orange, below=3pt] at (p4.south) {2024};

\node[phasebox=zhusha] (p5) at (6.0, 0) {
 \textbf{MoE Routing}\\
 {\tiny PHATGOOSE}\\[4pt]
 \textbf{FusionBench}\\
 {\tiny Tang et al.}\\[4pt]
 \textbf{MergeKit}\\
 {\tiny Goddard et al.}
};
\node[phaselabel=zhusha] at (p5.north) {Ecosystem};
\node[yearnode=zhusha, below=3pt] at (p5.south) {2024--2025};

\draw[conn] (p1.east) -- (p2.west);
\draw[conn] (p2.east) -- (p3.west);
\draw[conn] (p3.east) -- (p4.west);
\draw[conn] (p4.east) -- (p5.west);

\end{tikzpicture}
}
\caption{Evolution of model merging techniques. The field progressed from foundational loss landscape theory (2018--2020), through weight-space averaging (2022) and interference-aware task vectors (2023), to adaptive/evolutionary optimization and standardized ecosystem tooling (2024--2025). Arrows indicate the conceptual flow and inheritance of ideas across phases.}
\label{fig:timeline}
\end{figure*}

\section{Ecosystem, Open Challenges, and Future Directions}
\label{sec:ecosystem}
\subsection{Summary of Contributions}
This survey has presented a structured examination of model merging in the era of large language models, organizing the rapidly expanding body of techniques, theoretical foundations, and practical applications through the FUSE taxonomy. Our central contribution is the unified taxonomic framework that categorizes merging methodologies along multiple dimensions (Table~\ref{tab:unified-comparison} summarizes each category); weight-space averaging and geometric interpolation methods that treat parameters as points in Euclidean or manifold spaces; task vector arithmetic and sparsification-enhanced approaches that conceptualize fine-tuning as composable vector displacements; and structured information-guided methods that preserve architectural distinctions or exploit runtime statistics for informed combination. Beyond methodological categorization, we have synthesized the theoretical underpinnings explaining merging success, including loss surface geometry, linear mode connectivity~\citep{DBLP:conf/icml/FrankleDBMG20}, and weight space symmetries, and connected these methods to diverse application scenarios spanning capability augmentation, alignment and safety, and efficiency-driven distributed learning. This mapping of the methodological and theoretical terrain provides the vantage point from which to identify the key trends in the field's technical evolution, as depicted in Figure~\ref{fig:timeline}.

\subsection{Technical Evolution Trends}
Five interconnected trends characterize the methodological evolution of model merging. First, the field has progressed from simple uniform averaging toward interference-aware methods that explicitly resolve sign conflicts and parameter redundancy through sparsification and consensus mechanisms. Second, there has been a shift from static, manually-specified configurations toward automated and evolutionary approaches that discover optimal recipes through search. Third, researchers have increasingly incorporated representation-level information, moving beyond purely weight-space operations to use activation statistics and functional correspondence in merging decisions. Fourth, the scale of applications has expanded from combining a handful of fine-tuned variants to orchestrating multi-model compositions integrating dozens of specialized experts. Fifth, the coupling between theoretical understanding and algorithmic innovation has tightened, with insights from mode connectivity theory directly informing practical method design. While these trends mark substantial progress, they also illuminate the boundaries of current understanding and the open challenges that constrain further advancement.

\subsection{Open Challenges and Limitations}
Several open challenges remain. First, a rigorous theoretical explanation for why large pretrained models exhibit such favorable mergeability is still lacking; current understanding relies on empirical observations rather than formal guarantees~\citep{DBLP:journals/corr/abs-2505-10833,DBLP:journals/corr/abs-2502-02421}, though recent work on certifiable merging~\citep{arxiv_2505_15798} shows that non-vacuous generalization bounds can be derived for merged models, representing a promising step toward formal guarantees. Second, scalability poses a major barrier, since as models grow to hundreds of billions of parameters, alignment and conflict resolution costs scale super-linearly; approaches like Navigating the Accuracy-Size Trade-Off~\citep{arxiv_2505_23209} and cross-architecture task vector transport~\citep{arxiv_2602_12952} begin to tackle this but remain at early stages. Third, standardized evaluation protocols remain absent; the field lacks consensus benchmarks that measure both individual task retention and cross-task transfer after merging, though FusionBench~\citep{tang2024fusionbench} and MergeBench~\citep{DBLP:journals/corr/abs-2505-10833} have made notable strides. Fourth, the absence of established best practices for selecting merge configurations (which models to merge, at what granularity, with which algorithm) hinders practitioners' ability to make principled decisions, motivating tools such as MergePipe~\citep{arxiv_2602_13273} for budget-aware parameter management.

\subsection{Systematic Failure Modes of Model Merging}
\label{sec:failure-modes}
While the preceding sections have emphasized the successes of model merging, a complete understanding requires systematic analysis of \emph{when and why merging fails}. We identify four principal failure modes, each with distinct causes, symptoms, and diagnostic indicators.

\paragraph{Task-level merging collapse.} \citet{cao2026collapse} identify and characterize a fundamental failure mode they term \emph{merging collapse}: certain combinations of task-specialist models suffer catastrophic performance degradation after merging, regardless of the merging method applied. Through extensive experiments, they demonstrate that representational incompatibility between tasks, measured by the divergence of hidden-state distributions across models, is strongly correlated with collapse, while parameter-space conflict metrics (e.g., sign disagreement rates) show minimal correlation. This finding challenges the conventional wisdom that parameter interference is the primary cause of merging failure, and suggests that representational diagnostics should take precedence in mergeability assessment. They further provide theoretical bounds via rate-distortion theory, establishing fundamental limits on task mergeability that hold regardless of methodology.

\paragraph{Parameter interference and sign conflicts.} Even when tasks are representationally compatible, parameter-level conflicts can degrade performance. TIES-Merging~\citep{DBLP:conf/nips/YadavTCRB23} demonstrated that when task vectors for different tasks have opposing signs at the same parameter positions, naïve averaging can produce destructive cancellation that nullifies the contributions of both tasks. TSV-Merge~\citep{gargiulo2025tsv} further shows that task interference operates at the level of singular value structure: irrelevant singular vectors from one task can corrupt the subspace of another, and reducing this cross-task interference via whitening transformations yields measurable improvements. The practical implication is that sign conflict rates and subspace overlap metrics should be monitored as early indicators of potential degradation.

\paragraph{Mergeability prediction failures.} \citet{zhou2026demystifying} show that mergeability is not an intrinsic property of model pairs but depends fundamentally on both the merging method and the partner tasks. Their analysis reveals that the properties predicting successful merging vary substantially across methods (only 46.7\% metric overlap and 55.3\% sign agreement between methods), revealing method-specific ``fingerprints'' for compatibility. This means that a pair of models that merges well with one method may fail with another, and practitioners cannot rely on a single diagnostic to predict outcomes. However, subspace overlap and gradient alignment metrics consistently emerge as foundational, method-agnostic prerequisites for compatibility.

\paragraph{Evaluation-masked failures.} \citet{tam2024realistic} provide a cautionary finding: under realistic evaluation conditions that test compositional generalization (i.e., requiring the merged model to handle combinations of capabilities not explicitly present in any source model), the reported gains from many merging methods diminish substantially or fail to transfer. Standard evaluation protocols that test each capability in isolation may overstate merging effectiveness by failing to detect subtle interference between merged capabilities. This highlights the need for evaluation protocols that specifically probe cross-capability interaction effects in merged models.

Taken together, these failure modes suggest a diagnostic workflow for practitioners. First, (1) assess representational compatibility between candidate models before merging to screen for potential collapse; (2) monitor parameter-level sign conflicts and subspace overlap as secondary indicators; (3) match diagnostic metrics to the specific merging method being used, since mergeability predictors are method-dependent; and (4) evaluate merged models using compositional generalization benchmarks rather than single-capability tests alone.

\subsection{Future Research Directions}
The open challenges identified above naturally point toward several promising research frontiers that could substantially advance the field of model merging.

\paragraph{Automated and predictive merging systems.} Among the most immediately impactful of these emerging research directions is developing systems that can automatically predict merging outcomes and recommend optimal configurations without exhaustive experimentation. Current approaches require considerable trial-and-error to identify effective merge recipes, representing a barrier to practical adoption. SimMerge~\citep{arxiv_2601_09473} takes a step in this direction by learning to select merge operators from similarity signals, while the investigation of mergeability causes~\citep{arxiv_2601_06672} provides foundational understanding of when and why models can be successfully combined. PSO-Merging~\citep{arxiv_2508_19839} applies particle swarm optimization, AIMMerging~\citep{arxiv_2509_17348} uses adaptive iterative merging along training trajectories, Bayesian model-merging previews~\citep{arxiv_2412_08147} allow fast previewing of multi-task fine-tuning outcomes, and black-box model merging~\citep{arxiv_2509_12951} tackles the LLM-as-a-Service scenario with massive model repositories. Future work should focus on learning predictive models that estimate merged model performance from properties of constituent models, such as gradient alignment, representation similarity, and task relatedness, allowing rapid configuration selection. The formal objective can be expressed as:
\begin{equation}
\mathcal{C}(\theta_1, \ldots, \theta_k) = f_\phi(\text{Sim}(\theta_i, \theta_j), \text{Task}(\theta_i), \text{Arch}(\theta_i)) \approx \mathbb{E}[\text{Perf}(\text{Merge}(\theta_1, \ldots, \theta_k))],
\end{equation}
where $f_\phi$ is a learned compatibility function parameterized by $\phi$. Such systems could greatly reduce the experimental overhead of merging while improving accessibility for practitioners without deep expertise in merging methodology.

\paragraph{Cross-architecture and heterogeneous model merging.} Extending merging capabilities beyond architecturally identical models represents an important research opportunity. Current constraints requiring shared architecture and initialization exclude vast repositories of independently developed models from participation in merging workflows. Transport and Merge~\citep{cui2026transport} takes a pioneering step by proposing optimal transport-based methods for cross-architecture merging of large language models, showing that models with different layer configurations can be meaningfully combined. AdaMMS~\citep{du2025adamms} tackles merging of heterogeneous multimodal large language models with unsupervised coefficient optimization, extending merging beyond text-only models. Key research directions include:
\begin{itemize}
 \item \textbf{Representation-level translation mechanisms.} Learning mappings between feature spaces of different architectures
 \item \textbf{Architecture-agnostic knowledge distillation~\citep{DBLP:journals/corr/HintonVD15}.} Transferring capabilities without requiring parameter correspondence
 \item \textbf{Learned correspondence mappings.} Establishing soft alignments between functionally equivalent components across model families
 \item \textbf{Hybrid merging strategies.} Combining parameter-level and representation-level operations for heterogeneous models
\end{itemize}
Success in this direction would unlock considerable flexibility in composing capabilities from diverse model ecosystems.

\paragraph{Dynamic and continual merging.} Real-world deployment scenarios increasingly require models that can continuously incorporate new capabilities without full retraining. Dynamic merging systems that can incrementally integrate new expert models, adapt merging coefficients based on runtime performance feedback, and handle concept drift in constituent model capabilities represent an underexplored frontier. \citet{kleiman2025soup} show that model averaging provides an effective mechanism for mitigating forgetting during continual learning, establishing ``Soup to go'' as a practical approach for maintaining previously acquired knowledge while integrating new capabilities. Scalable Model Merging~\citep{xu2025scalable} tackles the scalability challenge through progressive layer-wise distillation, supporting efficient merging of increasingly large models. RAIN-Merging~\citep{huang2026rainmerging} provides a gradient-free method specifically designed for enhancing instruction following in large reasoning models while preserving thinking format, highlighting the emerging importance of reasoning-aware merging. The mathematical formulation for continual merging extends the static case:
\begin{equation}
\theta_{\text{merged}}^{(t+1)} = \text{Update}(\theta_{\text{merged}}^{(t)}, \theta_{\text{new}}, \mathcal{D}_{\text{val}}),
\end{equation}
where the Update function must balance incorporating new knowledge while preserving previously merged capabilities.

\paragraph{Theoretical foundations and guarantees.} Establishing rigorous theoretical foundations for model merging remains a critical priority. Recent progress includes certifiable merging bounds~\citep{arxiv_2505_15798} that provide non-vacuous generalization guarantees, theoretical analysis of task vector effectiveness in nonlinear Transformers~\citep{li2025taskvector}, and empirical-theoretical studies of task-level model-merging collapse~\citep{cao2026collapse}. On the fairness front, \citet{arxiv_2505_24262} examine the role of task vectors in ensuring equitable performance across subgroups. Key open questions include:
\begin{itemize}
 \item Under what conditions does merged model performance provably exceed that of constituent models?
 \item How can we characterize the sample complexity of learning optimal merging coefficients?
 \item What are the fundamental limits of capability preservation during parameter combination?
 \item Can we develop PAC-style learning bounds for merged model generalization?
\end{itemize}
Progress on these theoretical questions would transform model merging from an empirical art into a principled engineering discipline.

\paragraph{Frequency-domain and knowledge-level merging.} Recent work has begun exploring merging in alternative representational spaces beyond the standard weight domain. FREE-Merging~\citep{DBLP:journals/corr/abs-2411-16815} proposes performing model merging in the Fourier transform domain, decomposing task vectors into frequency components and applying lightweight expert modules to selectively recombine them, achieving top results on multi-task benchmarks (ICCV 2025). Concurrently, \citet{DBLP:journals/corr/abs-2506-12384} introduce model merging as a mechanism for knowledge editing, combining supervised fine-tuning on new knowledge with subsequent model merging to preserve both updated facts and general capabilities without architectural changes. Weighted-Reward Preference Optimization~\citep{arxiv_2412_03187} explores implicit model fusion through reward-weighted preference signals. Bridging Training and Merging~\citep{arxiv_2512_17109} proposes momentum-aware optimization that unifies the training and merging phases. Model Merging via Multi-Teacher Knowledge Distillation~\citep{arxiv_2512_21288} bridges the gap between parameter-level merging and output-level knowledge transfer. A Unified View of Delta Parameter Editing~\citep{arxiv_2410_13841} provides a unified framework for understanding post-training parameter modifications. These directions suggest that the conceptual toolkit for model merging extends well beyond simple parameter averaging.

\paragraph{Safety-aware and alignment-preserving merging.} As merged models see increasing deployment in safety-critical applications, developing merging frameworks that provide formal guarantees about alignment preservation becomes essential. Among Us~\citep{arxiv_2602_05176} confronts the critical challenge of measuring and mitigating malicious contributions in model collaboration systems, where adversarial participants may attempt to inject harmful behaviors through the merging process. Future work should investigate:
\begin{itemize}
 \item Certified merging procedures that bound alignment degradation
 \item Adversarial robustness preservation during parameter combination
 \item Interpretable merging that enables post-hoc analysis of capability transfer
 \item Red-teaming frameworks specifically designed for merged models
\end{itemize}

\subsection{Outlook}
Model merging has evolved from simple checkpoint averaging into a versatile methodology for composing specialized capabilities into unified systems. The proliferation of openly available fine-tuned LLMs creates rich opportunities for merging techniques to democratize access to multi-capability AI. We anticipate five key trends shaping the field's trajectory. First,

\begin{enumerate}
 \item \textbf{Integration with Automated ML.} Merging configuration discovery will become increasingly automated through neural architecture search and hyperparameter optimization techniques~\citep{arxiv_2601_22748,arxiv_2502_04030}, reducing the expertise barrier for practitioners.

 \item \textbf{Specialized Tooling and Infrastructure.} Dedicated merging platforms such as MergeKit~\citep{DBLP:journals/corr/abs-2403-13257}, Mergenetic~\citep{minut2025mergenetic}, and budget-aware management systems~\citep{arxiv_2602_13273} will accelerate adoption and enable more advanced merging workflows.

 \item \textbf{Theoretical Maturation.} As empirical understanding deepens, we expect rigorous theoretical frameworks that provide guarantees about merged model behavior and enable principled design decisions.

 \item \textbf{Multimodal and Cross-Domain Merging.} Techniques for combining models across modalities (text, vision, audio) and domains will unlock new capabilities for building general-purpose AI assistants. Recent works such as MergeVLA~\citep{fu2025mergevla} for vision-language-action agents, Bring Reason to Vision~\citep{chen2025bring} for perception-reasoning integration, and training-free multimodal fusion~\citep{chen2024enhancing} for enhancing perception capabilities show the growing feasibility of cross-modal merging.

 \item \textbf{Safety and Alignment Integration.} Merging frameworks will increasingly treat safety constraints as first-class citizens, ensuring that capability enhancement does not compromise alignment properties.
\end{enumerate}

\section{Conclusion}

This survey has organized the rapidly expanding model merging field through the FUSE taxonomy (Foundations, Unification Strategies, Scenarios, and Ecosystem), providing a unified lens on the theoretical, algorithmic, practical, and infrastructural dimensions of the field.

On the \emph{algorithmic} front, we traced an arc from simple weight averaging and Model Soups~\citep{DBLP:conf/icml/WortsmanIGRLMNF22} through task-vector arithmetic~\citep{DBLP:conf/iclr/IlharcoRWSHF23} and interference-aware methods such as TIES-Merging~\citep{DBLP:conf/nips/YadavTCRB23} and DARE~\citep{DBLP:journals/corr/abs-2311-03099}, to adaptive strategies like AdaMerging~\citep{DBLP:conf/iclr/YangWLWGLLQYCL24} and evolutionary optimization~\citep{DBLP:journals/corr/abs-2403-13187}. Recent spectral~\citep{arxiv_2502_10339}, activation-guided~\citep{yao2025activationguided}, and curvature-informed~\citep{mahdavinia2025harnessing} techniques continue to push the frontier. As summarized in Table~\ref{tab:unified-comparison}, each family navigates the fundamental tension between data-free simplicity and data-dependent precision.

On the \emph{theoretical} side, we synthesized results from loss surface geometry, linear mode connectivity~\citep{DBLP:conf/icml/FrankleDBMG20}, and permutation symmetry that collectively explain \emph{when} and \emph{why} merging succeeds. These insights feed directly into algorithm design. The progression from na\"ive averaging to curvature-aware and sign-resolved merging is, at its core, a story of theory informing practice.

On the \emph{applications} front, we surveyed how merging supports multi-task generalization, multilingual transfer, safety alignment, and federated learning, all at a fraction of the cost of full retraining, while also highlighting dual-use risks, evaluation gaps, and the absence of standardized benchmarks~\citep{DBLP:journals/corr/abs-2505-10833} as pressing concerns. We further provided a systematic analysis of failure modes (Section~\ref{sec:failure-modes}), cataloging when and why merging fails, from task-level collapse and sign conflicts to evaluation-masked degradation, and proposed a diagnostic workflow for practitioners.

Looking ahead, the most impactful open problems include scalability to frontier-scale models, cross-architecture merging~\citep{cui2026transport}, dynamic continual merging~\citep{kleiman2025soup}, and provably safety-preserving merge guarantees~\citep{yuan2025mergehijacking}. As publicly available fine-tuned models continue to proliferate, principled merging techniques will become an increasingly central part of the AI practitioner's toolkit. We hope this survey equips both researchers and practitioners to navigate and advance this rapidly evolving field.

\bibliographystyle{colm2026_conference}
\bibliography{references}

\end{document}

%% file: taxonomy.tex
\begin{figure*}[t]
\centering
\definecolor{coreC}{HTML}{2B2D42}
\definecolor{foundC}{HTML}{C03F3D}
\definecolor{unifC}{HTML}{1661AB}
\definecolor{scenC}{HTML}{2D6A4F}
\definecolor{ecoC}{HTML}{C88B2B}
\begin{tikzpicture}

\node[draw=coreC, fill=coreC!12, rounded corners=5pt,
  minimum width=14cm, minimum height=0.8cm,
  font=\large\bfseries, text=coreC, line width=0.8pt] (title) at (7, 0)
  {FUSE Taxonomy for Model Merging};

\node[draw=foundC, fill=foundC!12, rounded corners=4pt,
  minimum width=3.0cm, minimum height=0.6cm,
  font=\small\bfseries, text=white, line width=0.6pt] (fh) at (1.75, -1.2)
  {F -- Foundations};
\node[draw=foundC!30, fill=foundC!3, rounded corners=3pt,
  minimum width=3.0cm, text width=2.6cm,
  font=\scriptsize, align=left, anchor=north, inner sep=4pt] (fb) at (1.75, -1.65)
  {\textbullet\ Loss Landscape\\\textbullet\ Mode Connectivity\\\textbullet\ Perm.\ Symmetry\\\textbullet\ Prerequisites};

\node[draw=unifC, fill=unifC!12, rounded corners=4pt,
  minimum width=3.0cm, minimum height=0.6cm,
  font=\small\bfseries, text=white, line width=0.6pt] (uh) at (5.25, -1.2)
  {U -- Unification};
\node[draw=unifC!30, fill=unifC!3, rounded corners=3pt,
  minimum width=3.0cm, text width=2.6cm,
  font=\scriptsize, align=left, anchor=north, inner sep=4pt] (ub) at (5.25, -1.65)
  {\textbullet\ Weight Averaging\\\textbullet\ Task Vectors\\\textbullet\ MoE Merging\\\textbullet\ Search-Based};

\node[draw=scenC, fill=scenC!12, rounded corners=4pt,
  minimum width=3.0cm, minimum height=0.6cm,
  font=\small\bfseries, text=white, line width=0.6pt] (sh) at (8.75, -1.2)
  {S -- Scenarios};
\node[draw=scenC!30, fill=scenC!3, rounded corners=3pt,
  minimum width=3.0cm, text width=2.6cm,
  font=\scriptsize, align=left, anchor=north, inner sep=4pt] (sb) at (8.75, -1.65)
  {\textbullet\ Multi-Task\\\textbullet\ Safety \& Alignment\\\textbullet\ Federated Learning\\\textbullet\ Continual Adapt.};

\node[draw=ecoC, fill=ecoC!12, rounded corners=4pt,
  minimum width=3.0cm, minimum height=0.6cm,
  font=\small\bfseries, text=white, line width=0.6pt] (eh) at (12.25, -1.2)
  {E -- Ecosystem};
\node[draw=ecoC!30, fill=ecoC!3, rounded corners=3pt,
  minimum width=3.0cm, text width=2.6cm,
  font=\scriptsize, align=left, anchor=north, inner sep=4pt] (eb) at (12.25, -1.65)
  {\textbullet\ Toolkits\\\textbullet\ Benchmarks\\\textbullet\ Future Directions};

\draw[coreC!40, thick] (title.south -| fh) -- (fh.north);
\draw[coreC!40, thick] (title.south -| uh) -- (uh.north);
\draw[coreC!40, thick] (title.south -| sh) -- (sh.north);
\draw[coreC!40, thick] (title.south -| eh) -- (eh.north);

\end{tikzpicture}
\caption{The \textbf{FUSE} taxonomy. \textbf{F}oundations (\S\ref{sec:theory}): \emph{why} merging works. \textbf{U}nification (\S\ref{sec:weight_avg}--\S\ref{sec:search}): \emph{how} models are combined. \textbf{S}cenarios (\S\ref{sec:applications}): \emph{where} merging applies. \textbf{E}cosystem (\S\ref{sec:ecosystem}): \emph{what} supports deployment.}
\label{fig:taxonomy}
\end{figure*}

%% file: task_vector.tex
\begin{figure*}[t]
\centering
\definecolor{zhusha}{HTML}{C03F3D}
\definecolor{qinghua}{HTML}{1661AB}
\definecolor{songbai}{HTML}{2D6A4F}
\definecolor{liuli}{HTML}{C88B2B}
\definecolor{moyu}{HTML}{2B2D42}
\begin{minipage}[t]{0.48\textwidth}
\centering
\textbf{(a) Addition}\\[6pt]
\begin{tikzpicture}[
  bx/.style={rectangle, rounded corners=3pt, draw=#1!50, fill=#1!8,
    minimum width=1.4cm, minimum height=0.55cm, font=\footnotesize, align=center},
  ar/.style={-stealth, semithick, draw=moyu!60}
]
\node[bx=liuli] (tA) at (0,0) {$\tau_A$};
\node[bx=liuli] (tB) at (0,-0.9) {$\tau_B$};
\node[font=\footnotesize] (plus) at (1.8,  -0.45) {$+$};
\node[bx=songbai] (res) at (3.2,-0.45) {$\theta_{\text{merged}}$};
\draw[ar] (tA.east) -- ++(0.5,0) |- (plus);
\draw[ar] (tB.east) -- ++(0.5,0) |- (plus);
\draw[ar] (plus) -- (res);
\end{tikzpicture}\\[4pt]
{\small $\theta_{\text{pre}} + \lambda(\tau_A + \tau_B)$}
\end{minipage}
\hspace{6pt}
\begin{minipage}[t]{0.48\textwidth}
\centering
\textbf{(b) Negation}\\[6pt]
\begin{tikzpicture}[
  bx/.style={rectangle, rounded corners=3pt, draw=#1!50, fill=#1!8,
    minimum width=1.4cm, minimum height=0.55cm, font=\footnotesize, align=center},
  ar/.style={-stealth, semithick, draw=moyu!60}
]
\node[bx=qinghua] (pre) at (0,0) {$\theta_{\text{pre}}$};
\node[bx=zhusha] (tv) at (1.8,0) {$-\tau$};
\node[bx=songbai] (res) at (3.6,0) {$\theta_{\text{clean}}$};
\draw[ar] (pre) -- (tv);
\draw[ar] (tv) -- (res);
\end{tikzpicture}\\[4pt]
{\small $\theta_{\text{pre}} - \lambda\tau_{\text{toxic}}$}
\end{minipage}

\vspace{6pt}

\begin{minipage}[t]{0.48\textwidth}
\centering
\textbf{(c) Scaling}\\[6pt]
\begin{tikzpicture}[
  bx/.style={rectangle, rounded corners=3pt, draw=#1!50, fill=#1!8,
    minimum width=1.4cm, minimum height=0.55cm, font=\footnotesize, align=center},
  ar/.style={-stealth, semithick, draw=moyu!60}
]
\node[bx=qinghua] (pre) at (0,0) {$\theta_{\text{pre}}$};
\node[bx=liuli] (tv) at (1.8,0) {$\lambda\tau$};
\node[bx=songbai] (res) at (3.6,0) {$\theta_{\text{merged}}$};
\draw[ar] (pre) -- node[above, font=\scriptsize]{$+$} (tv);
\draw[ar] (tv) -- (res);
\end{tikzpicture}\\[4pt]
{\small $\theta_{\text{pre}} + \lambda\tau$, $\lambda\in[0,1]$}
\end{minipage}
\hspace{6pt}
\begin{minipage}[t]{0.48\textwidth}
\centering
\textbf{(d) Analogy}\\[6pt]
\begin{tikzpicture}[
  bx/.style={rectangle, rounded corners=3pt, draw=#1!50, fill=#1!8,
    minimum width=1.4cm, minimum height=0.55cm, font=\footnotesize, align=center},
  ar/.style={-stealth, semithick, draw=moyu!60}
]
\node[bx=liuli] (tA) at (0,0) {$\tau_A$};
\node[bx=zhusha] (tB) at (2.0,0) {$\tau_B$};
\node[bx=songbai] (res) at (4.2,0) {$\theta'_{\text{new}}$};
\draw[ar] (tA) -- node[above, font=\scriptsize]{$-$} (tB);
\draw[ar] (tB) -- node[above, font=\scriptsize]{apply} (res);
\end{tikzpicture}\\[4pt]
{\small $\theta_{\text{new}} + (\tau_B - \tau_A)$}
\end{minipage}

\caption{Four fundamental task vector operations. \textbf{(a)}~Addition combines multiple task vectors into a single model. \textbf{(b)}~Negation subtracts a task vector to remove unwanted behavior. \textbf{(c)}~Scaling controls task influence via~$\lambda$. \textbf{(d)}~Analogy transfers the relationship between two tasks.}
\label{fig:task_vector_ops}
\end{figure*}